\documentclass[10pt,twocolumn,letterpaper]{article}

\usepackage[pagenumbers]{cvpr}

\usepackage{algorithm}
\usepackage{algorithmic}
\usepackage{array}
\usepackage{microtype}
\usepackage{xurl}
\usepackage{float}
\usepackage{multicol}
\usepackage{bm}
\usepackage{mathtools}
\usepackage{amsthm}
\usepackage[export]{adjustbox}
\usepackage{siunitx}
\theoremstyle{plain}

\theoremstyle{definition}

\theoremstyle{remark}

\definecolor{cvprblue}{rgb}{0.21,0.49,0.74}
\usepackage[pagebackref,breaklinks,colorlinks,allcolors=cvprblue]{hyperref}

\hypersetup{
  pdftitle={LoCoVSR: Local Context Diffusion Posterior Sampling for Video Super-Resolution},
  pdfauthor={Matan Ben Chorin and Michael Elad},
  pdfsubject={Face video super-resolution using pixel-space diffusion posterior sampling with local temporal conditioning and shared noise trajectories},
  pdfkeywords={video super-resolution, VSR, face video super-resolution, video restoration, face restoration, diffusion models, diffusion posterior sampling, DPS, pixel-space diffusion, local temporal conditioning, shared noise trajectories, temporal consistency, measurement consistency, inverse problems}
}

\newcommand{\method}{LoCoVSR}
\title{\method{}: Local Context Diffusion Posterior Sampling \\for Video Super-Resolution}
\author{Matan Ben Chorin\\
Faculty of Electrical and Computer Engineering\\
Technion - Israel Institute of Technology\\
Haifa, Israel\\
{\tt\small ben-chorin@campus.technion.ac.il}
\and
Michael Elad\\
Faculty of Computer Science Department\\
Technion - Israel Institute of Technology\\
Haifa, Israel\\
{\tt\small elad@cs.technion.ac.il}
}

\begin{document}
\maketitle

\begin{abstract}
Video super-resolution (VSR) is an ill-posed inverse problem that aims to reconstruct a high-resolution (HR) video from a noisy, low-resolution (LR) version of it.
We present \method{}, a diffusion-based VSR framework that leverages pixel-space denoising diffusion probabilistic models. \method{} integrates the Diffusion Posterior Sampling technique with spatio-temporal context learning, operating in a moving-average form. A localized window of adjacent LR frames is used for recovering each center frame, while applying a shared noise trajectory across all frames. The localized windowing enables processing of long videos without length limitations, supports parallel inference, and prevents error accumulation that may occur in recursive processing. Unlike prior methods, \method{} offers a simple yet very effective VSR solution, avoiding explicit optical flow estimation, or information loss caused by latent space processing. 
Trained on the VFHQ face dataset, \method{} achieves accurate, temporally consistent and high-quality upscaling with competitive results against recent diffusion-based VSR approaches. 
\end{abstract}

\section{Introduction}
Video Super-Resolution (VSR) is an ill-posed inverse problem, the purpose of which is to reconstruct a high-resolution (HR) video sequence from its degraded low-resolution (LR) observations, ensuring consistency with the measurements while providing a high visual fidelity outcome. As multiple HR videos may correspond to the same LR input, VSR must utilize strong prior information in order to recover a faithful and stable solution~\citep{baniya2024vsrsurvey}.

Beyond restoring per-frame sharpness, a central challenge in VSR lies in recovering \emph{faithful} details while maintaining \emph{temporal consistency}. These requirements are particularly stringent in face-centric applications, our focus. Subtle attributes such as facial expressions, wrinkles, or teeth structure may be barely discernible in the LR input, yet remain important for faithful facial reconstruction~\citep{gu2022vfhq,zou2025flair}.

Diffusion models (DMs) have established themselves as powerful generative priors~\citep{ho2020denoising,dhariwal2021diffusion}. Their success has led to a growing interest in leveraging them for inverse problems (e.g.~\citep{kawar2022denoising,wang2022zero,chung2023diffusion}). In this work we follow this theme, and use DMs for VSR. In doing so, this work aims to offer a VSR recovery algorithm that should be as simple as possible, while being highly effective and competitive. We focus on a frame-by-frame recovery approach, so as to leverage existing single image DMs, and thus enabling VSR for arbitrarily long video sequences. 

In single image restoration tasks, Diffusion Posterior Sampling (DPS)~\citep{chung2023diffusion} is considered as one of the leading techniques, having demonstrated significant potential in handling super-resolution (SR) and other tasks. Our baseline to embark from is a naive frame-by-frame extension of DPS to VSR, which clearly does not maintain temporal consistency, thus causing severe visible flickering and inconsistent dynamics between frames~\citep{cao2025zero,yeh2024diffir2vr,zhou2024upscale}. Our goal is to augment this baseline solution in a simple way, while producing a highly competitive VSR solution. 

We present \textbf{\method{}}, a DM-based framework for VSR that leverages pixel-space denoising diffusion probabilistic models (DDPMs) and integrates DPS with spatio-temporal context. The context is acquired through a localized conditioning strategy: for each target (central) frame, the model exploits a compact temporal neighborhood of adjacent LR frames (e.g., $\pm 5$ frames).\footnote{With an empty local window, we get the baseline as a special case.} The denoiser is conditioned on this neighborhood for recovering the central frame. We further enforce measurement consistency by applying a DPS guidance term to the central frame.
This design enables the model to capture local motion dynamics directly from the measurements while avoiding explicit optical flow estimation. In contrast, alternative VSR methods tend to rely on explicit optical flow estimation to maintain temporal consistency~\citep{basicvsrpp,liang2022vrt,mgldvsr2024}. However, in LR or noisy videos, flow estimation is fragile; misalignment errors can propagate through the pipeline, resulting in warping artifacts and ghosting~\citep{cao2025zero}. Moreover, by operating in the pixel space, \method{} avoids potential complications associated with a latent representation, as commonly done in latent diffusion models (LDMs)~\citep{rombach2022highresolution}. While such latent mappings can improve computational efficiency, they may suppress subtle high-frequency details essential for high-fidelity restoration~\citep{yi2025tvt}, and add complications when handling inverse problems~\citep{rout2023solving,song2024resample,raphaeli2025silo}. 

To further enforce temporal coherence, we introduce a \emph{shared noise trajectory} tailored to our fully parallel frame-wise posterior sampling framework. While related stochastic synchronization strategies have been explored in prior video diffusion methods~\citep{kwon2025solving,cao2025zero}, our formulation shares the sampling stochasticity across independently reconstructed frames. This inductive bias reduces sampling-induced temporal variability, keeping fine details temporally coherent.

Localized moving-average-like windowing also yields favorable computational and algorithmic properties. As inference operates on bounded temporal neighborhoods, the approach naturally supports long sequences without requiring full video optimization, removes dependence on sequence length, and enables parallel processing of multiple windows.
This stands in contrast to strictly serial and recursive pipelines, which condition each frame on previous reconstructions and may suffer from error accumulation, where early mistakes compound over time~\citep{chan2022tradeoffs}.

In our experiments on the VFHQ face dataset, we consider $2,4 ~\mbox{and}~ 8\times$ VSR. We evaluate \method{} alongside several recent and leading diffusion and non-DM-based VSR methods: SVI-Diffusion~\citep{kwon2025solving}, Upscale-A-Video~\citep{zhou2024upscale}, VISION-XL~\citep{visionxl2025}, UltraVSR~\citep{ultravsr2024}, StableVSR~\citep{stablevsr2024}, PS-SR~\citep{wu2026pssr}, and BasicVSR++~\citep{basicvsrpp}. Performance is evaluated under a unified protocol, using PSNR, SSIM, LPIPS, and FVD to assess reconstruction fidelity, perceptual quality, and video level consistency. We further provide qualitative examples emphasizing the reliable restoration of delicate facial details. Our experiments show that 
\method{} achieves high-quality SR upscaling with competitive quantitative and qualitative results. Visual examples demonstrate faithful and temporally consistent recovery of fine facial attributes that are barely discernible in the LR input.

In summary, our main contributions are as follows:
\begin{itemize}
    \item \textbf{\method{}:} We introduce a simple yet highly effective pixel-space diffusion framework for high-fidelity face-centric VSR that integrates a spatio-temporal conditioned prior with DPS guidance. This enables \textbf{temporally consistent}, \textbf{high-quality reconstructions} that recover fine details while remaining \textbf{faithful} to the LR measurements.

    \item \textbf{Localized Conditioning Strategy:} We propose a scheme that utilizes a compact temporal neighborhood of adjacent LR frames to reconstruct each target frame. This approach enables the model to capture complex motion dynamics directly from the measurements, eliminating the need for \textbf{explicit optical flow estimation}. 
    \method{} enables processing of long videos \textbf{without length limitations}, supports \textbf{parallel inference}, and \textbf{prevents the error accumulation} that may occur in serial processing.
    
    \item \textbf{Temporal Stabilization:} To address the inherent sampling randomness of DMs, we implement a \textbf{shared noise trajectory} across frames. This provides a strong inductive bias that minimizes sampling-induced flickering and keeps recovered fine details \textbf{temporally coherent}.

    \item \textbf{High-Fidelity Restoration:} Evaluations on the \textbf{VFHQ face dataset} show that \method{} provides superior quantitative and qualitative results, faithfully restoring delicate facial attributes, such as textures and fine structures.

\end{itemize}

\section{Background}
We define a video as a frame sequence $\mathbf{x}=\{\mathbf{x}_i\}_{i=1}^{I}$. Its degraded observation is given by $\mathbf{y}=\{\mathbf{y}_i\}_{i=1}^{I}$, a sequence of the same length, where 
\begin{equation}
\label{eq:forward}
\mathbf{y}_i = \mathcal{A}(\mathbf{x}_i) + \mathbf{n}_i, \qquad i=1,\ldots,I.
\end{equation}
Here, $\mathbf{x}_i$ denotes the unknown HR frame, $\mathbf{y}_i$ is the observed LR (and potentially noisy) frame, and $\mathbf{n}_i$ represents measurement noise. Throughout this work, $\mathcal{A}$ denotes the known $\times s$ bicubic downsampling operator (in this work $s=2,4,~\mbox{or}~8$), and $\mathcal{A}^{\top}$ denotes its adjoint. Importantly, the noise term is \emph{frame dependent}: $\mathbf{n}_i$ varies across time, reflecting changes in sensor noise and acquisition conditions.
Due to the non-invertibility of $\mathcal{A}$, multiple HR videos may correspond to the same LR input. Thus, VSR must utilize a strong prior in order to recover a faithful solution. This brings us naturally to the topic of DMs. 

DMs may learn such a prior distribution $p(\mathbf{x})$, when aiming to generate high-quality samples through an iterative denoising (reverse) process. For inverse problems, the goal is to recover an unknown signal $\mathbf{x}$ that is perceptually plausible and consistent with the noisy measurements $\mathbf{y}$. A natural Bayesian approach is to perform posterior inference:
\begin{equation}
\label{eq:posterior}
p(\mathbf{x}\mid \mathbf{y}) \propto p(\mathbf{y}\mid \mathbf{x})\,p(\mathbf{x}),
\end{equation}
where $p(\mathbf{x})$ denotes the prior represented by the DM, and $p(\mathbf{y}\mid \mathbf{x})$ is the likelihood induced by the degradation model in Eq.~\eqref{eq:forward}. 
DPS applies this principle by injecting likelihood-based guidance along the reverse diffusion trajectory~\cite{chung2023diffusion}. DPS addresses the intractability of evaluating the likelihood at intermediate diffusion states (due to the dependence of $\mathbf{x}$ on the diffusion time $t$) via an approximation that is based on the posterior mean estimate $\hat{\mathbf{x}}_0$  (estimated clean sample), obtained at each denoising step through Tweedie’s formula. More specifically, given a diffusion state $\mathbf{x}_{i,t}$, DPS defines the data fidelity objective
\begin{equation}
\mathcal{L}_{\text{data}}(\mathbf{x}_{i,t})
= \left\|
\mathbf{y}_i - \mathcal{A}\!\left(\hat{\mathbf{x}}_{i,0}(\mathbf{x}_{i,t})\right)
\right\|_2^2,
\label{eq:dps_loss}
\end{equation}
and then steers the sampling by incorporating the guidance term $\nabla_{\mathbf{x}_{i,t}}\mathcal{L}_{\text{data}}(\mathbf{x}_{i,t})$ to repeatedly enforce measurement consistency throughout the generation (reverse) trajectory. 

All the above assumes that the DM deployed operates on the whole video as one entity $\mathbf{x}$, implying that the denoiser to be used operates on full video sequences, as recent video DMs do (e.g.,~\cite{zhan2025vdmvsr,chen2025dove}). Note, however, that the likelihood term above is separable, operating on a frame-by-frame basis. An appealing approach is to operate on single frames separately along these lines, using a single-image-based prior. \method{} adopts this principle: For the recovery of the frame $\mathbf{x}_i$, $\mathbf{y}_i$ and its neighboring frames provide context for temporal consistency, whereas the likelihood-based term is anchored to the central observation $\mathbf{y}_i$ to guarantee fidelity to the LR input, just as DPS does.

\section{Related work}

\noindent\textbf{Deep VSR.}
Traditional deep learning approaches for VSR primarily employ sliding window, recurrent, or attention-based architectures ~\citep{liang2022rvrt,zhang2024realviformer} to aggregate temporal information across frames.
Methods such as BasicVSR~\cite{basicvsr}, BasicVSR++~\cite{basicvsrpp}, EDVR~\cite{edvr}, and their variants often rely on explicit motion compensation, usually involving optical flow~\cite{basicvsrpp, liang2022vrt} or deformable alignment~\cite{edvr}, to register adjacent frames before reconstruction.
While effective on synthetic or mildly degraded inputs, explicit motion estimation becomes less accurate with increasing degradation severity, where misalignment can propagate through the pipeline and produce warping and ghosting artifacts~\citep{cao2025zero}.

Regression-based frameworks trained with pixelwise reconstruction losses (e.g., MSE or $\ell_1$) operate on a different point in the perception distortion trade-off~\cite{blau2018perception}, typically favoring distortion-oriented reconstruction over the synthesis of high perceptual quality outcomes~\cite{ledig2017srgan,baniya2024vsrsurvey}.

\paragraph{Diffusion solvers for inverse problems.}
DDPMs~\cite{ho2020denoising} have become powerful generative priors for image restoration applications.
Task-agnostic solvers incorporate measurement fidelity during reverse diffusion without retraining, deployed in various ways. Early approaches include consistency or projection based strategies (e.g., DDNM/DDRM style updates)~\citep{kawar2022denoising,wang2022zero}, while DPS~\cite{chung2023diffusion} derives likelihood gradient guidance from the denoiser’s posterior mean (Tweedie) estimate, enabling plug-and-play recovery under noisy measurements.
However, extending these image centric solvers to video is nontrivial: independently sampling each frame might amplify diffusion stochasticity, which turns into temporal inconsistency, thereby yielding perceptible flicker across time~\cite{zhou2024upscale,cao2025zero,yeh2024diffir2vr}. Recent inference-time video solvers therefore couple frames explicitly~\citep{cao2025zero,yeh2024diffir2vr,zhan2025vdmvsr}.
SVI-Diffusion~\cite{kwon2025solving}, for example, follows a batchwise sampling scheme, where all frames in the batch are processed simultaneously at each reverse step. This approach treats time as a batch dimension, synchronizes the stochastic noise across frames, and enforces spatio-temporal data consistency by applying conjugate gradient (CG) optimization. VISION-XL~\cite{visionxl2025} extends this approach to LDMs.

\paragraph{DMs for VSR.} 
Recent DM-based VSR methods commonly build on LDM backbones such as Stable-Diffusion or SDXL~\citep{rombach2022highresolution,podell2023sdxl}.
Methods such as StableVSR~\citep{stablevsr2024}, Upscale-A-Video~\citep{zhou2024upscale}, UltraVSR~\citep{ultravsr2024}, and PS-SR~\citep{wu2026pssr} introduce temporal components such as 3D convolutions, recurrent propagation, local/global processing, optical-flow-based alignment, or lightweight adaptation mechanisms to improve temporal coherence and inference efficiency. 
However, these often come at the cost of additional architectural or algorithmic complexity, and rely on inference schemes that either process the video jointly, imposing memory and length constraints, or propagate information serially, limiting parallelism and risking error accumulation~\citep{du2025patchvsr,chan2022tradeoffs}.

In addition, LDM-based restoration pipelines face fundamental obstacles when high-fidelity recovery is required.
First, the VAE encoder-decoder induces an information bottleneck: compressing to a latent space can discard subtle high-frequency cues (e.g., fine skin texture, small moles, tooth boundaries), which are difficult to reconstruct after decoding, reducing the fidelity of fine structures~\citep{yi2025tvt}.
Second, many widely used LDM priors are text conditioned; in restoration settings where reliable text conditioning is unavailable, applying them typically requires heuristic choices (e.g., null text prompting, prompt/embedding optimization, or regularization to keep latents feasible)~\citep{visionxl2025}. These choices can bias the solution towards semantic plausibility rather than strict measurement fidelity.
Moreover, enforcing data consistency is most straightforward in pixel space; in LDM pipelines, it often requires frequent decoding (and sometimes re-encoding) within iterative sampling ~\cite{visionxl2025}, which can introduce additional approximation error and destabilize fine temporal details. This is particularly problematic in face-centric VSR, where even subtle frame-to-frame inconsistencies are visually salient.

In contrast, our approach trains a pixel-space single-frame denoising prior with localized spatio-temporal conditioning, allowing the model to learn from the time dimension rather than relying only on inference-time synchronization across frames. The resulting framework is elegant: it captures local temporal dynamics without explicit optical-flow estimation, avoids full video joint processing, and prevents error accumulation associated with serial propagation. 

\begin{figure*}[t]
\centering
\includegraphics[
    width=\textwidth,
    trim=30 25 29 17,
    clip
]{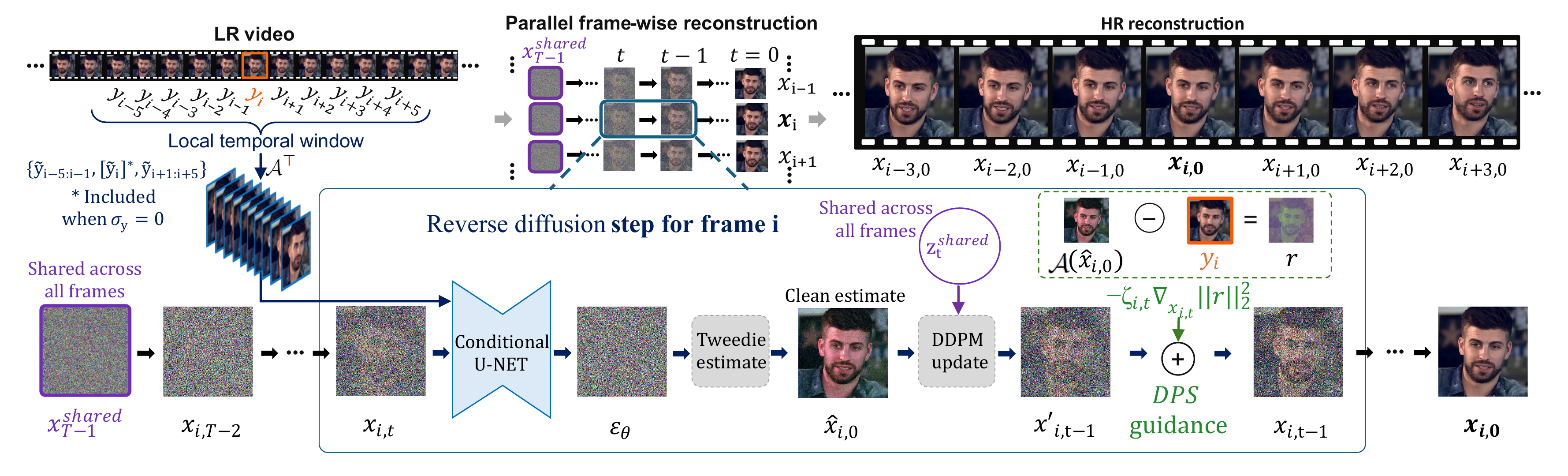}
\caption{
\textbf{Overview of \method{}.}
Given a LR video, each HR frame is reconstructed through a frame-wise reverse diffusion process. For a target frame $i$, a window of $\pm 5$ LR frames is back-projected to the HR grid ($\mathcal{A}^{\top}$) and concatenated channel-wise with the current diffusion state $\mathbf{x}_{i,t}$ as input to a pixel-space conditional U-Net. At each reverse step, Tweedie's formula yields the clean estimate $\hat{\mathbf{x}}_{i,0}$, followed by a DDPM update and a DPS correction that enforces consistency with the measurement $\mathbf{y}_i$. A single initial diffusion state and noise trajectory are shared across the video sequence to synchronize sampling stochasticity. Since no frame depends on another frame's reconstruction, all frames can be processed in parallel. The central LR observation is included in the denoiser conditioning in the noiseless setting.
}
\label{fig:overview}
\end{figure*}

\section{Method}
\label{sec:method}
We now present \method{}, a pixel-space diffusion framework for high-fidelity face VSR. Given LR observations $\mathbf{y}=\{\mathbf{y}_i\}_{i=1}^{I}$ generated according to the degradation model in Eq.~\eqref{eq:forward},  our goal is to sample from an approximation of the posterior in Eq.~\eqref{eq:posterior} while maintaining temporal coherence across the reconstructed frames. Our design follows this rationale: the conditional diffusion \emph{prior} provides face-domain realism and temporal awareness, while the \emph{likelihood} term enforces measurement fidelity. Accordingly, \method{} consists of three components: (i) \emph{localized spatio-temporal conditioning} of the denoiser (Sec.~\ref{sec:method_cond}); (ii) \emph{per-frame posterior guidance}, which anchors each reconstruction to its own LR measurement (Sec.~\ref{sec:method_guidance}); and (iii) a \emph{shared noise trajectory}, which synchronizes the sampling stochasticity across frames (Sec.~\ref{sec:method_noise}). An overview is given in Fig.~\ref{fig:overview} and the complete inference procedure is summarized in \Cref{alg:method}. \method{} requires neither explicit motion estimation nor joint processing of the full video, and therefore enables fully parallel frame-wise inference (Sec.~\ref{sec:method_inference}).

\subsection{Localized spatio-temporal conditioning}
\label{sec:method_cond}
A naive frame-by-frame application of an image diffusion prior ignores temporal dependencies, while joint video diffusion is costly and unnecessarily couples each frame to the entire video. To bridge these two extremes, we give the denoiser direct access to the local temporal context of the input frame. Specifically, rather than modeling the marginal frame prior $p(\mathbf{x}_i)$, we learn a \emph{conditional} prior $p(\mathbf{x}_i \mid \mathbf{c}_i)$, where $\mathbf{c}_i$ gathers the LR observations in a temporal window of radius $k$ ($k=5$ unless stated otherwise) around frame $i$:
$\mathbf{c}_i \;=\; \big(\, \tilde{\mathbf{y}}_{i+j} \,:\, j\in\mathcal{W} \,\big)
$,
where $\tilde{\mathbf{y}}_i=\mathcal{A}^\top(\mathbf{y}_i)$ and $\mathcal{W}=(-k,\ldots,-1,1,\ldots,k)$. The central observed frame is incorporated conditionally, as explained below. Adjacent LR frames often provide complementary observations of the same content under small inter-frame motion. Conditioning on this local temporal context steers the prior towards HR reconstructions that agree with the observed temporal evidence, enabling the recovery of fine details that are ambiguous in the target LR frame alone.

\textbf{Architecture.}
The conditioning is implemented by channel-wise concatenation. We train separate denoiser variants for the noiseless and noisy degradation
settings; thus, the treatment of the central LR observation is fixed for each
model rather than selected at inference time.
Let $\sigma_y$ denote the measurement noise standard deviation, and define
the optional central frame conditioning block for $k>0$ as

$\delta_{\sigma_y}(\tilde{\mathbf{y}}_i)\coloneqq\begin{cases}\tilde{\mathbf{y}}_i, & \sigma_y=0,\\\varnothing, & \sigma_y>0,\end{cases}$
where $\varnothing$ denotes omission from the channel wise concatenation.
The denoiser input for frame $i$ is then
$\tilde{\mathbf{x}}_{i,t}
=
\left[
\tilde{\mathbf{y}}_{i-k:i-1},
\delta_{\sigma_y}(\tilde{\mathbf{y}}_i),
\mathbf{x}_{i,t},
\tilde{\mathbf{y}}_{i+1:i+k}
\right]$ for $k>0$.
The model predicts
$\boldsymbol{\epsilon}_{\theta}(\tilde{\mathbf{x}}_{i,t},t)$
from the noisy target frame and its local LR temporal context. With the default $k=5$, this yields $36$ input channels for the noiseless model and $33$ channels for the noisy model (due to the color-layers). The network thus treats the temporal window as a single multi-channel image: apart from widening the first convolution, the U-Net backbone~\citep{dhariwal2021diffusion} is left entirely unchanged without explicit motion estimation or dedicated temporal modules. As shown empirically, this \emph{lightweight conditioning interface is sufficient} for the denoiser to capture temporal dependencies and leverage motion cues directly from the LR measurements (see \cref{sec:ablations} and App.~\ref{app:full_temporal_ablation}).

\paragraph{Training.}
We fine-tune a pretrained FFHQ pixel-space DDPM on VFHQ (see \cref{sec:experiments_setup}) using the standard DDPM~\citep{ho2020denoising} $\epsilon$-prediction objective. For a clean HR frame $\mathbf{x}_i$, noise $\boldsymbol{\epsilon}\sim\mathcal{N}(0,I)$, and diffusion time $t$, we form
$\mathbf{x}_{i,t}=\sqrt{\bar\alpha_t}\,\mathbf{x}_i+\sqrt{1-\bar\alpha_t}\,\boldsymbol{\epsilon}
$, 
and minimize
$\mathcal{L}_{\mathrm{diff}}
=\mathbb{E}_{i,t,\boldsymbol{\epsilon}}
\left\|\boldsymbol{\epsilon}-\boldsymbol{\epsilon}_{\theta}\!\left(\tilde{\mathbf{x}}_{i,t},t\right)\right\|_2^2$,
with learned variance as in the underlying guided-DM (full details in App.~\ref{app:training_details}).
For sampling, we use the equivalent conditional score estimate
\begin{equation}
\label{eq:score_from_epsilon}
\mathbf{s}_{\theta}
\left(\tilde{\mathbf{x}}_{i,t},t\right)
\coloneqq
-
\frac{
\boldsymbol{\epsilon}_{\theta}
\left(\tilde{\mathbf{x}}_{i,t},t\right)}
{\sqrt{1-\bar{\alpha}_t}},
\end{equation}
which approximates
$\nabla_{\mathbf{x}_{i,t}}
\log p_t(\mathbf{x}_{i,t}\mid\mathbf{c}_i)$.

\subsection{Per-frame posterior guidance}
\label{sec:method_guidance}

The temporally conditioned denoiser provides a powerful prior, but conditioning alone does not guarantee fidelity to the target observation; we therefore explicitly enforce consistency with the corresponding LR measurement. Following the Bayesian formulation of \cref{eq:posterior}, we sample from the posterior
$p(\mathbf{x}_i \mid \mathbf{y}_i, \mathbf{c}_i)\;\propto\;p(\mathbf{y}_i \mid \mathbf{x}_i)\;p_\theta(\mathbf{x}_i \mid \mathbf{c}_i)$,
where the likelihood is induced by the degradation of \cref{eq:forward} and is anchored \emph{only} to the central observed frame $\mathbf{y}_i$, while the neighboring frames act as context through the prior. This
separation allows the temporal context to contribute complementary information
without treating potentially misaligned neighboring frames as direct
measurement constraints, retaining independent and parallel sampling of
the target frames.

Sampling proceeds with the DPS scheme~\citep{chung2023diffusion}
applied to the conditioned score estimator. We use $T=1000$ reverse diffusion steps per frame. At each reverse step $t$, the posterior mean estimate is obtained via Tweedie's formula,
$\hat{\mathbf{x}}_{i,0}
\left(\mathbf{x}_{i,t},\mathbf{c}_i\right)
\approx
\frac{1}{\sqrt{\bar{\alpha}_t}}
\left(
\mathbf{x}_{i,t}
+
\left(1-\bar{\alpha}_t\right)
\mathbf{s}_{\theta}
\left(\tilde{\mathbf{x}}_{i,t},t\right)
\right)$.

A standard ancestral DDPM update produces an intermediate iterate $\mathbf{x}'_{i,t-1}$, and the data-fidelity objective of \cref{eq:dps_loss} is enforced through a gradient correction,
$\mathbf{x}_{i,t-1}
=\mathbf{x}'_{i,t-1}
-\zeta_{i,t}\,\nabla_{\mathbf{x}_{i,t}}
\big\|\mathbf{y}_i-\mathcal{A}\big(\hat{\mathbf{x}}_{i,0}\big)\big\|_2^2$.
As in DPS, the step size is normalized by the current residual,
$\zeta_{i,t}=\frac{\zeta'}{\big\|\mathbf{y}_i-\mathcal{A}\big(\hat{\mathbf{x}}_{i,0}\big)\big\|_2}$,
leaving a single scalar hyperparameter $\zeta'$, selected once per degradation setting on the validation split. 

\subsection{Shared noise trajectory}
\label{sec:method_noise}

Diffusion sampling is stochastic, and independent noise realizations across frames can translate into temporally inconsistent fine details that are perceived as flicker. We therefore synchronize the sampling randomness across the entire video sequence. Specifically, the initial Gaussian state and every stochastic noise injection in the reverse process are sampled once and reused for all frames:
$\mathbf{x}_{i,{T-1}} = \mathbf{x}_{T-1}^{\mathrm{shared}}, \quad
\mathbf{z}_{i,t} = \mathbf{z}_t^{\mathrm{shared}}, \quad \forall i $.
As a result, all frames are generated under the same stochastic realization, while their differences arise from the frame-specific measurement $\mathbf{y}_i$ and temporal context $\mathbf{c}_i$. This aligns the sampling randomness across time, so high-frequency details change only in response to the actual frame content rather than independent noise realizations. The resulting sequence-level coupling method enhances temporal coherence without additional training or inference costs.

\subsection{Parallel frame-wise inference}
\label{sec:method_inference}

\textbf{Boundary handling.}
For target frames near the beginning or end of the sequence, missing neighbors are filled by mirror padding with respect to the available side, so that every frame is processed with a complete $\pm k$ window and no artificial discontinuities are introduced at sequence edges.

\textbf{Parallel inference.}
Each frame's reverse process depends only on its measurement and local LR window; the shared noise trajectory synchronizes stochastic variation without adding dependencies between reconstructions. Thus, (i) all frames can be sampled in parallel rather than serially; (ii) memory scales with the window size rather than with the sequence, enabling arbitrarily long videos without whole video joint batch memory constraints; and (iii) independence from previous reconstructions eliminates autoregressive error accumulation and drift. Overall, \method{} delivers accurate, temporally consistent reconstructions with faithful detail recovery, while preserving the scalability and robustness of fully parallel frame-wise sampling for high-fidelity VSR.

\begin{algorithm}[tb]
\small
\caption{\method{} inference}
\label{alg:method}
\begin{algorithmic}[1]
\REQUIRE LR frames $\{\mathbf{y}_i\}_{i=1}^{I}$, conditional score model $\bm{s}_\theta$, operator $\mathcal{A}$, measurement noise standard deviation $\sigma_y$, window size $k$, step size $\zeta'$, diffusion steps $T$, diffusion schedule
$\{\alpha_t,\bar{\alpha}_t,\beta_t,\tilde{\sigma}_t\}_{t=0}^{T-1}$,  

\STATE $\mathbf{x}_{T-1}^{\mathrm{shared}},\{\mathbf{z}_t^{\mathrm{shared}}\}_{t=1}^{T-1}\overset{\mathrm{i.i.d.}}{\sim}\mathcal{N}(\mathbf{0},\mathbf{I})$; $\mathbf{z}_0^{\mathrm{shared}}=\mathbf{0}$

\STATE $\displaystyle
\bar{\mathbf{y}}_{1-k:I+k}
\leftarrow
\operatorname{MirrorPad}_{k}
\left(\mathbf{y}_{1:I}\right)$
\FOR{$i=1$ {\bfseries to} $I$ \textbf{in parallel}}   
   
    \STATE $\mathbf{x}_{i,{T-1}} \leftarrow \mathbf{x}_{T-1}^{\mathrm{shared}}$
 
    \STATE $\displaystyle
    \tilde{\mathbf{y}}_{i-k:i+k}
    \leftarrow
    \mathcal{A}^{\top}\!\left(
    \bar{\mathbf{y}}_{i-k:i+k}
    \right)$
  
    \FOR{$t=T-1$ {\bfseries downto} $1$}
        \STATE $\displaystyle
        \widehat{\mathbf{s}}_{i,t}
        \leftarrow
        \bm{s}_{\theta}\!\left(
        \left[
        \tilde{\mathbf{y}}_{i-k:i-1},
        \delta_{\sigma_y}(\tilde{\mathbf{y}}_i),
        \mathbf{x}_{i,t},
        \tilde{\mathbf{y}}_{i+1:i+k}
        \right],
        t
        \right)$
        
        \STATE $\displaystyle
        \widehat{\mathbf{x}}_{i,0}
        \leftarrow
        \frac{1}{\sqrt{\bar{\alpha}_{t}}}
        \left(
        \mathbf{x}_{i,t}
        +
        \left(1-\bar{\alpha}_{t}\right)
                \widehat{\mathbf{s}}_{i,t}\right)$

        \STATE
        \resizebox{\linewidth}{!}{$\displaystyle
        \mathbf{x}'_{i,t-1}
        \leftarrow
        \frac{\sqrt{\alpha_t}\left(1-\bar{\alpha}_{t-1}\right)}
             {1-\bar{\alpha}_t}\mathbf{x}_{i,t}
        +
        \frac{\sqrt{\bar{\alpha}_{t-1}}\beta_t}
             {1-\bar{\alpha}_t}\widehat{\mathbf{x}}_{i,0}
        +
        \tilde{\sigma}_t\mathbf{z}_{t}^{\mathrm{shared}}
        $}
        
        \STATE $\displaystyle
        \zeta_{i,t}
        \leftarrow
        \frac{\zeta'}
        {\left\|
        \mathbf{y}_i-
        \mathcal{A}\!\left(\widehat{\mathbf{x}}_{i,0}\right)
        \right\|_2}$
        
        \STATE $\displaystyle
        \mathbf{x}_{i,t-1}
        \leftarrow
        \mathbf{x}'_{i,t-1}
        -
        \zeta_{i,t}
        \nabla_{\mathbf{x}_{i,t}}
        \left\|
        \mathbf{y}_i-
        \mathcal{A}\!\left(\widehat{\mathbf{x}}_{i,0}\right)
        \right\|_2^2$
    \ENDFOR
\ENDFOR

\STATE \textbf{Output:} $\{{\mathbf{x}}_{i,0}\}_{i=1}^{I}$
\end{algorithmic}
\end{algorithm}

\section{Experiments}
\label{sec:experiments}
We evaluate \textbf{\method{}} under both noiseless and noisy degradation, compare with recent VSR methods under the same degradation and metric protocol, and ablate architectural choices and hyperparameters.

\begin{figure*}[t]
\centering
\setlength{\tabcolsep}{0pt}
\renewcommand{\arraystretch}{1.0}

\begin{tabular}{@{}c c c c c c@{}}
&
\small\textbf{LR input}
&
\small\textbf{SVI-Diffusion}
&
\small\textbf{Upscale-A-Video}
&
\small\textbf{Ours}
&
\small\textbf{Ground truth}
\\

\raisebox{-.5\totalheight}{%
    \makebox[0.070\textwidth][c]{%
        \shortstack{%
            \small\bfseries Noiseless\\
            \small $\sigma_y=0$
        }%
    }%
}
&
\raisebox{-.5\totalheight}{%
    \includegraphics[interpolate=false,width=0.185\textwidth]
    {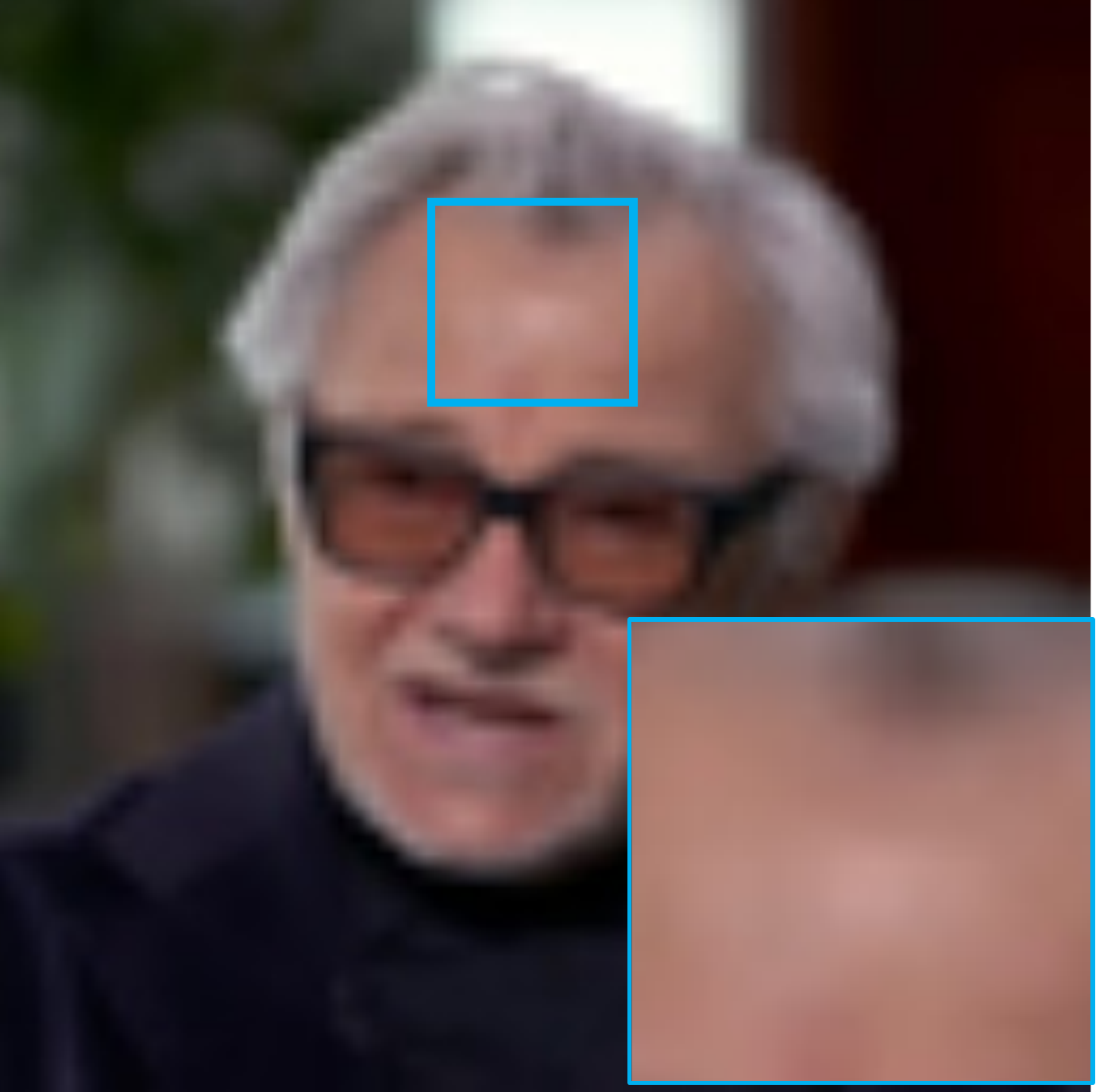}%
}
&
\raisebox{-.5\totalheight}{%
    \includegraphics[interpolate=false,width=0.185\textwidth]
    {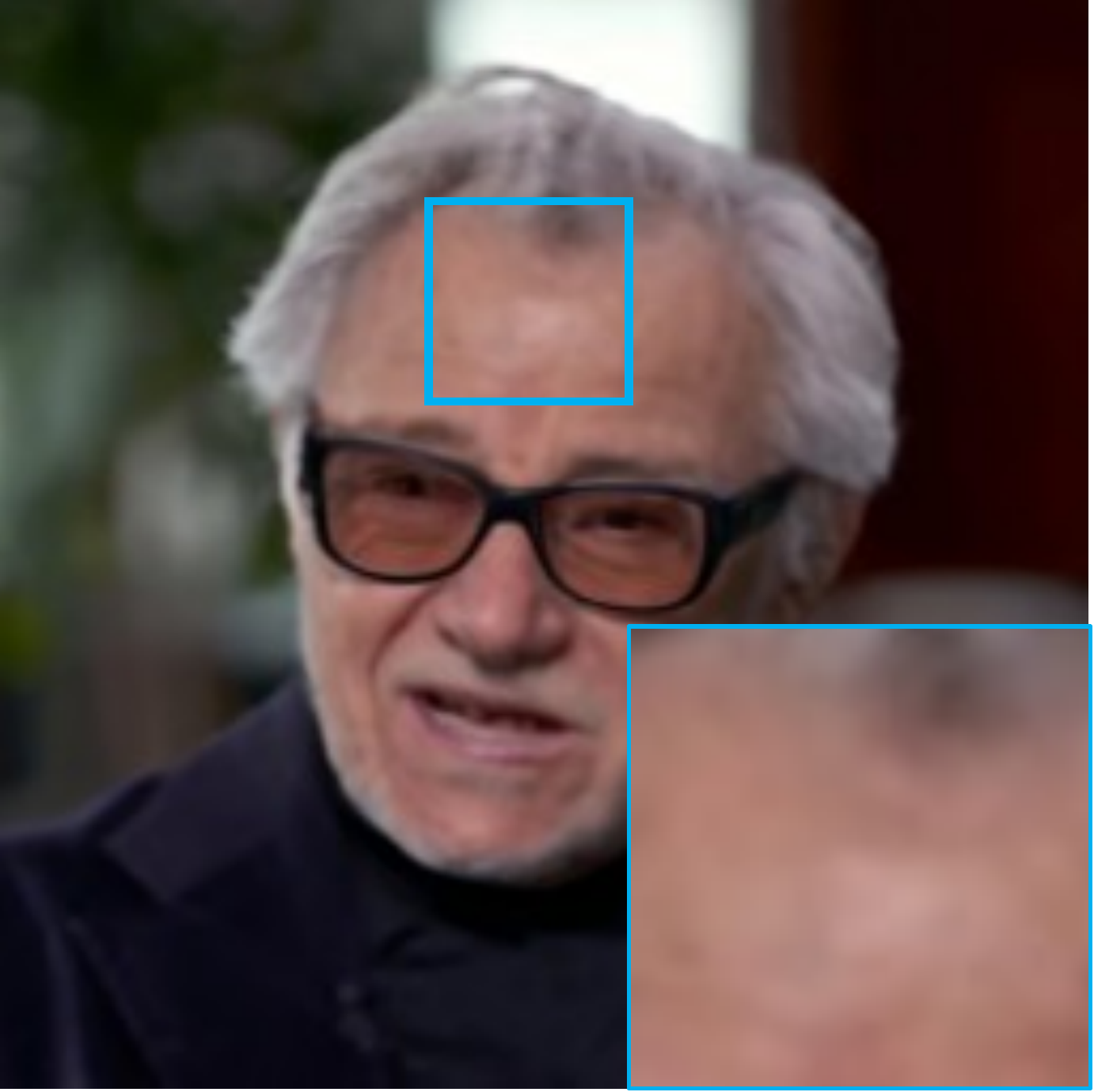}%
}
&
\raisebox{-.5\totalheight}{%
    \includegraphics[interpolate=false,width=0.185\textwidth]
    {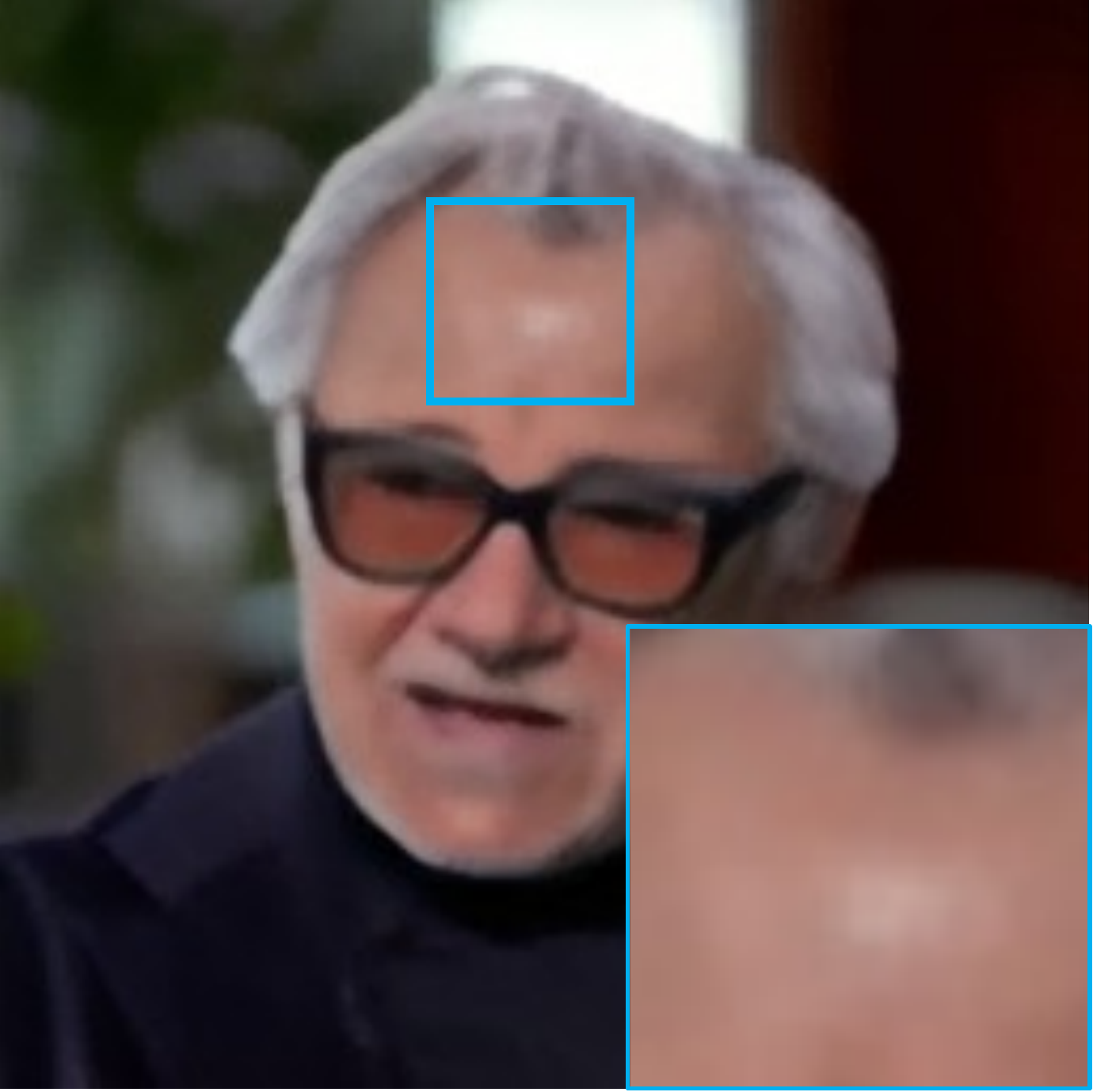}%
}
&
\raisebox{-.5\totalheight}{%
    \includegraphics[interpolate=false,width=0.185\textwidth]
    {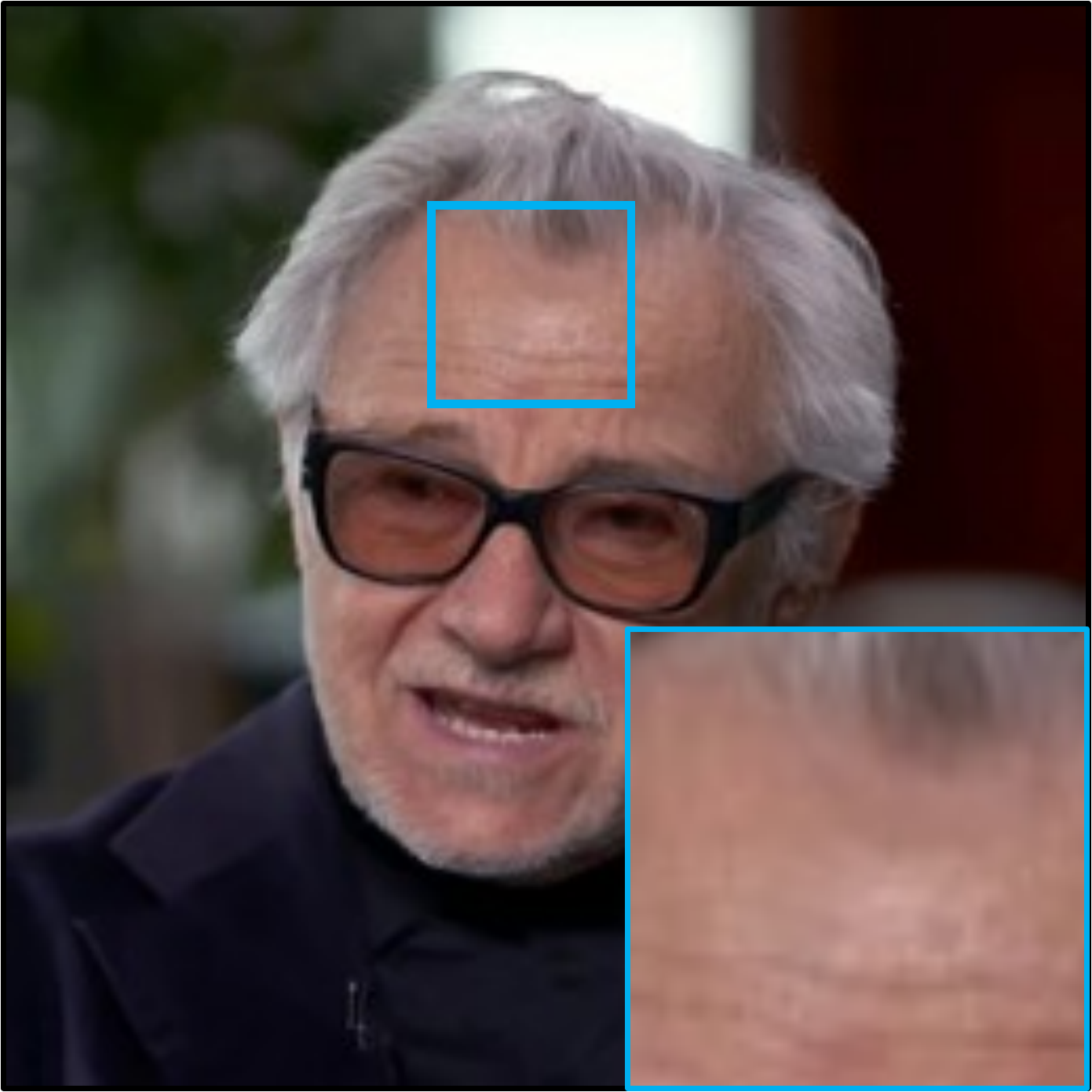}%
}
&
\raisebox{-.5\totalheight}{%
    \includegraphics[interpolate=false,width=0.185\textwidth]
    {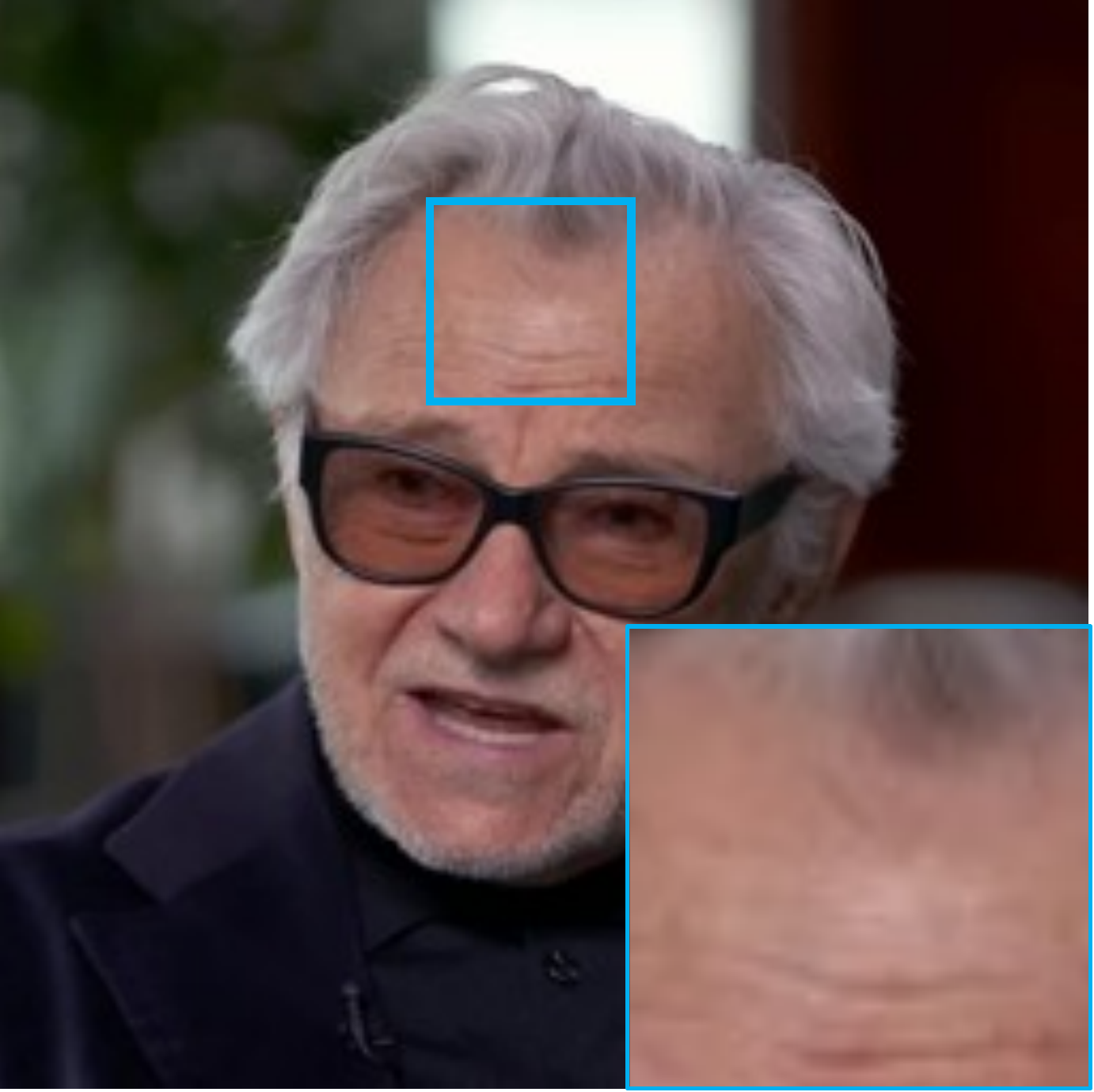}%
}
\\

\raisebox{-.5\totalheight}{%
    \makebox[0.077\textwidth][c]{%
        \shortstack{%
            \small\bfseries Noisy\\
            \small $\sigma_y=0.05$
        }%
    }%
}
&
\raisebox{-.5\totalheight}{%
    \includegraphics[interpolate=false,width=0.185\textwidth]
    {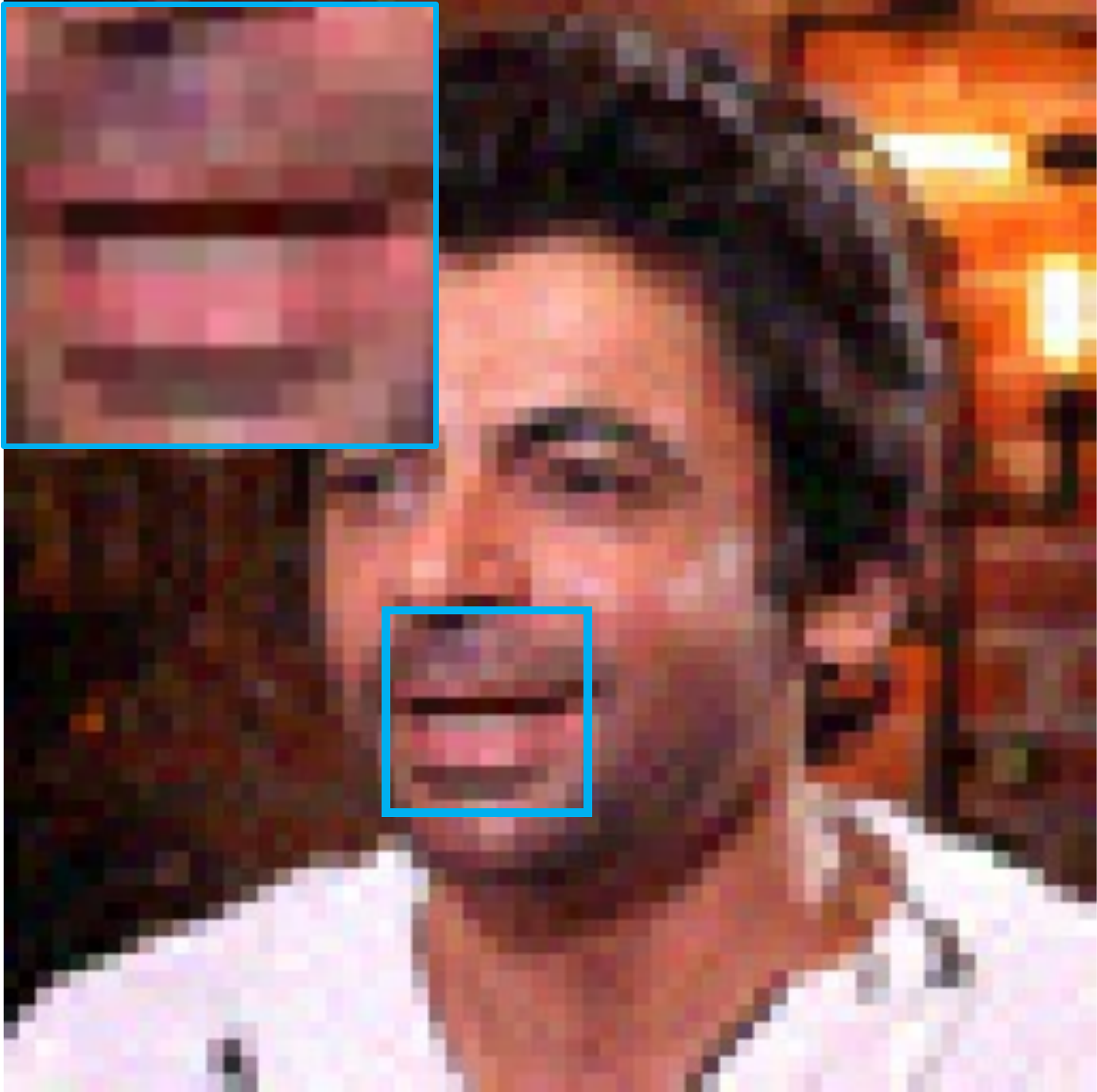}%
}
&
\raisebox{-.5\totalheight}{%
    \includegraphics[interpolate=false,width=0.185\textwidth]
    {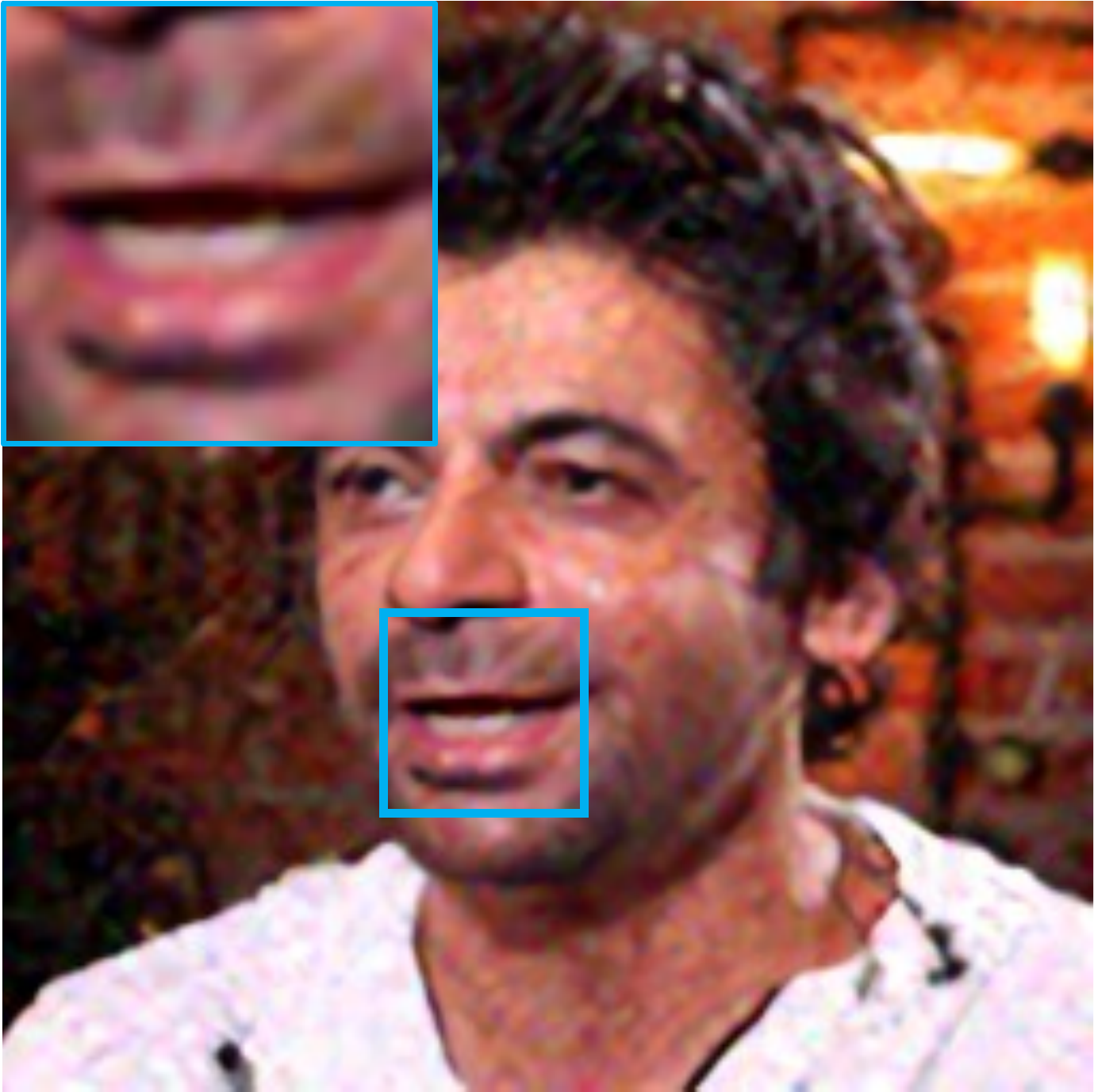}%
}
&
\raisebox{-.5\totalheight}{%
    \includegraphics[interpolate=false,width=0.185\textwidth]
    {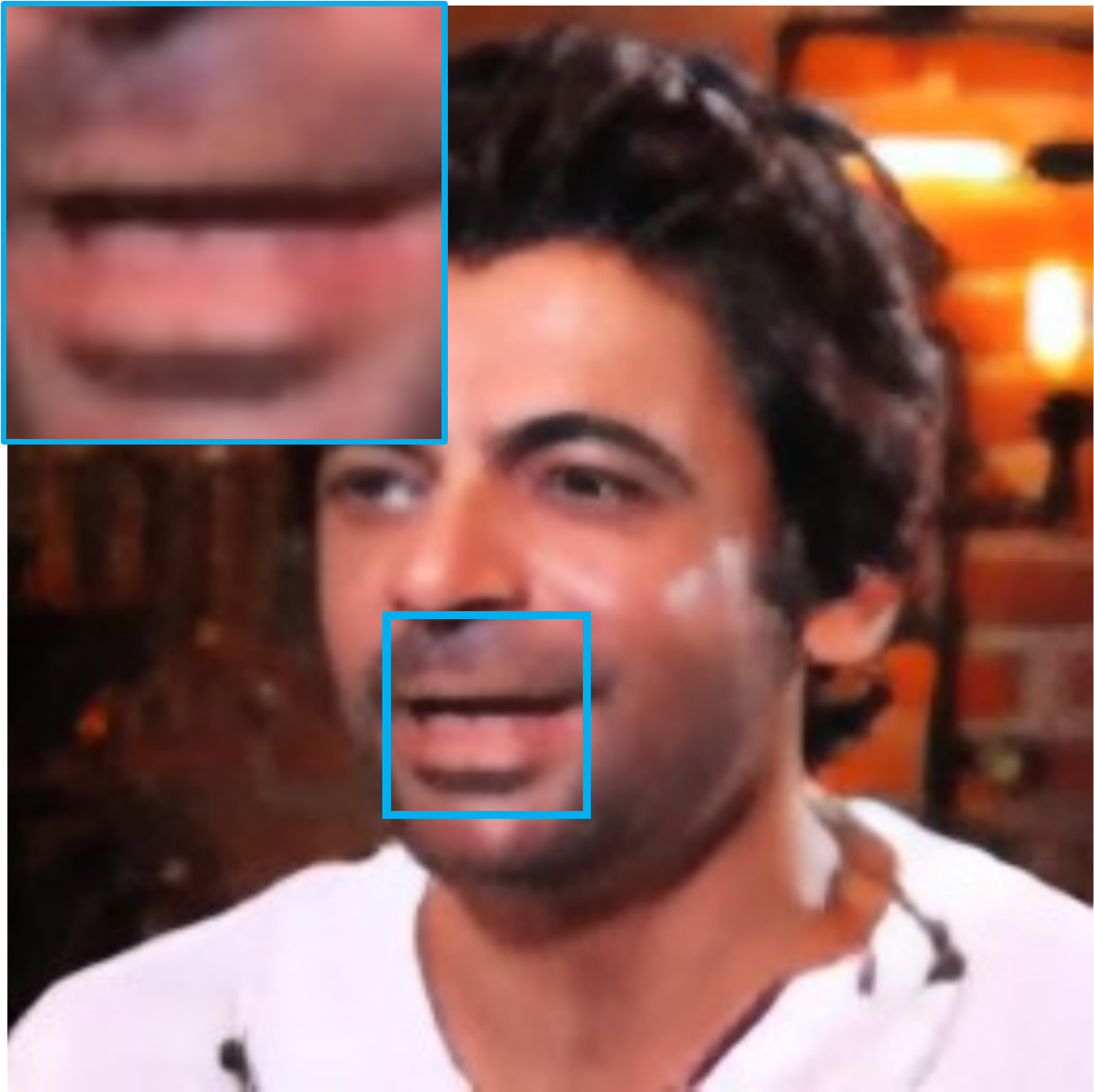}%
}
&
\raisebox{-.5\totalheight}{%
    \includegraphics[interpolate=false,width=0.185\textwidth]
    {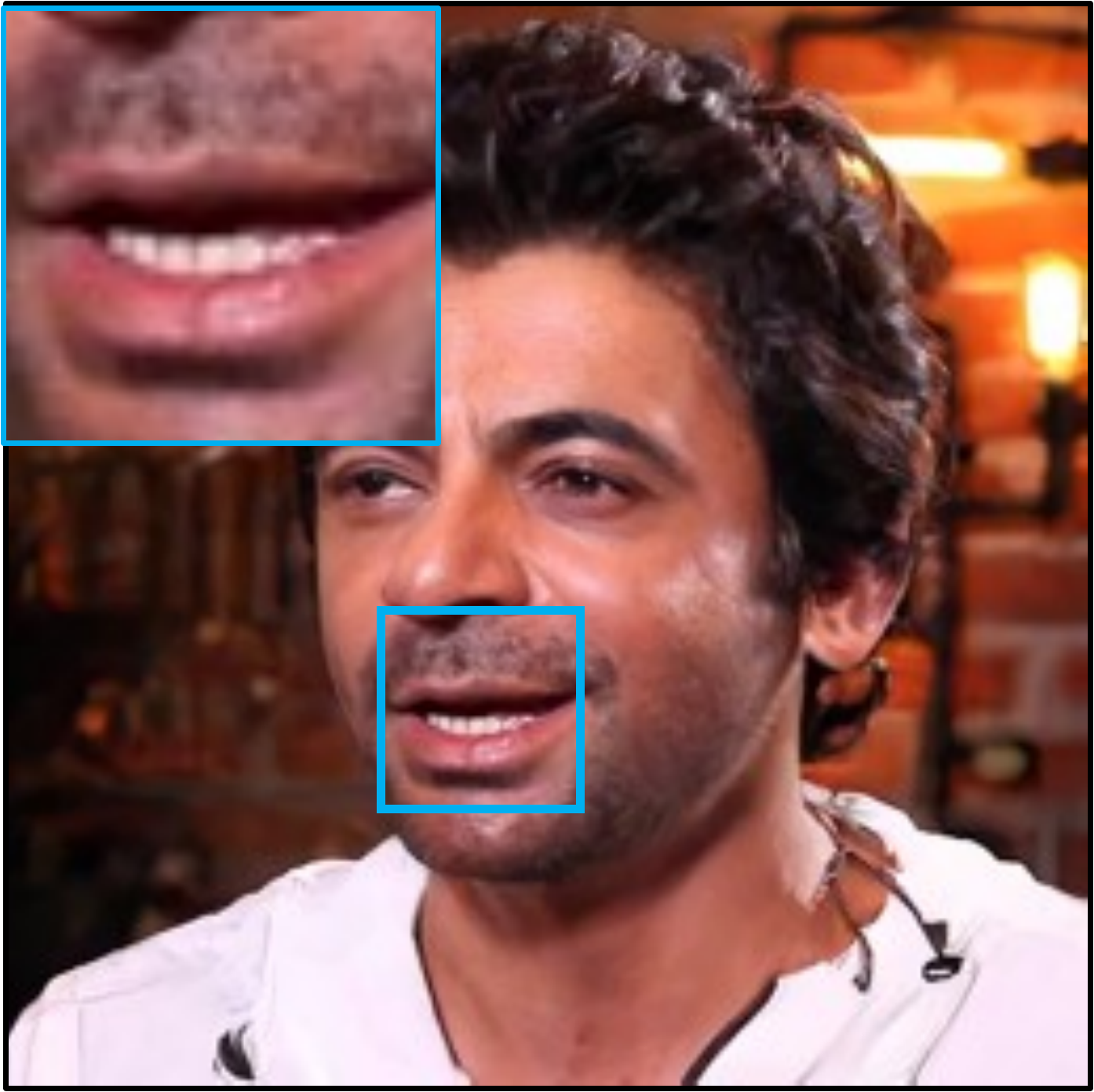}%
}
&
\raisebox{-.5\totalheight}{%
    \includegraphics[interpolate=false,width=0.185\textwidth]
    {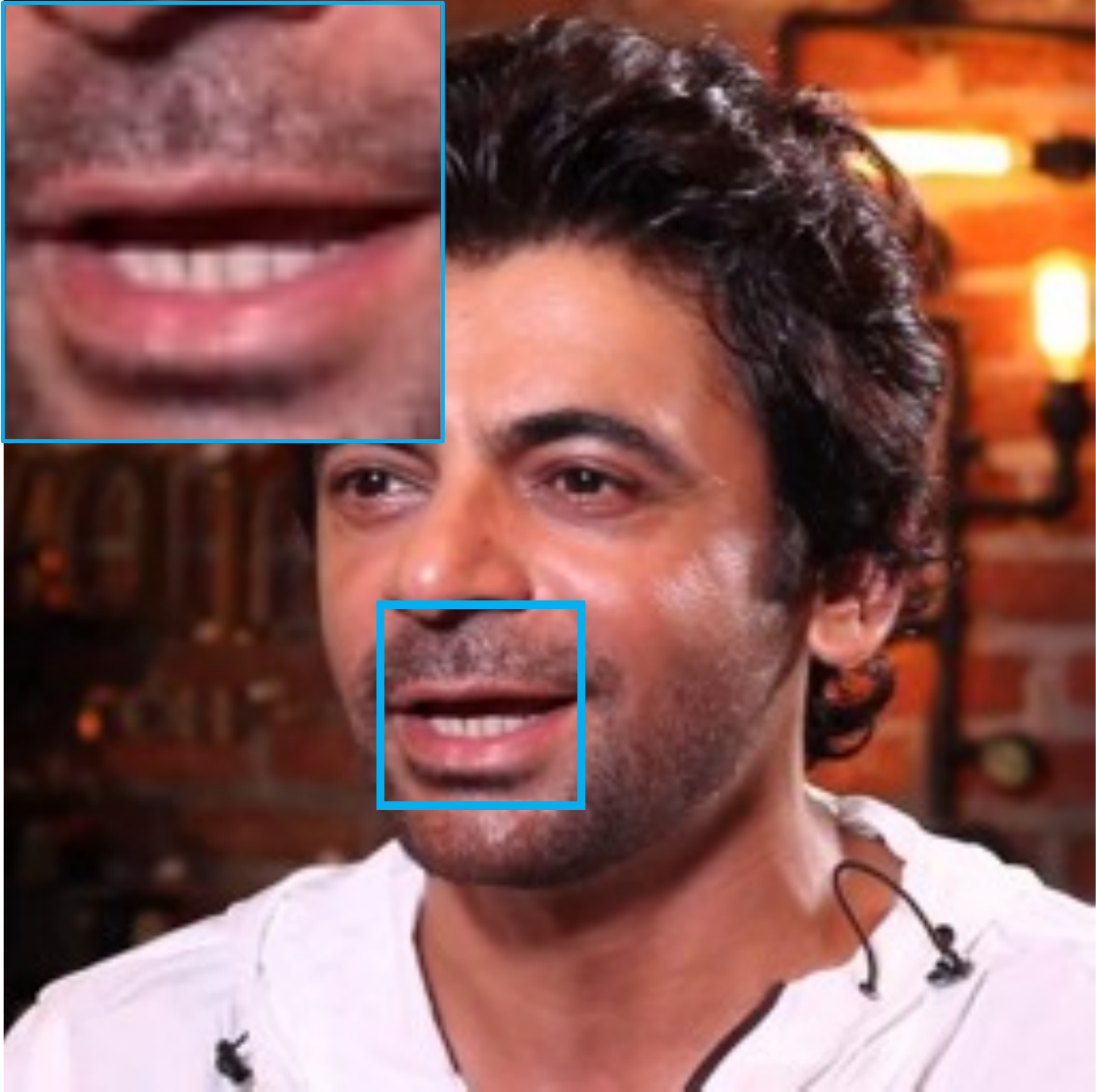}%
}
\end{tabular}

\caption{
\textbf{Qualitative comparison on VFHQ.}
Representative $4\times$ VSR results under noiseless bicubic degradation
($\sigma_y=0$, top) and additive Gaussian noise
($\sigma_y=0.05$, bottom).
Columns show the enlarged LR input, competing methods,
\method{}, and the ground truth.
Within each row, all methods reconstruct the same target frame, with the
corresponding regions enlarged for  comparison.
Across both settings, \method{}  faithfully recovers fine structures while avoiding over-smoothed or synthetic details.
Best viewed zoomed in.
}
\label{fig:qualitative}
\end{figure*}

\subsection{Experimental setup}
\label{sec:experiments_setup}
\textbf{Datasets.} We train and evaluate \method{} on VFHQ~\citep{gu2022vfhq}, a high-quality face video dataset. The training split comprises 13,867 video sequences from 5,956 distinct movies, totaling 3,175,749 frames. For main design choices and hyperparameter tuning, we randomly sample a validation split of 5  sequences from 5 distinct movies, containing 701 frames in total. The final test set consists of 100 sequences randomly sampled from 100 distinct movies, comprising 14,304 frames. All splits are constructed by random sampling while ensuring that they are mutually disjoint at the video level. Across the processed VFHQ sequences, frame rates range from 15 to 60 fps (mean 26.66, median 25), while the test split spans 24 to 60 fps (mean 26.42, median 25). At the median frame rate of 25 fps, the default temporal radius $k{=}5$ covers approximately $\pm0.20$\,s around the target frame. 

\textbf{Degradation Protocol.} We evaluate all methods under a unified degradation pipeline. The main evaluation in this section uses a $\times 4$ bicubic downsampling operator with anti-aliasing. We consider two distinct settings: a \textit{noiseless} setting where no additional noise is applied, and a \textit{noisy} setting where Additive White Gaussian Noise with a standard deviation of $\sigma_y = 0.05$ is added to the degraded frames, where the noise level is defined in the normalized image range $[-1,1]$. 
Additional evaluations for $\times 2$ and $\times 8$ scale-up under the same noiseless and noisy settings are reported in App.~\ref{app:additional_scales}.
Additional implementation details are in App.~\ref{app:experimental_details}.

\textbf{Baselines.} We compare \method{} with recent DM-based methods for VSR: SVI-Diffusion~\citep{kwon2025solving}, VISION-XL~\citep{visionxl2025}, Upscale-A-Video~\citep{zhou2024upscale}, UltraVSR~\citep{ultravsr2024}, StableVSR~\citep{stablevsr2024}, and PS-SR~\citep{wu2026pssr}. All methods are evaluated on identical degraded inputs under the same degradation and metric protocols using their official implementations and released checkpoints, which support our domain. The competing methods are evaluated in their released operating regimes and rely on large-scale pretrained diffusion priors: the VSR baselines use their pretrained VSR models, while the inverse-problem solvers use their pretrained image priors; for SVI-Diffusion, we use an FFHQ prior matching the facial domain. \method{} is fine-tuned only on the VFHQ training split. Method-specific inference configurations are selected on the validation split and fixed before test set evaluation, including an autograd-defined adjoint for baselines with CG updates and a BM3D-preprocessed StableVSR variant for noisy SR. Method implementation details are provided in App.~\ref{app:experimental_details}.
We also compare with the non-DM-based BasicVSR++~\citep{basicvsrpp} in App.~\ref{app:expanded_main_test}.

\textbf{Metrics.} We evaluate complementary aspects of video reconstruction quality. Distortion level fidelity is measured using PSNR and SSIM~\citep{wang2004ssim}, perceptual similarity is assessed with LPIPS~\citep{zhang2018perceptual}, and video level perceptual quality, reflecting both spatial appearance and temporal dynamics, is evaluated using FVD~\citep{unterthiner2019fvd}. For FVD, we pool non-overlapping 16-frame clips from all test videos and compute a single dataset level score. All methods are evaluated with the same public video quality evaluation toolkit\footnote{\href{https://github.com/JunyaoHu/common_metrics_on_video_quality}{common\_metrics\_on\_video\_quality repository}} under an identical metric pipeline. Higher PSNR and SSIM indicate better reconstruction fidelity, while lower LPIPS and FVD indicate better perceptual and video level quality.

\begin{table*}[t]
\centering
\footnotesize
\caption{
Main quantitative comparison of DM-based methods on the held-out VFHQ test set
(100 sequences, 14{,}304 frames) for $\times4$ face VSR under noisy and noiseless bicubic degradation.
Best results are shown in \textbf{bold} and second-best results are \underline{underlined}.
}
\label{tab:main-test}
\begin{tabular}{l cccc c cccc}
\toprule
& \multicolumn{4}{c}{\textbf{Noisy} ($\sigma_y=0.05$)} & & \multicolumn{4}{c}{\textbf{Noiseless}} \\
\cmidrule(lr){2-5}\cmidrule(lr){7-10}
Method & PSNR$\uparrow$ & SSIM$\uparrow$ & LPIPS$\downarrow$ & FVD$\downarrow$ & & PSNR$\uparrow$ & SSIM$\uparrow$ & LPIPS$\downarrow$ & FVD$\downarrow$ \\
\midrule
\method{}        & \textbf{30.026} & \textbf{0.862} & \textbf{0.063} & \textbf{17.476} & & \textbf{31.778} & \textbf{0.905} & \textbf{0.043} & \textbf{8.526} \\
SVI-Diffusion           & 26.489 & 0.616 & 0.431 & 865.181 & & \underline{31.032} & \underline{0.893} & 0.099 & 30.309 \\
Upscale-A-Video         & \underline{28.970} & \underline{0.838} & 0.154 & 194.191 & & 29.223 & 0.856 & 0.149 & 139.189 \\
VISION-XL               & 26.993 & 0.701 & 0.421 & 875.044 & & 29.473 & 0.850 & 0.197 & 94.496 \\
UltraVSR                & 27.440 & 0.823 & \underline{0.123} & 198.494 & & 27.853 & 0.838 & 0.110 & 93.860 \\
PS-SR                   & 23.658 & 0.792 & 0.166 & \underline{174.431} & & 23.619 & 0.788 & 0.166 & 174.811 \\
StableVSR               & 25.444 & 0.527 & 0.432 & 1159.674 & & 30.471 & 0.878 & \underline{0.068} & \underline{25.498} \\
StableVSR+BM3D          & 28.864 & 0.821 & 0.160 & 231.221 & & -- & -- & -- & -- \\
\bottomrule
\end{tabular}
\end{table*}

\textbf{Implementation details.} \method{} uses the OpenAI \texttt{guided-diffusion} training framework~\citep{dhariwal2021diffusion} and DPS codebase~\citep{chung2023diffusion}. The pixel-space DDPM U-Net operates at $256{\times}256$ RGB resolution, is initialized from the FFHQ checkpoint used in DPS~\citep{chung2023diffusion}, and is fine-tuned on VFHQ. Only the first convolution is widened for additional conditioning channels, with weights initialized by replicating pretrained RGB filters and fixed channel ordering across training and inference. Conditioning, central frame handling, mirror padding, and shared noise sampling follow Sec.~\ref{sec:method}; inference uses the same degradation operator as evaluation protocol. Guidance step size is selected on validation split: $\zeta'=5.2$ for noiseless and $\zeta'=1.2$ for noisy reconstruction. Full architecture, training, checkpoint selection, and sampling details appear in App.~\ref{app:training_details}; test set runtime and peak memory for all methods appear in App.~\ref{app:efficiency}. Code and checkpoints will be released upon publication.

\subsection{Main results}
\label{sec:main_results}
Tab.~\ref{tab:main-test} compares recent DM-based methods for $\times4$ face VSR on the held-out VFHQ test set across noisy and noiseless bicubic degradation settings. Under the unified degradation and evaluation protocol described above, \method{} achieves the best results across all reported metrics among the evaluated DM-based methods in both degradation regimes. At the frame level, \method{} attains the highest PSNR, improving over the leading competing result by approximately $1.1$ dB in the noisy setting and by around $0.75$ dB in the noiseless setting. At the video level, \method{} also achieves substantially lower FVD, achieving almost an order of magnitude reduction under noise and about a threefold reduction in the noiseless setting. Together with the best SSIM and LPIPS scores, these results indicate that \method{} improves reconstruction fidelity, perceptual similarity, and temporal coherence simultaneously, without trading measurement fidelity for perceptual quality.
We evaluate \method{} at $\times2$ and $\times8$ scale factors,
with quantitative results in App.~\ref{app:additional_scales}.

Qualitative comparisons in Fig.~\ref{fig:qualitative} corroborate the quantitative results. 
\method{} yields sharper and more faithful facial reconstructions, preserving fine structures around the eyes and mouth, as well as fine details such as teeth, wrinkles, and hair. 
Competing VSR methods more frequently exhibit oversmoothed facial regions, unnatural local textures, or reduced temporal stability, especially under additive noise. 
More visual comparisons appear in App.~\ref{app:additional_visuals}.

\subsection{Ablation studies}
\label{sec:ablations}
We ablate the main components of \method{} on the VFHQ validation split under the main $\times4$ setting. Unless ablating stochasticity itself, we use matched sampling noise trajectories across variants to control for sampling randomness. Unless stated otherwise, each configuration is independently fine-tuned on VFHQ, and its best checkpoint is selected based on validation PSNR; $\zeta'$ is then tuned independently under the same model-selection and evaluation protocol.

\begin{table}[t]
\centering
\caption{
\textbf{Validation ablations for $\times4$ VSR}.
Architecture variants are reported using their best validation PSNR checkpoint and optimal guidance scale $\zeta'$; inference ablations use the selected
architecture and guidance settings. Boldface marks the selected configurations. Full sweeps and metrics appear in App.~\ref{app:extra_ablations}.
}
\label{tab:compact_ablations}

\footnotesize
\setlength{\tabcolsep}{2.2pt}
\renewcommand{\arraystretch}{1.06}
\sisetup{detect-weight=true,detect-family=true}

\begin{tabular*}{0.97\columnwidth}{
@{\extracolsep{\fill}}
l
S[table-format=2.1]
S[table-format=2.3]
S[table-format=3.3]
S[table-format=2.1]
S[table-format=2.3]
S[table-format=3.3]
@{}
}
\toprule
&
\multicolumn{3}{c}{Noiseless} &
\multicolumn{3}{c}{Noisy} \\
\cmidrule(lr){2-4}
\cmidrule(l){5-7}
{Variant} &
{$\zeta'$} &
{PSNR$\uparrow$} &
{FVD$\downarrow$} &
{$\zeta'$} &
{PSNR$\uparrow$} &
{FVD$\downarrow$} \\
\midrule

\multicolumn{7}{@{}l}{
\textit{Temporal context}
{\normalfont\,(without central LR denoiser conditioning)}
} \\
\quad $k=0$
    & 1.3
    & 27.469
    & 161.338
    & 1.2
    & 26.700
    & 446.632 \\
\quad $k=2$
    & 5.3
    & 29.354
    & 69.429
    & 1.2
    & 27.750
    & 152.476 \\
\quad $k=5$
    & \bfseries 5.0
    & \bfseries 29.568
    & 72.177
    & \bfseries 1.2
    & \bfseries 28.254
    & 130.086 \\
\quad $k=7$
    & 5.5
    & 29.479
    & 72.258
    & 1.2
    & 28.222
    & 118.395 \\
\quad $k=10$
    & 5.4
    & 29.408
    & 69.126
    & 1.2
    & 27.756
    & 120.337 \\

\midrule

\multicolumn{7}{@{}l}{
\textit{Central LR denoiser conditioning}
{\normalfont\,($k=5$)}
} \\
\quad Excluded
    & 5.0
    & 29.568
    & 72.177
    & \bfseries 1.2
    & \bfseries 28.254
    & 130.086 \\
\quad Included
    & \bfseries 5.2
    & \bfseries 29.694
    & 68.720
    & 0.4
    & 28.198
    & 125.455 \\

\midrule

\multicolumn{7}{@{}l}{
\textit{Data-consistency guidance scale}
{\normalfont\,(selected architecture)}
} \\
\quad Off
    & 0.0
    & 29.424
    & 69.709
    & 0.0
    & 27.530
    & 144.390 \\
\quad Low
    & 2.0
    & 29.594
    & 68.623
    & 0.4
    & 28.133
    & 140.676 \\
\quad Moderate
    & 4.0
    & 29.679
    & 68.351
    & 0.8
    & 28.232
    & 134.818 \\
\quad Selected
    & \bfseries 5.2
    & \bfseries 29.694
    & 68.720
    & \bfseries 1.2
    & \bfseries 28.254
    & 130.086 \\
\quad High
    & 7.0
    & 29.684
    & 72.399
    & 1.6
    & 28.233
    & 133.309 \\
\quad Very high
    & 10.0
    & 29.645
    & 78.273
    & 2.0
    & 28.181
    & 151.374 \\

\midrule

\multicolumn{7}{@{}l}{
\textit{Boundary handling}
{\normalfont\,(selected architecture and guidance)}
} \\
\quad Zero
    & 5.2
    & 29.685
    & 68.766
    & 1.2
    & 28.242
    & 130.520 \\
\quad Mirroring
    & \bfseries 5.2
    & \bfseries 29.694
    & 68.720
    & \bfseries 1.2
    & \bfseries 28.254
    & 130.086 \\
    
\midrule

\multicolumn{7}{@{}l}{
\textit{Shared noise trajectory}
{\normalfont\,(selected architecture and guidance)}
} \\
\quad Excluded
    & 5.2
    & 29.702
    & 82.795
    & 1.2
    & 28.240
    & 164.753 \\
\quad Included
    & \bfseries 5.2
    & 29.694
    & \bfseries 68.720
    & \bfseries 1.2
    & 28.254
    & \bfseries 130.086 \\
    
\bottomrule
\end{tabular*}
\end{table}

\textbf{Temporal context.}
Tab.~\ref{tab:compact_ablations} examines the temporal conditioning radius
$k\in\{0,2,5,7,10\}$. Ablating temporal context ($k{=}0$) substantially degrades performance in both settings. Compared with $k{=}0$, the selected temporal conditioning ($k{=}5$) improves PSNR by approximately $2.10$ dB in the noiseless setting and $1.55$ dB under noise, while also improving FVD by approximately $55\%$ and $71\%$, respectively. This confirms that the denoiser effectively leverages information from neighboring frames. Expanding the context from $\pm2$ to $\pm5$, excluding the central LR observation, consistently enhances PSNR in both degradation settings. This indicates that additional nearby LR observations offer complementary evidence for detail recovery. Reconstruction fidelity peaks at $k{=}5$ and decreases for wider windows, indicating that distant frames are not merely redundant: as temporal offsets increase, motion, expression changes, and occlusions reduce their relevance and can interfere with target-frame reconstruction. Although broader context may reduce FVD, these gains come at the expense of PSNR, exposing a trade-off between
sequence-level temporal quality and frame-level fidelity. We explored adaptive temporal window sizes, including varying the denoiser context size $k$ during sampling according to local motion, as well as switching between denoisers during sampling; both introduced temporal inconsistencies. Conditioning on the previously reconstructed HR frame led to cumulative error propagation and provided no consistent benefit over the fixed $k{=}5$ LR conditioning scheme. We therefore adopt $k{=}5$, as the optimal balance between
useful temporal support and accurate detail recovery.

\textbf{Central LR observation.}
As described in Sec.~\ref{sec:method_cond}, the role of the central LR
observation depends on measurement reliability.
Tab.~\ref{tab:compact_ablations} shows that including it improves noiseless
PSNR from $29.568$ to $29.694$ dB, whereas excluding it yields the best
noisy PSNR ($28.254$ dB). We therefore include the central LR observation
only in the noiseless denoiser; in both settings, the central measurement
remains enforced through DPS. Full metrics and qualitative analysis are
provided in App.~\ref{app:full_temporal_ablation} and
App.~\ref{app:visual_ablations}.

\textbf{Data-consistency guidance scale.}
Tab.~\ref{tab:compact_ablations} isolates the contribution of the DPS likelihood guidance. Setting $\zeta'=0$ and sampling from the conditioned prior alone, reduces PSNR by $0.27$ dB in the noiseless setting and $0.72$ dB under noise, demonstrating that explicit measurement consistency contributes beyond temporal conditioning. $\zeta'$ is tuned per setting on the validation split, peaking at $\zeta'=5.2$ in the noiseless setting and
at $\zeta'=1.2$ in the noisy setting; the smaller optimal scale under noise reflects the weaker reliability of the noisy measurement.

\textbf{Boundary handling.}
At video sequence boundaries, where a complete temporal neighborhood is unavailable, we mirror the valid observations rather than introduce zero-valued conditioning frames. Tab.~\ref{tab:compact_ablations} shows that the aggregate gain is modest, it is consistent and incurs no additional computational cost; we therefore use mirror padding. 

\textbf{Shared noise trajectory.}
Independent per-frame diffusion stochasticity produces a visible flicker. Sharing the diffusion noise trajectory leaves frame level fidelity essentially unchanged, while substantially improving video level consistency: FVD decreases from 82.795 to 68.720 in the noiseless setting and from 164.753 to 130.086 under noise, with negligible changes in PSNR. This confirms that shared
stochasticity primarily acts as a temporal stabilization mechanism rather
than improving individual-frame reconstruction.

\section{Conclusion}
\label{sec:conclusion}
We introduced \method{}, a pixel-space diffusion framework for high-fidelity face VSR integrating localized spatio-temporal conditioning, per-frame DPS guidance, and a shared noise trajectory. These components are complementary: neighboring LR frames provide temporal evidence, DPS anchors each target frame to its observation, and shared noise trajectory maintains a coherent stochastic evolution across the entire sequence.
Under a unified evaluation protocol on VFHQ, \method{} achieves the best performance among the evaluated DM-based methods across all reported metrics in noisy and noiseless $\times4$ settings. Relative to the strongest competitor, it improves PSNR by $1.06$ dB under noise and $0.75$ dB without noise, while reducing FVD
from $174.431$ to $17.476$ and from $25.498$ to $8.526$, respectively.
Experiments at $\times2$ and $\times8$ demonstrate \method{}'s applicability across different scale factors.
Ablations identify $k{=}5$ as the best balance between useful temporal
support and target-frame fidelity, and show that the guidance scale $\zeta'$ is critical for balancing the conditioned prior with measurement consistency.
\method{} requires neither optical flow nor latent compression, supports parallel frame-wise inference with no architectural dependence on sequence length, and avoids autoregressive error accumulation.

\paragraph{Limitations and future work.}
Our study focuses on face VSR under controlled, known bicubic degradations and evaluates two measurement noise regimes, $\sigma_y{=}0$ and $\sigma_y{=}0.05$. This controlled setting allows us to isolate the effects of temporal conditioning and posterior guidance but does not capture the full complexity of real-world video degradations, such as unknown blur and compression. The diffusion sampler remains computationally expensive. Future work will explore accelerated sampling, broader and blind degradation models, and extension beyond the face domain.

\paragraph{Ethical considerations.}
As with other generative restoration methods, \method{} may reconstruct plausible facial details that are not uniquely determined by the LR observations. Its outputs should therefore not be treated as ground truth in forensic or other high stakes identification settings.

{
\small
\bibliographystyle{ieeenat_fullname}
\bibliography{paper}

@article{baniya2024vsrsurvey,
  author  = {Baniya, Arbind Agrahari and Lee, Tsz-Kwan and Eklund, Peter W. and Aryal, Sunil},
  title   = {A Survey of Deep Learning Video Super-Resolution},
  journal = {IEEE Transactions on Emerging Topics in Computational Intelligence},
  year    = {2024},
  volume  = {8},
  number  = {4},
  pages   = {2655--2676},
  month   = aug,
  doi     = {10.1109/TETCI.2024.3398015}
}

@inproceedings{gu2022vfhq,
  author    = {Xie, Liangbin and Wang, Xintao and Zhang, Honglun and Dong, Chao and Shan, Ying},
  title     = {{VFHQ}: A High-Quality Dataset and Benchmark for Video Face Super-Resolution},
  booktitle = {Proceedings of the IEEE/CVF Conference on Computer Vision and Pattern Recognition (CVPR) Workshops},
  month     = jun,
  year      = {2022},
  pages     = {657--666}
}

@inproceedings{zou2025flair,
  author    = {Zou, Zihao and Liu, Jiaming and Shoushtari, Shirin and Wang, Yubo and Kamilov, Ulugbek S.},
  title     = {{FLAIR}: A Conditional Diffusion Framework with Applications to Face Video Restoration},
  booktitle = {Proceedings of the Winter Conference on Applications of Computer Vision (WACV)},
  month     = feb,
  year      = {2025},
  pages     = {5228--5238}
}

@inproceedings{ho2020denoising,
  author    = {Ho, Jonathan and Jain, Ajay and Abbeel, Pieter},
  title     = {Denoising Diffusion Probabilistic Models},
  booktitle = {Advances in Neural Information Processing Systems},
  volume    = {33},
  pages     = {6840--6851},
  year      = {2020}
}

@inproceedings{dhariwal2021diffusion,
  author    = {Dhariwal, Prafulla and Nichol, Alexander},
  title     = {Diffusion Models Beat {GANs} on Image Synthesis},
  booktitle = {Advances in Neural Information Processing Systems},
  volume    = {34},
  pages     = {8780--8794},
  year      = {2021}
}

@inproceedings{kawar2022denoising,
  author    = {Kawar, Bahjat and Elad, Michael and Ermon, Stefano and Song, Jiaming},
  title     = {Denoising Diffusion Restoration Models},
  booktitle = {Advances in Neural Information Processing Systems},
  volume    = {35},
  pages     = {23593--23606},
  year      = {2022}
}

@inproceedings{wang2022zero,
  title     = {Zero-Shot Image Restoration Using Denoising Diffusion Null-Space Model},
  author    = {Wang, Yinhuai and Yu, Jiwen and Zhang, Jian},
  booktitle = {The Eleventh International Conference on Learning Representations},
  year      = {2023},
}

@inproceedings{chung2023diffusion,
  title     = {Diffusion Posterior Sampling for General Noisy Inverse Problems},
  author    = {Chung, Hyungjin and Kim, Jeongsol and McCann, Michael Thompson and Klasky, Marc Louis and Ye, Jong Chul},
  booktitle = {The Eleventh International Conference on Learning Representations},
  year      = {2023},
}

@inproceedings{cao2025zero,
  title     = {Zero-Shot Video Restoration and Enhancement Using Pre-Trained Image Diffusion Model},
  author    = {Cao, Cong and Yue, Huanjing and Liu, Xin and Yang, Jingyu},
  booktitle = {Proceedings of the AAAI Conference on Artificial Intelligence},
  volume    = {39},

  pages     = {1935--1943},
  year      = {2025}
}

@article{yeh2024diffir2vr,
  title         = {{DiffIR2VR-Zero}: Zero-Shot Video Restoration with Diffusion-based Image Restoration Models},
  author        = {Yeh, Chang-Han and Lin, Chin-Yang and Wang, Zhixiang and Hsiao, Chi-Wei and Chen, Ting-Hsuan and Shiu, Hau-Shiang and Liu, Yu-Lun},
  journal       = {arXiv preprint arXiv:2407.01519},
  year          = {2024},
  eprint        = {2407.01519},
  archivePrefix = {arXiv},
  primaryClass  = {cs.CV},
}

@inproceedings{zhou2024upscale,
  author    = {Zhou, Shangchen and Yang, Peiqing and Wang, Jianyi and Luo, Yihang and Loy, Chen Change},
  title     = {Upscale-A-Video: Temporal-Consistent Diffusion Model for Real-World Video Super-Resolution},
  booktitle = {Proceedings of the IEEE/CVF Conference on Computer Vision and Pattern Recognition (CVPR)},
  month     = jun,
  year      = {2024},
  pages     = {2535--2545}
}

@inproceedings{basicvsrpp,
  author    = {Chan, Kelvin C. K. and Zhou, Shangchen and Xu, Xiangyu and Loy, Chen Change},
  title     = {{BasicVSR}++: Improving Video Super-Resolution with Enhanced Propagation and Alignment},
  booktitle = {Proceedings of the IEEE/CVF Conference on Computer Vision and Pattern Recognition (CVPR)},
  month     = jun,
  year      = {2022},
  pages     = {5972--5981}
}

@article{liang2022vrt,
  author  = {Liang, Jingyun and Cao, Jiezhang and Fan, Yuchen and Zhang, Kai and Ranjan, Rakesh and Li, Yawei and Timofte, Radu and Van Gool, Luc},
  title   = {{VRT}: A Video Restoration Transformer},
  journal = {IEEE Transactions on Image Processing},
  year    = {2024},
  volume  = {33},
  pages   = {2171--2182},
  doi     = {10.1109/TIP.2024.3372454}
}

@inproceedings{mgldvsr2024,
  author    = {Yang, Xi and He, Chenhang and Ma, Jianqi and Zhang, Lei},
  title     = {Motion-Guided Latent Diffusion for Temporally Consistent Real-World Video Super-Resolution},
  booktitle = {Computer Vision -- ECCV 2024},
  year      = {2024},
  pages     = {224--242},
  publisher = {Springer},
  doi       = {10.1007/978-3-031-72784-9_13}
}

@inproceedings{rombach2022highresolution,
  author    = {Rombach, Robin and Blattmann, Andreas and Lorenz, Dominik and Esser, Patrick and Ommer, Bj{\"o}rn},
  title     = {High-Resolution Image Synthesis with Latent Diffusion Models},
  booktitle = {Proceedings of the IEEE/CVF Conference on Computer Vision and Pattern Recognition (CVPR)},
  month     = jun,
  year      = {2022},
  pages     = {10684--10695}
}

@inproceedings{yi2025tvt,
  author    = {Yi, Qiaosi and Li, Shuai and Wu, Rongyuan and Sun, Lingchen and Wu, Yuhui and Zhang, Lei},
  title     = {Fine-structure Preserved Real-world Image Super-resolution via Transfer {VAE} Training},
  booktitle = {Proceedings of the IEEE/CVF International Conference on Computer Vision (ICCV)},
  month     = oct,
  year      = {2025},
  pages     = {12415--12426}
}

@inproceedings{kwon2025solving,
  title     = {Solving Video Inverse Problems Using Image Diffusion Models},
  author    = {Kwon, Taesung and Ye, Jong Chul},
  booktitle = {The Thirteenth International Conference on Learning Representations},
  year      = {2025},
}

@inproceedings{chan2022tradeoffs,
  author    = {Chan, Kelvin C. K. and Zhou, Shangchen and Xu, Xiangyu and Loy, Chen Change},
  title     = {Investigating Tradeoffs in Real-World Video Super-Resolution},
  booktitle = {Proceedings of the IEEE/CVF Conference on Computer Vision and Pattern Recognition (CVPR)},
  month     = jun,
  year      = {2022},
  pages     = {5962--5971}
}

@inproceedings{liang2022rvrt,
  author    = {Liang, Jingyun and Fan, Yuchen and Xiang, Xiaoyu and Ranjan, Rakesh and Ilg, Eddy and Green, Simon and Cao, Jiezhang and Zhang, Kai and Timofte, Radu and Van Gool, Luc},
  title     = {{RVRT}: Recurrent Video Restoration Transformer with Guided Deformable Attention},
  booktitle = {Advances in Neural Information Processing Systems},
  volume    = {35},
  pages     = {378--393},
  year      = {2022}
}

@inproceedings{zhang2024realviformer,
  author    = {Zhang, Yuehan and Yao, Angela},
  title     = {{RealViformer}: Investigating Attention for Real-World Video Super-Resolution},
  booktitle = {Computer Vision -- ECCV 2024},
  year      = {2024},
  pages     = {412--428},
  publisher = {Springer},
  doi       = {10.1007/978-3-031-73397-0_24}
}

@inproceedings{basicvsr,
  author    = {Chan, Kelvin C. K. and Wang, Xintao and Yu, Ke and Dong, Chao and Loy, Chen Change},
  title     = {{BasicVSR}: The Search for Essential Components in Video Super-Resolution and Beyond},
  booktitle = {Proceedings of the IEEE/CVF Conference on Computer Vision and Pattern Recognition (CVPR)},
  month     = jun,
  year      = {2021},
  pages     = {4947--4956}
}

@inproceedings{edvr,
  author    = {Wang, Xintao and Chan, Kelvin C. K. and Yu, Ke and Dong, Chao and Loy, Chen Change},
  title     = {{EDVR}: Video Restoration With Enhanced Deformable Convolutional Networks},
  booktitle = {Proceedings of the IEEE/CVF Conference on Computer Vision and Pattern Recognition (CVPR) Workshops},
  month     = jun,
  year      = {2019}
}

@inproceedings{blau2018perception,
  author    = {Blau, Yochai and Michaeli, Tomer},
  title     = {The Perception-Distortion Tradeoff},
  booktitle = {Proceedings of the IEEE Conference on Computer Vision and Pattern Recognition (CVPR)},
  month     = jun,
  year      = {2018},
  pages     = {6228--6237}
}

@inproceedings{ledig2017srgan,
  author    = {Ledig, Christian and Theis, Lucas and Husz{\'a}r, Ferenc and Caballero, Jose and Cunningham, Andrew and Acosta, Alejandro and Aitken, Andrew and Tejani, Alykhan and Totz, Johannes and Wang, Zehan and Shi, Wenzhe},
  title     = {Photo-Realistic Single Image Super-Resolution Using a Generative Adversarial Network},
  booktitle = {Proceedings of the IEEE Conference on Computer Vision and Pattern Recognition (CVPR)},
  month     = jul,
  year      = {2017},
  pages     = {4681--4690}
}

@inproceedings{zhan2025vdmvsr,
  author    = {Zhan, Zhihao and Pang, Wang and Zhu, Xiang and Bai, Yechao},
  title     = {Rethinking Video Super-Resolution: Towards Diffusion-Based Methods without Motion Alignment},
  booktitle = {2025 17th International Conference on Signal Processing Systems (ICSPS)},
  year      = {2025},
  doi       = {10.1109/ICSPS66615.2025.11347706}
}

@inproceedings{visionxl2025,
  author    = {Kwon, Taesung and Ye, Jong Chul},
  title     = {{VISION-XL}: High Definition Video Inverse Problem Solver using Latent Image Diffusion Models},
  booktitle = {Proceedings of the IEEE/CVF International Conference on Computer Vision (ICCV)},
  year      = {2025},
  pages     = {10465--10474}
}

@inproceedings{podell2023sdxl,
  title     = {{SDXL}: Improving Latent Diffusion Models for High-Resolution Image Synthesis},
  author    = {Podell, Dustin and English, Zion and Lacey, Kyle and Blattmann, Andreas and Dockhorn, Tim and M{\"u}ller, Jonas and Penna, Joe and Rombach, Robin},
  booktitle = {The Twelfth International Conference on Learning Representations},
  year      = {2024},
}

@inproceedings{stablevsr2024,
  author    = {Rota, Claudio and Buzzelli, Marco and van de Weijer, Joost},
  title     = {Enhancing Perceptual Quality in Video Super-Resolution Through Temporally-Consistent Detail Synthesis Using Diffusion Models},
  booktitle = {Computer Vision -- ECCV 2024},
  year      = {2024},
  pages     = {36--53},
  publisher = {Springer},
  doi       = {10.1007/978-3-031-73254-6_3}
}

@inproceedings{ultravsr2024,
  author    = {Liu, Yong and Pan, Jinshan and Li, Yinchuan and Dong, Qingji and Zhu, Chao and Guo, Yu and Wang, Fei},
  title     = {{UltraVSR}: Achieving Ultra-Realistic Video Super-Resolution with Efficient One-Step Diffusion Space},
  booktitle = {Proceedings of the 33rd ACM International Conference on Multimedia},
  year      = {2025},
  pages     = {7785--7794},
  doi       = {10.1145/3746027.3755117}
}

@inproceedings{du2025patchvsr,
  author    = {Du, Shian and Xia, Menghan and Liu, Chang and Wang, Xintao and Wang, Jing and Wan, Pengfei and Zhang, Di and Ji, Xiangyang},
  title     = {{PatchVSR}: Breaking Video Diffusion Resolution Limits with Patch-wise Video Super-Resolution},
  booktitle = {Proceedings of the Computer Vision and Pattern Recognition Conference (CVPR)},
  month     = jun,
  year      = {2025},
  pages     = {17799--17809}
}

@article{wang2004ssim,
  author  = {Wang, Zhou and Bovik, Alan C. and Sheikh, Hamid R. and Simoncelli, Eero P.},
  title   = {Image Quality Assessment: From Error Visibility to Structural Similarity},
  journal = {IEEE Transactions on Image Processing},
  year    = {2004},
  volume  = {13},
  number  = {4},
  pages   = {600--612},
  month   = apr,
  doi     = {10.1109/TIP.2003.819861}
}

@inproceedings{zhang2018perceptual,
  author    = {Zhang, Richard and Isola, Phillip and Efros, Alexei A. and Shechtman, Eli and Wang, Oliver},
  title     = {The Unreasonable Effectiveness of Deep Features as a Perceptual Metric},
  booktitle = {Proceedings of the IEEE Conference on Computer Vision and Pattern Recognition (CVPR)},
  month     = jun,
  year      = {2018},
  pages     = {586--595}
}

@inproceedings{unterthiner2019fvd,
  title     = {{FVD}: A new Metric for Video Generation},
  author    = {Unterthiner, Thomas and van Steenkiste, Sjoerd and Kurach, Karol and Marinier, Raphael and Michalski, Marcin and Gelly, Sylvain},
  booktitle = {ICLR 2019 Workshop Deep Generative Models for Highly Structured Data},
  year      = {2019},
}

@inproceedings{karras2019ffhq,
  author    = {Karras, Tero and Laine, Samuli and Aila, Timo},
  title     = {A Style-Based Generator Architecture for Generative Adversarial Networks},
  booktitle = {Proceedings of the IEEE/CVF Conference on Computer Vision and Pattern Recognition (CVPR)},
  month     = jun,
  year      = {2019},
  pages     = {4401--4410}
}

@article{dabov2007bm3d,
  author  = {Dabov, Kostadin and Foi, Alessandro and Katkovnik, Vladimir and Egiazarian, Karen},
  title   = {Image Denoising by Sparse 3-D Transform-Domain Collaborative Filtering},
  journal = {IEEE Transactions on Image Processing},
  year    = {2007},
  volume  = {16},
  number  = {8},
  pages   = {2080--2095},
  month   = aug,
  doi     = {10.1109/TIP.2007.901238}
}

@inproceedings{rout2023solving,
  title     = {Solving Linear Inverse Problems Provably via Posterior Sampling with Latent Diffusion Models},
  author    = {Rout, Litu and Raoof, Negin and Daras, Giannis and Caramanis, Constantine and Dimakis, Alexandros G. and Shakkottai, Sanjay},
  booktitle = {Advances in Neural Information Processing Systems},
  volume    = {36},
  year      = {2023},
  pages     = {49960--49990}
}

@inproceedings{song2024resample,
  title     = {Solving Inverse Problems with Latent Diffusion Models via Hard Data Consistency},
  author    = {Song, Bowen and Kwon, Soo Min and Zhang, Zecheng and Hu, Xinyu and Qu, Qing and Shen, Liyue},
  booktitle = {The Twelfth International Conference on Learning Representations},
  year      = {2024}
}

@inproceedings{raphaeli2025silo,
  title     = {{SILO}: Solving Inverse Problems with Latent Operators},
  author    = {Raphaeli, Ron and Man, Sean and Elad, Michael},
  booktitle = {Proceedings of the IEEE/CVF International Conference on Computer Vision},
  year      = {2025},
  pages     = {10570--10580}
}

@inproceedings{wu2026pssr,
  author    = {Wu, Aiqiu and Qiu, Zhaofan and Yao, Ting and Mei, Tao},
  title     = {{PS-SR}: Pseudo-Single-Step Video Super-Resolution via Speculative Diffusion},
  booktitle = {Proceedings of the IEEE/CVF Conference on Computer Vision and Pattern Recognition (CVPR)},
  month     = jun,
  year      = {2026},
  pages     = {38218--38227}
}

@inproceedings{chen2025dove,
  title={DOVE: Efficient One-Step Diffusion Model for Real-World Video Super-Resolution},
  author={Chen, Zheng and Zou, Zichen and Zhang, Kewei and Su, Xiongfei and Yuan, Xin and Guo, Yong and Zhang, Yulun},
  booktitle={Advances in Neural Information Processing Systems},
  volume={38},
  pages={94905--94924},
  year={2025}
}
}

\clearpage
\maketitlesupplementary
\raggedbottom
\setlength{\multicolsep}{5pt plus 1pt minus 1pt}
\setlength{\premulticols}{0pt}
\setlength{\postmulticols}{0pt}
\makeatletter
\renewcommand\paragraph{\@startsection{paragraph}{4}{\z@}%
  {1.2ex plus .3ex minus .2ex}{-1em}{\normalfont\normalsize\bfseries}}
\makeatother
\newenvironment{suppfigure}{%
  \par\addvspace{6pt}\noindent\begin{minipage}{\textwidth}%
  \captionsetup{type=figure,skip=6pt}\centering
}{\end{minipage}\par\addvspace{6pt}}
\newenvironment{supptable}{%
  \par\addvspace{6pt}\noindent\begin{minipage}[b]{\textwidth}%
  \captionsetup{type=table,skip=6pt}\centering
}{\end{minipage}\par\addvspace{11.5pt}}

\appendix
\renewcommand{\theHsection}{appendix.\Alph{section}}
\renewcommand{\theHsubsection}{appendix.\Alph{section}.\arabic{subsection}}

\section{Additional experiments and implementation details}
\label{app:supplementary}

\subsection{Additional experimental details}
\label{app:experimental_details}

\paragraph{Evaluation protocol.}
We evaluate all methods under a unified degradation and metric pipeline. Given a clean HR video $x$, the LR measurement is generated as 
$y = \mathcal{A}(x) + n$,
where $\mathcal{A}$ denotes a $\times 2$, $\times 4$, or $\times 8$ bicubic downsampling operator with anti aliasing. We implement $\mathcal{A}$ using the public \texttt{Resizer} implementation\footnote{\url{https://github.com/assafshocher/resizer}} with \texttt{scale=0.5}, \texttt{scale=0.25}, or \texttt{scale=0.125}, respectively, \texttt{kernel="cubic"}, and \texttt{antialiasing=True}. These correspond to $128 \times 128 \rightarrow 256 \times 256$, $64 \times 64 \rightarrow 256 \times 256$, and $32 \times 32 \rightarrow 256 \times 256$ SR, respectively. For the noiseless setting we set $n=0$, while for the noisy setting we sample $n \sim \mathcal{N}(0,\sigma_y^2 I)$ with $\sigma_y=0.05$ and add it to the degraded LR frames, where the noise level is defined in the normalized image range $[-1,1]$. All reconstructed frames are saved as 16 bit PNG files. We report PSNR, SSIM, LPIPS, and FVD using the same public video quality evaluation toolkit\footnote{\url{https://github.com/JunyaoHu/common_metrics_on_video_quality/tree/main}} for all methods.

\paragraph{Baseline configuration.}
All baseline-specific inference adaptations described below are selected exclusively on the validation split and fixed before test evaluation; no test-set-specific tuning is performed. For SVI-Diffusion~\citep{kwon2025solving}, we use the FFHQ~\citep{karras2019ffhq} pretrained image diffusion prior released with DPS~\citep{chung2023diffusion}, matching the face domain setting of our experiments. The original implementation processes each video in non-overlapping chunks of 16 frames; we evaluate this protocol and additionally run a full video variant in which all frames are optimized jointly. Tab.~\ref{tab:main-test} reports the full-video variant. Since SVI-Diffusion solves a CG subproblem during sampling, the adjoint operator must be consistent with the forward degradation. We therefore compute $\mathcal{A}^{\top}$ using \texttt{torch.autograd}, rather than approximating it with an independent bicubic upsampling operator. Following the original SR configuration, we use 100 sampling steps, $\eta=0.8$, and $l=5$ inner optimization iterations. For the noisy SR setting, we use 3 inner optimization iterations instead of 5 and set the CG stopping tolerance to $\epsilon_{\mathrm{CG}}=10^{-3}$ instead of $10^{-5}$, which improves numerical stability under the noisy measurement model.

For VISION-XL~\citep{visionxl2025}, we follow the SDXL-based~\citep{podell2023sdxl} null text protocol used by the authors to avoid prompt induced content changes. We use 25 sampling steps, initialize the latent inversion at $\tau=0.3T$, and set the low pass filtering scale to $\lambda=2$. Since the released SR configuration assumes a different degradation implementation, directly applying the default CG setting to our cubic Resizer degradation led to unstable convergence. We therefore use the same autograd-defined adjoint $\mathcal{A}^{\top}$ as above and reduce the number of CG iterations to 7 rather than 10. For the noisy SR setting, we further reduce the number of CG iterations to 3 and set the CG stopping tolerance to $\epsilon_{\mathrm{CG}}=10^{-3}$ instead of $10^{-5}$. 

For Upscale A Video~\citep{zhou2024upscale}, UltraVSR~\citep{ultravsr2024}, StableVSR~\citep{stablevsr2024}, PS-SR~\citep{wu2026pssr}, and BasicVSR++~\citep{basicvsrpp}, which do not expose an explicit measurement operator at inference time, we provide the same degraded frames. For Upscale A Video, we disable text-based guidance by using empty positive and negative prompts, set the guidance scale to zero, enable the video VAE, use $N=20$ sampling steps, set the noise level to $s=50$, and use propagation steps $\{14,15,16,17\}$. StableVSR is evaluated with $T=50$ diffusion steps under the same input protocol. For the noisy task, we also evaluate StableVSR and BasicVSR++ variants with an initial BM3D~\citep{dabov2007bm3d} denoising stage applied to the noisy LR frames using $\sigma_y=0.05$. These adaptations are used to match our degradation model and ensure numerically stable inference. All adapted settings are selected on the validation split and fixed before test evaluation.

\subsection{Expanded main test results}
\label{app:expanded_main_test}

Tab.~\ref{tab:main-test-expanded} expands the main quantitative comparison in
Tab.~\ref{tab:main-test} by reporting result variability and including
BasicVSR++~\citep{basicvsrpp} as a representative non-DM baseline.
PSNR, SSIM, and LPIPS are reported as mean $\pm$
standard deviation. FVD is reported as a single video level score.

\paragraph{Comparison with the non DM baseline.}
The comparison with BasicVSR++~\citep{basicvsrpp} highlights a different operating point between
distortion and perceptual quality. In the noiseless setting,
BasicVSR++ achieves higher PSNR and SSIM than \method{}, whereas \method{}
achieves substantially better perceptual and video level scores. This pattern is
consistent with the perception distortion trade-off \citep{blau2018perception}:
methods optimized toward distortion-oriented fidelity can attain higher
PSNR/SSIM, while generative restoration methods may occupy a different point
on the trade-off and provide improved perceptual quality. Under noisy
degradation, LoCoVSR outperforms both BasicVSR++ variants across all four metrics.

\onecolumn

\begin{supptable}
\centering
\caption{
\textbf{Expanded quantitative comparison of DM-based and non-DM-based methods
on the held-out VFHQ test set.}
Results correspond to the same 100 sequences (14{,}304 frames), $\times4$
degradation settings, and evaluation protocol as Tab.~\ref{tab:main-test}.
PSNR, SSIM, and LPIPS are reported as mean $\pm$ standard deviation.
FVD is reported as a single video level score because the evaluation summaries
do not provide an FVD standard deviation. Best results are shown in \textbf{bold} and second best results are
\underline{underlined}.
}
\label{tab:main-test-expanded}

\footnotesize
\setlength{\tabcolsep}{2.4pt}
\renewcommand{\arraystretch}{0.94}

\begin{tabular}{@{}l cccc c cccc@{}}
\toprule
& \multicolumn{4}{c}{\textbf{Noisy} ($\sigma_y=0.05$)}
& &
\multicolumn{4}{c}{\textbf{Noiseless}} \\
\cmidrule(lr){2-5}\cmidrule(lr){7-10}
Method &
PSNR$\uparrow$ &
SSIM$\uparrow$ &
LPIPS$\downarrow$ &
FVD$\downarrow$ &
&
PSNR$\uparrow$ &
SSIM$\uparrow$ &
LPIPS$\downarrow$ &
FVD$\downarrow$ \\
\midrule

\multicolumn{10}{@{}l}{\textit{DM-based methods}} \\

\method{}
& \textbf{30.026{\tiny$\pm$2.21}}
& \textbf{0.862{\tiny$\pm$0.04}}
& \textbf{0.063{\tiny$\pm$0.02}}
& \textbf{17.476}
&
& \underline{31.778{\tiny$\pm$2.53}}
& \underline{0.905{\tiny$\pm$0.04}}
& \textbf{0.043{\tiny$\pm$0.02}}
& \textbf{8.526} \\

SVI-Diffusion
& 26.489{\tiny$\pm$1.01}
& 0.616{\tiny$\pm$0.03}
& 0.431{\tiny$\pm$0.08}
& 865.181
&
& 31.032{\tiny$\pm$2.43}
& 0.893{\tiny$\pm$0.04}
& 0.099{\tiny$\pm$0.04}
& 30.309 \\

Upscale-A-Video
& \underline{28.970{\tiny$\pm$1.93}}
& \underline{0.838{\tiny$\pm$0.05}}
& 0.154{\tiny$\pm$0.04}
& 194.191
&
& 29.223{\tiny$\pm$2.01}
& 0.856{\tiny$\pm$0.05}
& 0.149{\tiny$\pm$0.04}
& 139.189 \\

VISION-XL
& 26.993{\tiny$\pm$1.36}
& 0.701{\tiny$\pm$0.03}
& 0.421{\tiny$\pm$0.07}
& 875.044
&
& 29.473{\tiny$\pm$2.16}
& 0.850{\tiny$\pm$0.04}
& 0.197{\tiny$\pm$0.04}
& 94.496 \\

UltraVSR
& 27.440{\tiny$\pm$1.55}
& 0.823{\tiny$\pm$0.05}
& \underline{0.123{\tiny$\pm$0.04}}
& 198.494
&
& 27.853{\tiny$\pm$1.65}
& 0.838{\tiny$\pm$0.05}
& 0.110{\tiny$\pm$0.04}
& 93.860 \\

PS-SR
& 23.658{\tiny$\pm$1.72}
& 0.792{\tiny$\pm$0.05}
& 0.166{\tiny$\pm$0.04}
& \underline{174.431}
&
& 23.619{\tiny$\pm$1.73}
& 0.788{\tiny$\pm$0.05}
& 0.166{\tiny$\pm$0.04}
& 174.811 \\

StableVSR
& 25.444{\tiny$\pm$0.97}
& 0.527{\tiny$\pm$0.04}
& 0.432{\tiny$\pm$0.08}
& 1159.674
&
& 30.471{\tiny$\pm$2.40}
& 0.878{\tiny$\pm$0.05}
& \underline{0.068{\tiny$\pm$0.03}}
& 25.498 \\

StableVSR+BM3D
& 28.864{\tiny$\pm$1.89}
& 0.821{\tiny$\pm$0.05}
& 0.160{\tiny$\pm$0.04}
& 231.221
&
& --
& --
& --
& -- \\

\midrule
\multicolumn{10}{@{}l}{\textit{Non-DM-based methods}} \\

BasicVSR++
& 25.834{\tiny$\pm$2.15}
& 0.643{\tiny$\pm$0.05}
& 0.383{\tiny$\pm$0.08}
& 958.208
&
& \textbf{32.974{\tiny$\pm$3.16}}
& \textbf{0.924{\tiny$\pm$0.04}}
& 0.088{\tiny$\pm$0.04}
& \underline{19.891} \\

BasicVSR++ + BM3D
& 27.996{\tiny$\pm$2.59}
& 0.814{\tiny$\pm$0.06}
& 0.242{\tiny$\pm$0.06}
& 256.849
&
& --
& --
& --
& -- \\

\bottomrule
\end{tabular}

\end{supptable}

\begin{multicols}{2}
\raggedcolumns
\subsection{Computational cost}
\label{app:efficiency}

Tab.~\ref{tab:efficiency} reports the measured end-to-end runtime and peak memory usage on the full held-out VFHQ test set under the main $\times4$ setting for the evaluated methods.
\end{multicols}

\begin{supptable}
\centering
\caption{
\textbf{Measured computational cost on the held-out VFHQ test set under the main $\times4$ setting.}
We report end-to-end wall clock time and peak VRAM per device over the same 100 sequences (14{,}304 frames) as Tab.~\ref{tab:main-test}. All methods are evaluated on NVIDIA A100 80GB GPUs. \method{} uses four GPUs for parallel target frame processing, while all baselines use a single GPU.  Wallclock times are not normalized by GPU count.
}

\label{tab:efficiency}

\footnotesize
\setlength{\tabcolsep}{3.8pt}
\renewcommand{\arraystretch}{0.94}

\begin{tabular}{@{}l c rr c rr@{}}
\toprule
& & \multicolumn{2}{c}{\textbf{Noisy} ($\sigma_y=0.05$)}
& &
\multicolumn{2}{c}{\textbf{Noiseless}} \\
\cmidrule(lr){3-4}\cmidrule(lr){6-7}
Method &
Devices &
Wall time (s) &
Peak VRAM / device (MiB) &
&
Wall time (s) &
Peak VRAM / device (MiB) \\
\midrule

\multicolumn{7}{@{}l}{\textit{DM-based methods}} \\

\method{}
& 4 GPUs
& 87{,}780 & 80{,}107
&
& 88{,}140 & 80{,}195 \\

SVI-Diffusion
& 1 GPU
& 49{,}945 & 10{,}367
&
& 52{,}713 & 10{,}367 \\

Upscale-A-Video
& 1 GPU
& 61{,}182 & 3{,}553
&
& 61{,}172 & 3{,}553 \\

VISION-XL
& 1 GPU
& 53{,}845 & 8{,}589
&
& 58{,}093 & 8{,}589 \\

UltraVSR
& 1 GPU
& 20{,}973 & 5{,}247
&
& 20{,}976 & 5{,}247 \\

PS-SR
& 1 GPU
& 14{,}022 & 7{,}486
&
& 13{,}984 & 7{,}486 \\

StableVSR
& 1 GPU
& 433{,}222 & 6{,}115
&
& 432{,}473 & 6{,}115 \\

StableVSR+BM3D
& 1 GPU
& 433{,}362 & 6{,}115
&
& -- & -- \\

\midrule
\multicolumn{7}{@{}l}{\textit{Non-DM-based methods}} \\

BasicVSR++
& 1 GPU
& 2{,}720 & 2{,}410
&
& 2{,}720 & 2{,}410 \\

BasicVSR++ + BM3D
& 1 GPU
& 2{,}693 & 2{,}410
&
& -- & -- \\

\bottomrule
\end{tabular}
\end{supptable}

\begin{multicols}{2}
\raggedcolumns
\subsection{Architecture and training details}
\label{app:training_details}

\paragraph{Backbone architecture.}
\method{} uses a pixel-space DDPM denoiser based on the U-Net architecture from OpenAI \texttt{guided-diffusion}~\citep{dhariwal2021diffusion}. The model operates at $256{\times}256$ RGB resolution and follows the FFHQ face-model configuration used by DPS~\citep{chung2023diffusion}: 128 base channels, one residual block per resolution, attention at resolution 16, four attention heads with 64 channels per head, learned variance prediction, no class conditioning, scale-shift normalization, residual up/downsampling, zero dropout, and full-precision training. We use $1000$ diffusion steps with a linear noise schedule and learned-range variance. The model remains fully pixel-space and does not introduce an explicit temporal axis, 3D convolutions, optical flow, or transformer-based temporal modules.

\paragraph{Channel-wise temporal conditioning.}
For a target frame $i$, the denoiser receives the noisy diffusion state $\mathbf{x}_{i,t}$ together with a local temporal window of LR observations. LR measurements are generated using the same degradation operator used in the evaluation protocol at scale factors $\times2$, $\times4$, and $\times8$, producing $128{\times}128$, $64{\times}64$, and $32{\times}32$ frames, respectively, and are then back-projected to the $256{\times}256$ HR grid using $\mathcal{A}^{\top}$ before concatenation. With a $\pm5$ window, the model uses ten neighboring LR frames. In the noiseless model, we additionally concatenate the back-projected central observation $\tilde{\mathbf{y}}_i=\mathcal{A}^{\top}(\mathbf{y}_i)$, resulting in $36$ input channels: $3$ channels for $\mathbf{x}_{i,t}$, $30$ channels for the neighboring frames, and $3$ channels for the central observation. In the noisy model, the central degraded observation is omitted, resulting in $33$ input channels, and the conditioning frames are perturbed during training with i.i.d. Gaussian noise of standard deviation $\sigma_y=0.05$ in the normalized image range $[-1,1]$. The channel ordering is fixed across training and inference.

\paragraph{Initialization and optimization.}
We train separate models for the noiseless and noisy settings. The $\times4$ models are fine-tuned from the pretrained FFHQ checkpoint released with DPS and fine-tuned on VFHQ, with the newly added input channels. Since temporal conditioning only changes the input dimensionality, only the first convolution is widened. The original weights corresponding to the noisy image channels are copied from the pretrained checkpoint, and the newly added conditioning-channel weights are initialized by replicating the pretrained RGB filters. Training follows the \texttt{guided-diffusion} pipeline with AdamW, learning rate $10^{-4}$, batch size $39$, and exponential moving average (EMA) decay $0.9999$. We train separate models for the noiseless and noisy settings on 4 NVIDIA A100 80GB GPUs. We select checkpoints by validation denoising quality: for each candidate checkpoint we evaluate the PSNR of the Tweedie clean image estimate $\hat{\mathbf{x}}_0$ on the validation set across diffusion timesteps, at 100-timestep intervals, and choose the model with the best integrated denoising performance. The selected $\times4$ checkpoints correspond to approximately 24 epochs ($1.97\mathrm{M}$ iterations) for the noiseless model and 26 epochs ($2.12\mathrm{M}$ iterations) for the noisy model.

For the $\times2$ and $\times8$ settings, we continue fine-tuning from the selected $\times4$ checkpoints using the corresponding scale-specific degradation, while keeping the same optimization and checkpoint selection protocol. The $\times2$ models are trained for approximately $19$ additional epochs (1.53M iterations) in the noiseless setting and $15$ epochs (1.19M iterations) in the noisy setting, while the $\times8$ models are trained for approximately $16$ epochs (1.33M iterations) and $9$ epochs (0.69M iterations), respectively.

\paragraph{Sampling and guidance.}
At inference, the noise prediction of the trained denoiser is converted to the equivalent conditional score through
\cref{eq:score_from_epsilon}, and the resulting score estimator is used inside the $1000$-step DDPM DPS sampler. The likelihood term is anchored only to the central LR observation, while neighboring LR frames enter solely as denoiser conditioning. The forward degradation operator $\mathcal{A}$ is the same anti aliased bicubic operator, with the corresponding scale factor, used to generate the LR measurements. The data consistency gradient is computed by automatic differentiation through the implemented forward operator, yielding the required vector-Jacobian product without explicitly materializing an adjoint matrix or introducing a separately hand-coded upsampling approximation. The DPS guidance scale is tuned on the validation split. Unless otherwise stated, both models share all remaining sampling settings.

\paragraph{Boundary handling and temporal stochasticity.}
For frames near the beginning or end of a video sequence, missing conditioning frames are filled by mirror padding with respect to the available side, so that the same window size is used for all target frames. To reduce sampling-induced flicker, we draw the full sampling noise trajectory once per video sequence and reuse it for all frames in that video sequence. This keeps stochastic high-frequency details more temporally aligned, while the DPS guidance term preserves consistency with each frame's own LR measurement.

\textbf{Design rationale.} Two natural alternatives are worth contrasting. First, conditioning each frame on \emph{previously reconstructed} frames creates a serial dependency in which early errors are recycled and compound over time; in our experiments such autoregressive conditioning caused pronounced error accumulation and degraded reconstruction quality, which motivated anchoring the conditioning to the fixed, reliable LR measurements instead. Second, explicitly aligning neighbors via optical flow is fragile at LR: our attempts to warp neighboring frames (during training, as additional conditioning, or as guidance interpolation at inference) introduced flicker and warping artifacts rather than removing them. Delegating motion handling entirely to the learned conditional prior demonstrated more elegance and reliability.

\subsection{Additional scale factors}
\label{app:additional_scales}

We additionally evaluate \method{} at $\times2$ and $\times8$ using the same held-out VFHQ test set and metric pipeline as in the main $\times4$ evaluation, with the corresponding scale-specific bicubic downsampling operator. Tab.~\ref{tab:additional_scales} reports the results for all three scale factors. The corresponding LR resolutions are $128{\times}128$, $64{\times}64$, and $32{\times}32$ for $\times2$, $\times4$, and $\times8$, respectively, with reconstruction to $256{\times}256$. Representative qualitative results for the $\times2$ and $\times8$ settings
are shown in Figure~\ref{fig:additional_scales_visual}.

For each additional scale factor, we independently select the DPS guidance scale $\zeta'$ on the VFHQ validation split, following the same protocol used for $\times4$. For $\times2$, the selected values are $\zeta'=1.6$ and $\zeta'=1.2$ for the noiseless and noisy settings, respectively. For $\times8$, the corresponding values are $\zeta'=11.0$ and $\zeta'=1.3$.

As expected, the reconstruction becomes increasingly challenging as the scale factor grows, with lower PSNR and SSIM and higher LPIPS and FVD at larger upscaling factors. Nevertheless, \method{} remains effective across all three scales under both noiseless and noisy degradation. This is particularly notable at $\times8$, which constitutes a substantially more difficult and severely underdetermined reconstruction problem: a $256{\times}256$ HR frame must be recovered from a measurement of only $32{\times}32$ pixels. Such aggressive downsampling results in substantial information loss and makes the recovery of fine facial details considerably more challenging. Despite this extreme degradation, \method{} continues to produce high-quality reconstructions, demonstrating robustness even in the highly demanding $\times8$ setting.
\end{multicols}

\clearpage

\begin{supptable}
\centering
\caption{
Quantitative performance of \method{} across $\times2$, $\times4$, and $\times8$ face VSR on the held-out VFHQ test set (100 sequences, 14{,}304 frames) under noisy and noiseless bicubic degradation. The $\times4$ results are repeated from Tab.~\ref{tab:main-test} for reference.
}
\label{tab:additional_scales}
\begin{tabular}{c cccc c cccc}
\toprule
& \multicolumn{4}{c}{\textbf{Noisy} ($\sigma_y=0.05$)} & & \multicolumn{4}{c}{\textbf{Noiseless}} \\
\cmidrule(lr){2-5}\cmidrule(lr){7-10}
Scale & PSNR$\uparrow$ & SSIM$\uparrow$ & LPIPS$\downarrow$ & FVD$\downarrow$ & & PSNR$\uparrow$ & SSIM$\uparrow$ & LPIPS$\downarrow$ & FVD$\downarrow$ \\
\midrule
$\times2$ & 35.358 & 0.942 & 0.017 & 2.115 & & 38.858 & 0.977 & 0.006 & 0.340 \\
$\times4$ & 30.026 & 0.862 & 0.063 & 17.476 & & 31.778 & 0.905 & 0.043 & 8.526 \\
$\times8$ & 25.168 & 0.727 & 0.142 & 89.930 & & 26.783 & 0.775 & 0.108 & 50.981 \\
\bottomrule
\end{tabular}
\end{supptable}

\begin{suppfigure}
\centering
\setlength{\tabcolsep}{0.7pt}
\renewcommand{\arraystretch}{0.92}

\begin{tabular}{@{}c@{\hspace{1pt}}c@{\hspace{1pt}}c@{\hspace{1pt}}c@{}}
&
\small\textbf{LR input}
&
\small\textbf{\method{}}
&
\small\textbf{Ground truth}
\\

\raisebox{0.08\textwidth}{%
    \rotatebox{90}{\small\textbf{$\times2$, $\sigma_y=0$}}%
}
&
\includegraphics[interpolate=false,width=0.225\textwidth]{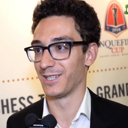}
&
\includegraphics[interpolate=false,width=0.225\textwidth]{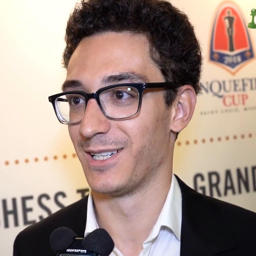}
&
\includegraphics[interpolate=false,width=0.225\textwidth]{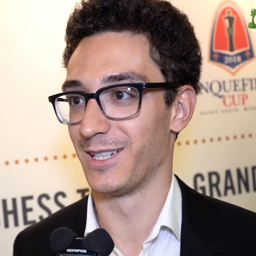}
\\

\raisebox{0.08\textwidth}{%
    \rotatebox{90}{\small\textbf{$\times2$, $\sigma_y=0.05$}}%
}
&
\includegraphics[interpolate=false,width=0.225\textwidth]{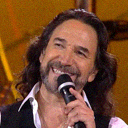}
&
\includegraphics[interpolate=false,width=0.225\textwidth]{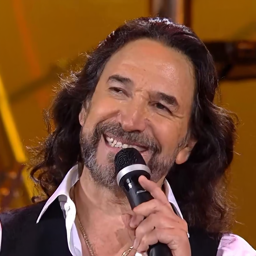}
&
\includegraphics[interpolate=false,width=0.225\textwidth]{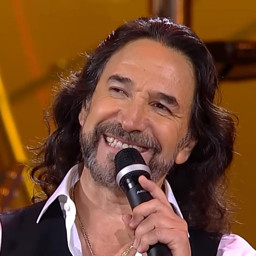}
\\

\raisebox{0.08\textwidth}{%
    \rotatebox{90}{\small\textbf{$\times8$, $\sigma_y=0$}}%
}
&
\includegraphics[interpolate=false,width=0.225\textwidth]{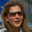}
&
\includegraphics[interpolate=false,width=0.225\textwidth]{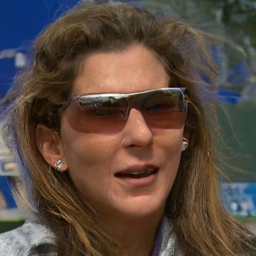}
&
\includegraphics[interpolate=false,width=0.225\textwidth]{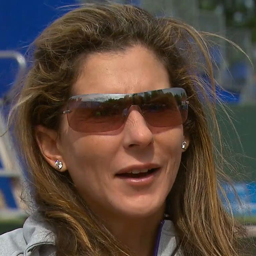}
\\

\raisebox{0.08\textwidth}{%
    \rotatebox{90}{\small\textbf{$\times8$, $\sigma_y=0.05$}}%
}
&
\includegraphics[interpolate=false,width=0.225\textwidth]{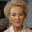}
&
\includegraphics[interpolate=false,width=0.225\textwidth]{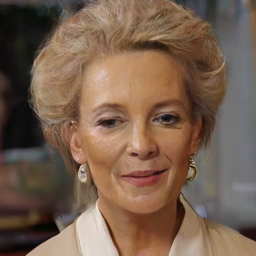}
&
\includegraphics[interpolate=false,width=0.225\textwidth]{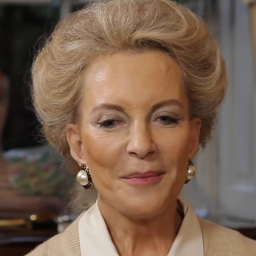}
\\

\end{tabular}

\caption{
\textbf{Qualitative comparison on the VFHQ test set for $\times2$ and $\times8$ VSR.}
Representative results are brought for noiseless bicubic degradation ($\sigma_y=0$) and additive Gaussian noise ($\sigma_y=0.05$). Columns show the enlarged LR input, \method{}, and the ground truth. Within each row, all images correspond to the same target frame. Across both scale factors and degradation settings, \method{} faithfully recovers fine facial and background structures.
}
\label{fig:additional_scales_visual}
\end{suppfigure}

\clearpage

\begin{multicols}{2}
\raggedcolumns
\section{Extended ablation studies}
\label{app:extra_ablations}

This section expands the ablations summarized in Sec.~\ref{sec:experiments}. Unless stated otherwise, all variants are evaluated on the VFHQ validation split under the main $\times4$ setting using matched sampling noise trajectories. Architecture choices and guidance scales are selected by validation PSNR; SSIM, LPIPS, and FVD are then reported. This protocol avoids selecting a separate configuration for each metric and makes the trade-offs between frame-level fidelity and video-level quality explicit.

\subsection{Temporal conditioning, central observation, and guidance}
\phantomsection
\label{app:full_temporal_ablation}

Tab.~\ref{tab:temporal_guidance_ablation} extends the compact ablation summary in Tab.~\ref{tab:compact_ablations} by reporting the full set of end-to-end reconstruction metrics and additional guidance scale values. We first analyze the denoiser itself across diffusion timesteps, and then evaluate the corresponding design choices within the full reconstruction pipeline. 

\paragraph{Denoiser level analysis.} Before evaluating the complete reconstruction pipeline, we isolate the effect of the conditioning architecture on the learned denoiser. For each architecture, we independently select its best training checkpoint according to the validation denoising criterion described in App.~\ref{app:training_details}. Figure~\ref{fig:denoiser_performance} reports the PSNR of the Tweedie clean estimate $\hat{\mathbf{x}}_0$ on the validation set across diffusion timesteps, evaluated at 100-timestep intervals.

\paragraph{End-to-end reconstruction.} We next evaluate the same design choices within the complete reconstruction pipeline. Tab.~\ref{tab:temporal_guidance_ablation} reports the full quantitative results, while Figures~\ref{fig:temporal_context_noiseless} and~\ref{fig:temporal_context_noisy} show the joint effects of temporal context and guidance strength. Figures~\ref{fig:guidance_sweep_noiseless} and~\ref{fig:guidance_sweep_noisy} further isolate the contribution of temporal conditioning by directly comparing the selected $k=5$ configuration with the $k=0$ baseline across guidance scales. Temporal conditioning consistently improves all four reconstruction metrics. The resulting PSNR-optimal operating points are $\zeta'=5.2$ and $\zeta'=1.2$ for the noiseless and noisy settings, respectively.

\paragraph{Temporal context radius.}
The temporal radius ablation reveals two complementary effects. First, temporal conditioning is essential: removing neighboring frames ($k=0$) substantially degrades both reconstruction fidelity and video-level consistency, reducing PSNR to $27.469$ dB and $26.700$ dB and increasing FVD to $161.338$ and $446.632$ in the noiseless and noisy settings, respectively. Second, increasing the temporal radius is beneficial only up to a point. In the noiseless setting, $k=5$ achieves the best PSNR, while nearby radii can yield slightly better LPIPS or FVD. Under noise, $k=5$ again gives the best PSNR and LPIPS, whereas $k=7$ attains the best SSIM and FVD. Overall, these results indicate a clear trade-off: additional neighboring observations improve temporal stability and reconstruction quality, but increasingly distant frames provide diminishing benefit as motion, expression changes, and occlusions reduce their relevance to the target frame. We therefore select $k=5$ as the best overall operating point.
\end{multicols}

\begin{suppfigure}
\centering
\subcaptionbox{Noiseless degradation.}{
    \includegraphics[width=0.475\textwidth]{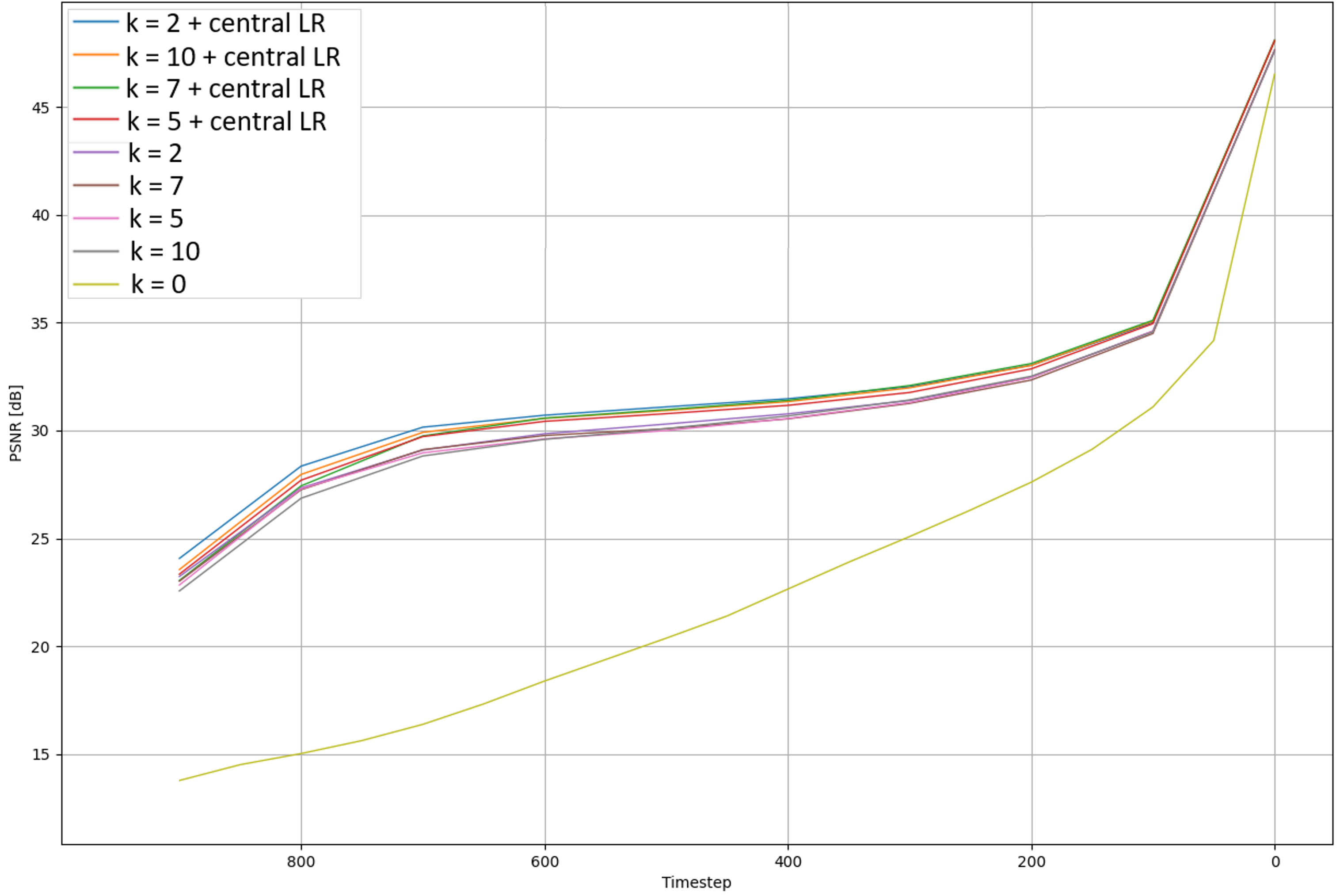}
}
\hspace{0.02\textwidth}
\subcaptionbox{Noisy degradation ($\sigma_y=0.05$).}{
    \includegraphics[width=0.475\textwidth]{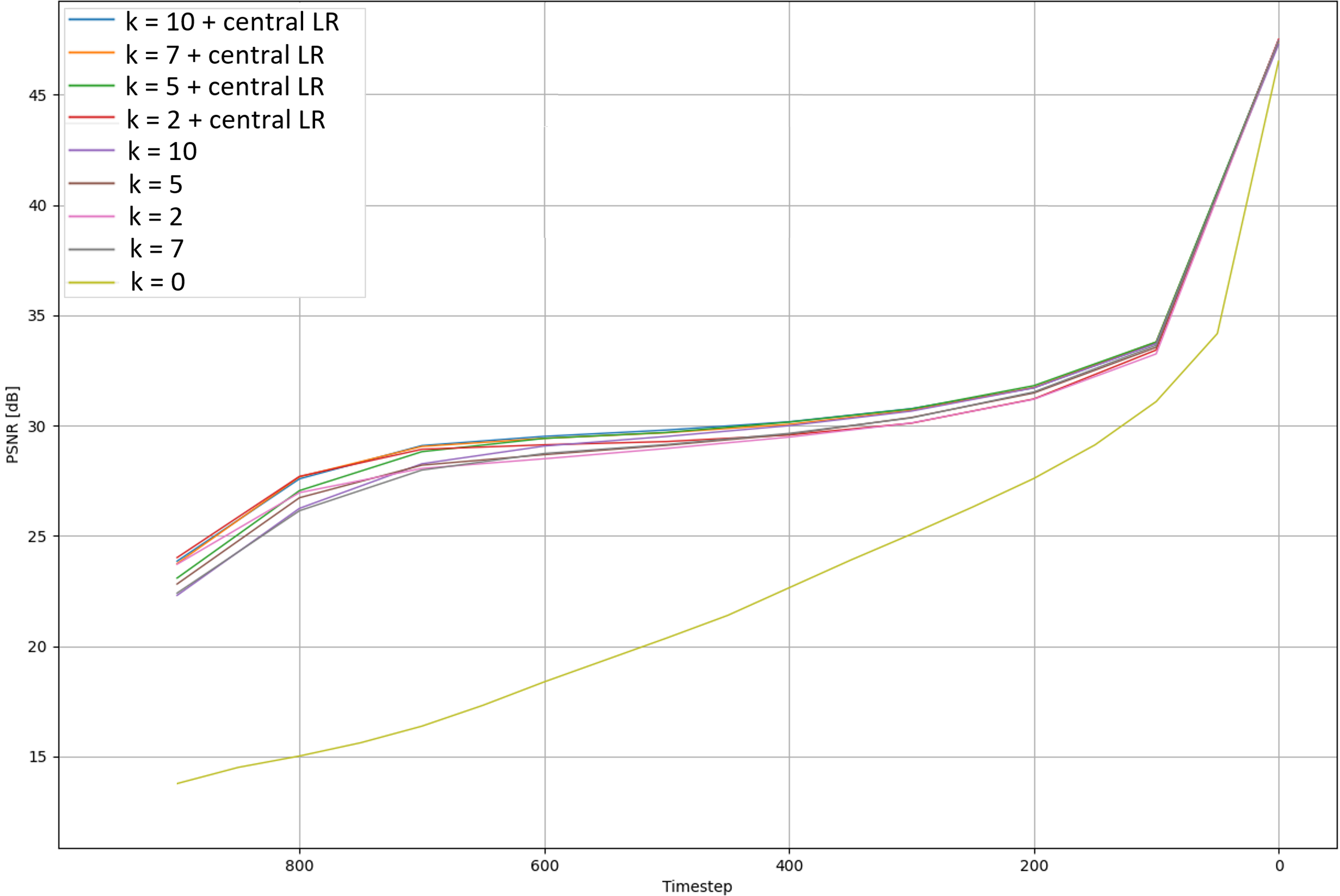}
}
\caption{
\textbf{Denoiser performance across diffusion timesteps.}
For each conditioning architecture, we report the validation set PSNR of the
Tweedie clean estimate $\hat{\mathbf{x}}_0$ using its independently selected
best training checkpoint. PSNR is evaluated at 100-timestep intervals.
$k$ denotes the temporal conditioning radius. Legend entries are ordered by the integrated PSNR across diffusion timesteps.
}
\label{fig:denoiser_performance}
\end{suppfigure}

\clearpage

\begin{suppfigure}
\centering
\includegraphics[interpolate=false,width=0.85\textwidth]{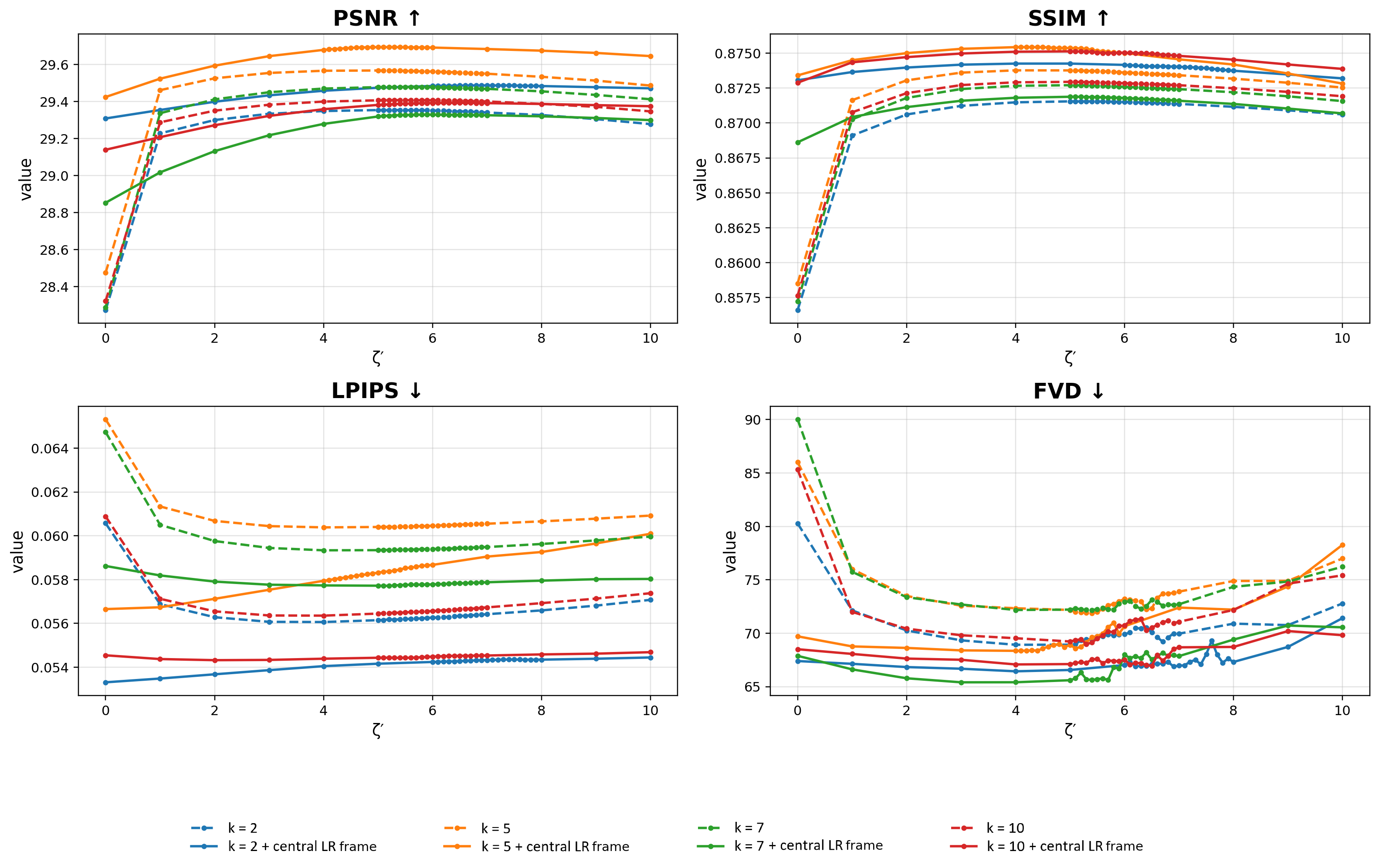}
\caption{
\textbf{Temporal-context ablation under noiseless degradation.}
We vary the number of neighboring LR frames on each side of the target and whether the back-projected central LR observation is included in the denoiser input. Dashed curves omit the central observation, whereas solid curves include it. Colors indicate the temporal-context radius. All variants are evaluated using matched sampling-noise trajectories, while sweeping the DPS guidance step size $\zeta'$.}
\label{fig:temporal_context_noiseless}
\end{suppfigure}

\begin{suppfigure}
\centering
\includegraphics[interpolate=false,width=0.85\textwidth]{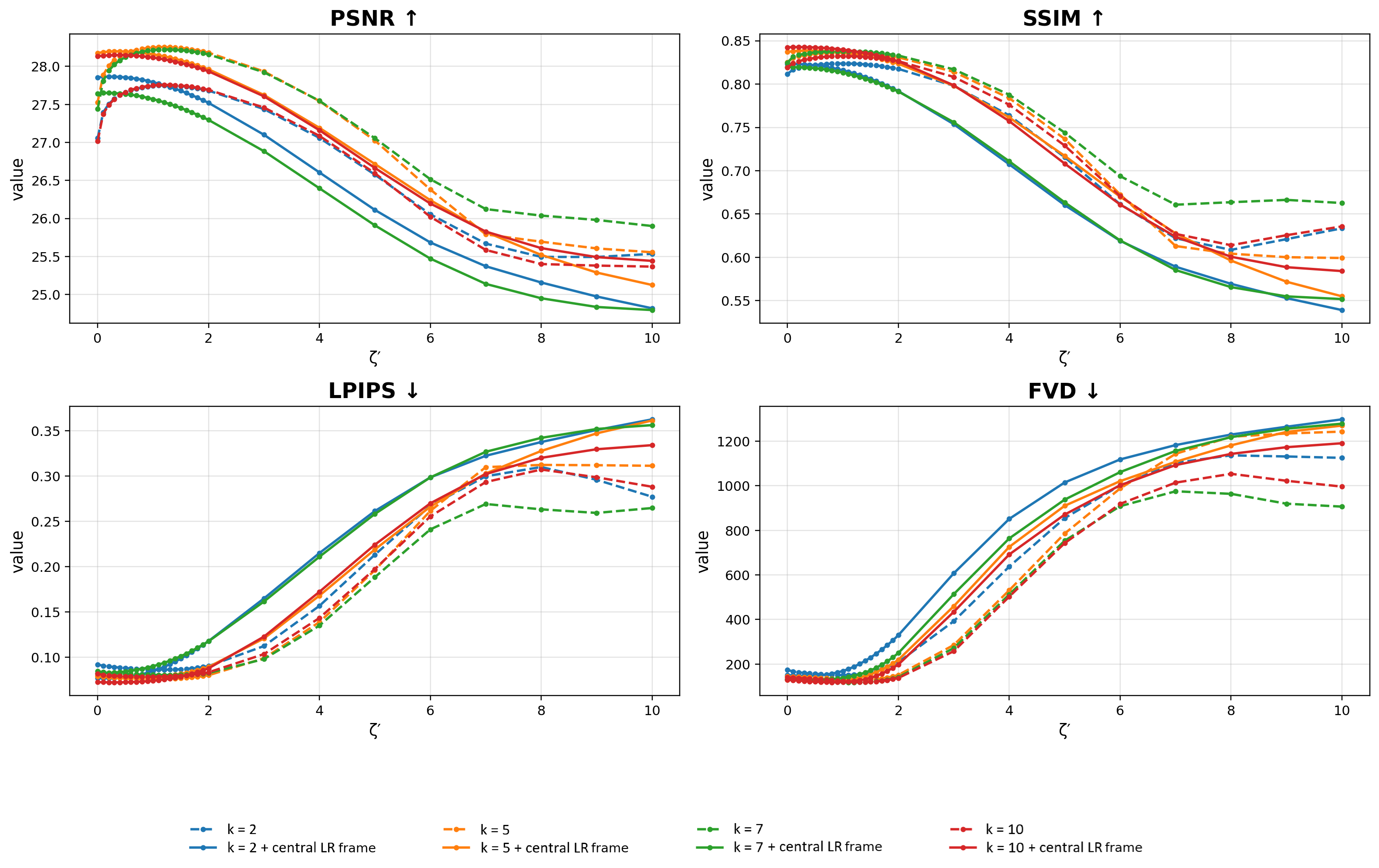}
\caption{
\textbf{Temporal-context ablation under noisy degradation.}
Dashed curves omit the noisy central LR observation from the denoiser input, whereas solid curves include it. Colors indicate the temporal-context radius, and all variants use matched sampling-noise trajectories. The results expose a trade-off between frame-level fidelity and video-level quality: central-frame conditioning improves SSIM and FVD for some configurations, but does not improve validation PSNR or LPIPS.}
\label{fig:temporal_context_noisy}
\end{suppfigure}

\clearpage

\begin{suppfigure}
\centering
\includegraphics[interpolate=false,width=0.85\textwidth,trim=0bp 14bp 0bp 6bp,clip]{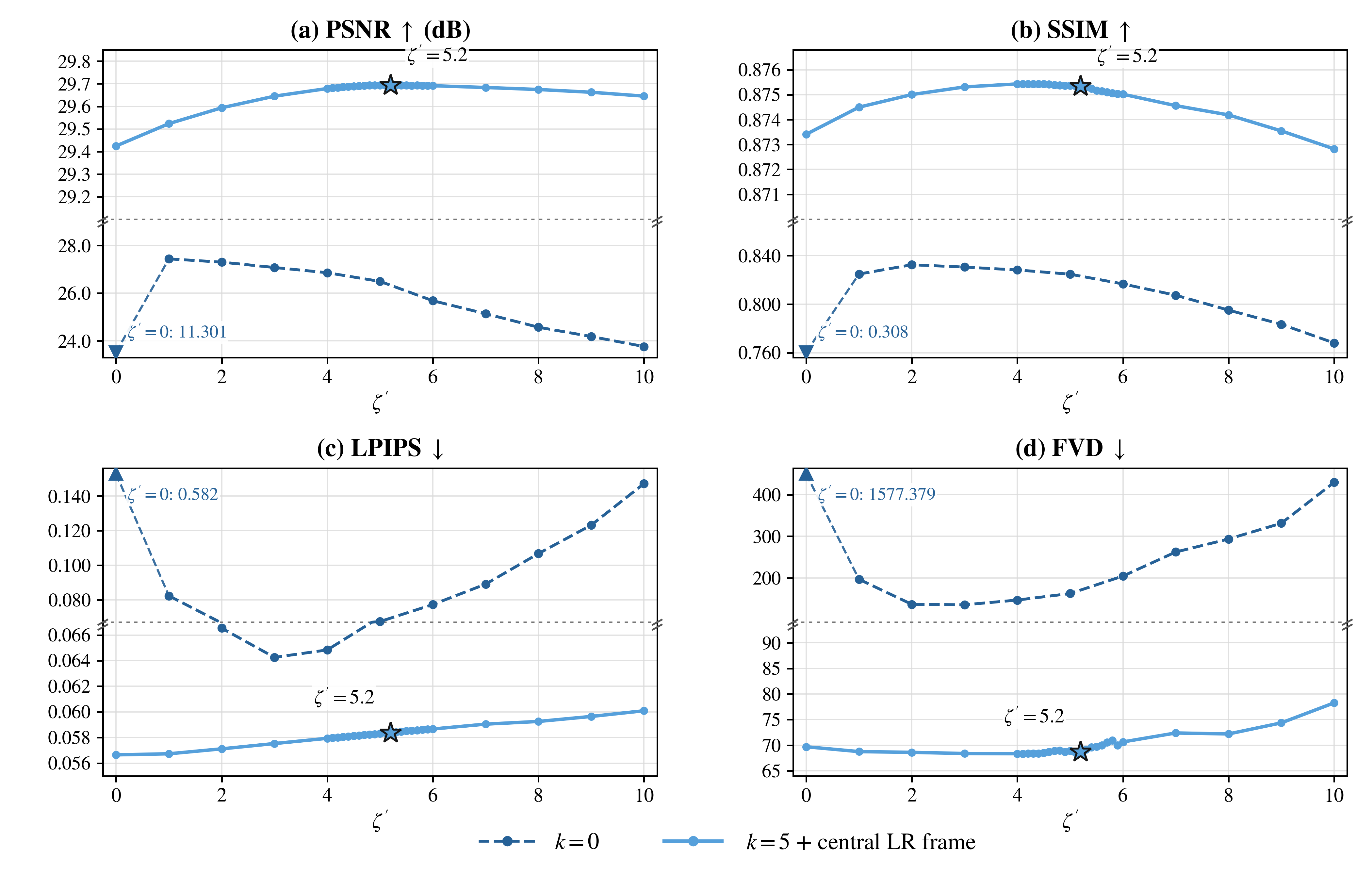}
\caption{
\textbf{DPS guidance scale sensitivity under noiseless degradation.} 
PSNR, SSIM, LPIPS, and FVD on the VFHQ validation split for $\times4$ VSR, comparing the frame-wise baseline ($k=0$, without LR denoiser conditioning; dashed) with the selected temporal configuration ($k=5$, solid). For $k=5$, the denoiser conditions on ten neighboring LR frames together with the central LR observation. Stars mark the PSNR-selected operating point $\zeta'=5.2$, used for all metrics. Piecewise linear $y$ axes are separated by horizontal dashed lines; triangles indicate the off-scale $k=0,\zeta'=0$ values reported alongside. Markers denote measured values, with unsmoothed lines connecting them. 
}
\label{fig:guidance_sweep_noiseless}
\end{suppfigure}

\begin{suppfigure}
\centering
\includegraphics[interpolate=false,width=0.85\textwidth,trim=0bp 14bp 0bp 6bp,clip]{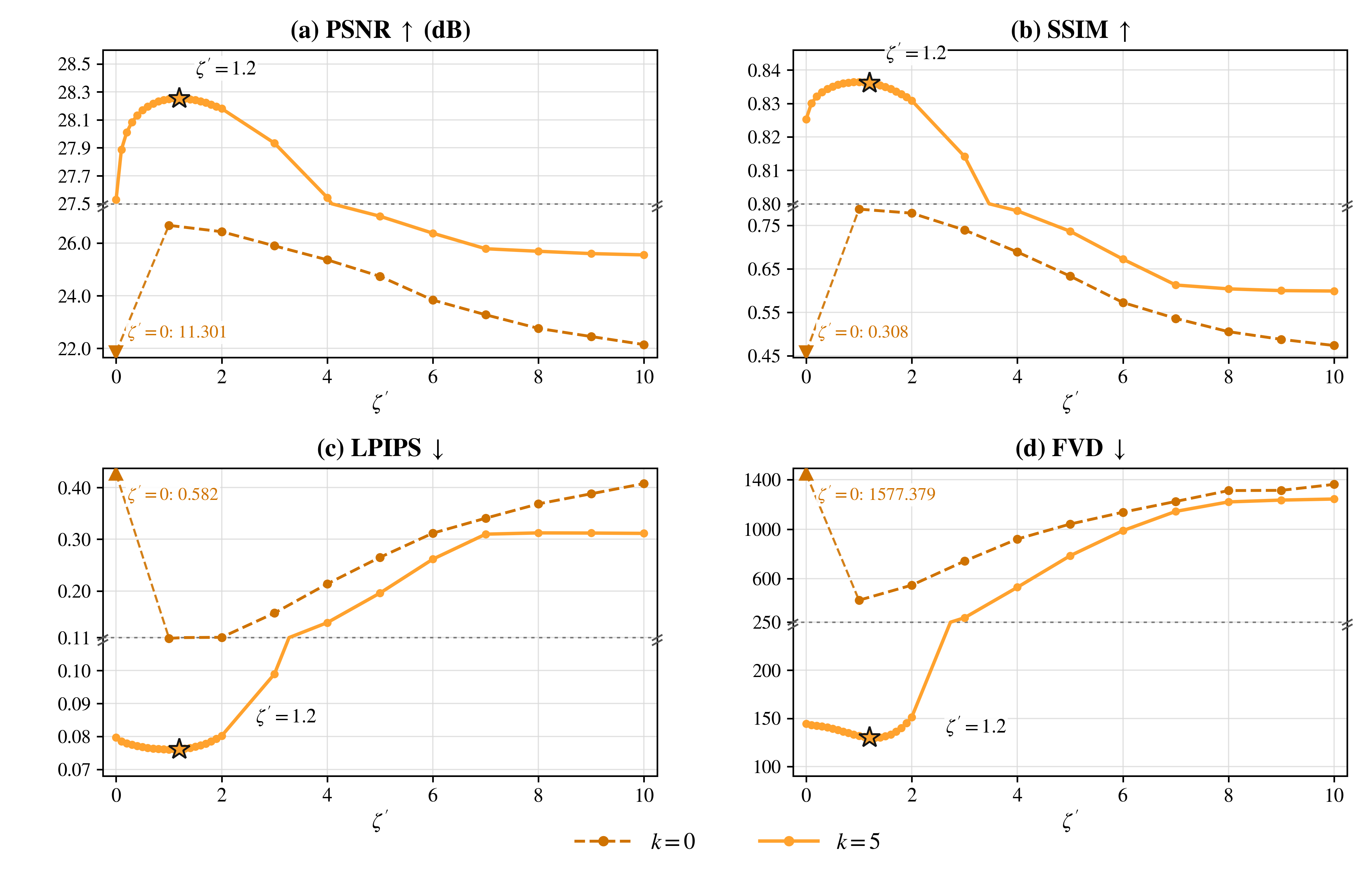}
\caption{
\textbf{DPS guidance scale sensitivity under noisy degradation.}
PSNR, SSIM, LPIPS, and FVD on the VFHQ validation split for
$\times4$ VSR with $\sigma_y=0.05$, comparing the frame-wise baseline ($k=0$, without LR denoiser conditioning; dashed) with
the selected $k=5$ temporal conditioning (solid).
For $k=5$, the denoiser conditions on ten neighboring LR frames and
excludes the central LR frame, while DPS retains the central measurement.
Stars mark the PSNR-selected operating point $\zeta'=1.2$, used for all
metrics. Axis and marker conventions follow
Figure~\ref{fig:guidance_sweep_noiseless}.
}
\label{fig:guidance_sweep_noisy}
\end{suppfigure}

\clearpage

\begin{multicols}{2}
\raggedcolumns
\paragraph{Central LR observation.}
In the noiseless setting, adding the back-projected central observation to the denoiser input improves the selected PSNR from $29.568$ to $29.694$ dB and also improves SSIM, LPIPS, and FVD. This is consistent with the central LR frame acting as a reliable, spatially aligned anchor for the conditional prior. Under measurement noise, the behavior changes: excluding the central LR observation gives the best validation PSNR ($28.254$ dB) and LPIPS, whereas including it gives slightly better SSIM and FVD. We therefore use the central LR observation as denoiser conditioning only in the noiseless model. In both settings, the central measurement still enters the DPS likelihood term.

\columnbreak
\paragraph{Data-consistency guidance.}
Turning off DPS guidance degrades reconstruction fidelity in both regimes. In the noiseless setting, PSNR increases from $29.424$ dB at $\zeta'=0$ to $29.694$ dB at the selected $\zeta'=5.2$. In the noisy setting, it increases from $27.530$ dB to $28.254$ dB at $\zeta'=1.2$. The sweep also shows that stronger guidance is not monotonically better: overly large values eventually worsen both frame-level and video-level metrics. The lower selected guidance scale in the noisy setting is consistent with the reduced reliability of the noisy measurement.
\end{multicols}

\begin{supptable}
\centering
\caption{
\textbf{Expanded validation ablations for $\times4$ VSR.}
This table directly expands Tab.~\ref{tab:compact_ablations}: the same
configurations, guidance scales $\zeta'$, PSNR, and FVD values are retained,
while SSIM, LPIPS, and per-video standard deviations for the frame-level
metrics are additionally reported.
The symbol $\star$ denotes the configuration selected for subsequent experiments.
}
\label{tab:temporal_guidance_ablation}
\vspace{1.5pt}

\small
\setlength{\aboverulesep}{0.3ex}
\setlength{\belowrulesep}{0.45ex}
\setlength{\tabcolsep}{3.2pt}
\renewcommand{\arraystretch}{0.90}

\textbf{(a) Temporal context}
\vspace{0pt}

\begin{minipage}[t]{0.485\textwidth}
\centering
\textit{Noiseless: without central LR denoiser conditioning}

\vspace{0pt}
\begin{tabular}{@{}cccccc@{}}
\toprule
$k$ &
$\zeta'$ &
PSNR$\uparrow$ &
SSIM$\uparrow$ &
LPIPS$\downarrow$ &
FVD$\downarrow$ \\
\midrule

0
& 1.3
& 27.469{\tiny$\pm$2.57}
& 0.8308{\tiny$\pm$0.08}
& 0.0736{\tiny$\pm$0.01}
& 161.338 \\

2
& 5.3
& 29.354{\tiny$\pm$2.71}
& 0.8716{\tiny$\pm$0.07}
& 0.0562{\tiny$\pm$0.03}
& 69.429 \\

$5^{\star}$
& 5.0
& \textbf{29.568{\tiny$\pm$2.73}}
& 0.8738{\tiny$\pm$0.08}
& 0.0604{\tiny$\pm$0.03}
& 72.177 \\

7
& 5.5
& 29.479{\tiny$\pm$2.70}
& 0.8727{\tiny$\pm$0.08}
& 0.0594{\tiny$\pm$0.04}
& 72.258 \\

10
& 5.4
& 29.408{\tiny$\pm$2.68}
& 0.8729{\tiny$\pm$0.08}
& 0.0565{\tiny$\pm$0.03}
& 69.126 \\

\bottomrule
\end{tabular}
\end{minipage}
\hfill
\begin{minipage}[t]{0.485\textwidth}
\centering
\textit{Noisy: without central LR denoiser conditioning}

\vspace{0pt}
\begin{tabular}{@{}cccccc@{}}
\toprule
$k$ &
$\zeta'$ &
PSNR$\uparrow$ &
SSIM$\uparrow$ &
LPIPS$\downarrow$ &
FVD$\downarrow$ \\
\midrule

0
& 1.2
& 26.700{\tiny$\pm$2.31}
& 0.7900{\tiny$\pm$0.08}
& 0.1046{\tiny$\pm$0.02}
& 446.632 \\

2
& 1.2
& 27.750{\tiny$\pm$2.18}
& 0.8235{\tiny$\pm$0.08}
& 0.0864{\tiny$\pm$0.03}
& 152.476 \\

$5^{\star}$
& 1.2
& \textbf{28.254{\tiny$\pm$2.27}}
& 0.8361{\tiny$\pm$0.07}
& 0.0762{\tiny$\pm$0.02}
& 130.086 \\

7
& 1.2
& 28.222{\tiny$\pm$2.49}
& 0.8377{\tiny$\pm$0.08}
& 0.0805{\tiny$\pm$0.03}
& 118.395 \\

10
& 1.2
& 27.756{\tiny$\pm$2.46}
& 0.8322{\tiny$\pm$0.08}
& 0.0787{\tiny$\pm$0.03}
& 120.337 \\

\bottomrule
\end{tabular}
\end{minipage}

\vspace{1.5pt}

\textbf{(b) Central LR denoiser conditioning}
\vspace{0pt}

\begin{minipage}[t]{0.485\textwidth}
\centering
\textit{Noiseless: $k=5$}

\vspace{0pt}
\begin{tabular}{@{}lccccc@{}}
\toprule
Central LR &
$\zeta'$ &
PSNR$\uparrow$ &
SSIM$\uparrow$ &
LPIPS$\downarrow$ &
FVD$\downarrow$ \\
\midrule

Excluded
& 5.0
& 29.568{\tiny$\pm$2.73}
& 0.8738{\tiny$\pm$0.08}
& 0.0604{\tiny$\pm$0.03}
& 72.177 \\

Included $\star$
& 5.2
& \textbf{29.694{\tiny$\pm$2.82}}
& 0.8753{\tiny$\pm$0.08}
& 0.0584{\tiny$\pm$0.03}
& 68.720 \\

\bottomrule
\end{tabular}
\end{minipage}
\hfill
\begin{minipage}[t]{0.485\textwidth}
\centering
\textit{Noisy: $k=5$}

\vspace{0pt}
\begin{tabular}{@{}lccccc@{}}
\toprule
Central LR &
$\zeta'$ &
PSNR$\uparrow$ &
SSIM$\uparrow$ &
LPIPS$\downarrow$ &
FVD$\downarrow$ \\
\midrule

Excluded $\star$
& 1.2
& \textbf{28.254{\tiny$\pm$2.27}}
& 0.8361{\tiny$\pm$0.07}
& 0.0762{\tiny$\pm$0.02}
& 130.086 \\

Included
& 0.4
& 28.198{\tiny$\pm$2.39}
& 0.8386{\tiny$\pm$0.08}
& 0.0782{\tiny$\pm$0.02}
& 125.455 \\

\bottomrule
\end{tabular}
\end{minipage}

\vspace{1.5pt}

\textbf{(c) Data-consistency guidance scale}
\vspace{0pt}

\begin{minipage}[t]{0.485\textwidth}
\centering
\textit{Noiseless: $k=5$, with central LR}

\vspace{0pt}
\begin{tabular}{@{}ccccc@{}}
\toprule
$\zeta'$ &
PSNR$\uparrow$ &
SSIM$\uparrow$ &
LPIPS$\downarrow$ &
FVD$\downarrow$ \\
\midrule

0.0
& 29.424{\tiny$\pm$2.73}
& 0.8734{\tiny$\pm$0.08}
& 0.0567{\tiny$\pm$0.03}
& 69.709 \\

2.0
& 29.594{\tiny$\pm$2.79}
& 0.8750{\tiny$\pm$0.08}
& 0.0571{\tiny$\pm$0.03}
& 68.623 \\

4.0
& 29.679{\tiny$\pm$2.82}
& 0.8754{\tiny$\pm$0.08}
& 0.0579{\tiny$\pm$0.03}
& 68.351 \\

$5.2^{\star}$
& \textbf{29.694{\tiny$\pm$2.82}}
& 0.8753{\tiny$\pm$0.08}
& 0.0584{\tiny$\pm$0.03}
& 68.720 \\

7.0
& 29.684{\tiny$\pm$2.81}
& 0.8746{\tiny$\pm$0.08}
& 0.0590{\tiny$\pm$0.03}
& 72.399 \\

10.0
& 29.645{\tiny$\pm$2.78}
& 0.8728{\tiny$\pm$0.08}
& 0.0601{\tiny$\pm$0.03}
& 78.273 \\

\bottomrule
\end{tabular}
\end{minipage}
\hfill
\begin{minipage}[t]{0.485\textwidth}
\centering
\textit{Noisy: $k=5$, without central LR}

\vspace{0pt}
\begin{tabular}{@{}ccccc@{}}
\toprule
$\zeta'$ &
PSNR$\uparrow$ &
SSIM$\uparrow$ &
LPIPS$\downarrow$ &
FVD$\downarrow$ \\
\midrule

0.0
& 27.530{\tiny$\pm$2.23}
& 0.8253{\tiny$\pm$0.07}
& 0.0797{\tiny$\pm$0.03}
& 144.390 \\

0.4
& 28.133{\tiny$\pm$2.23}
& 0.8344{\tiny$\pm$0.07}
& 0.0771{\tiny$\pm$0.03}
& 140.676 \\

0.8
& 28.232{\tiny$\pm$2.26}
& 0.8363{\tiny$\pm$0.07}
& 0.0762{\tiny$\pm$0.02}
& 134.818 \\

$1.2^{\star}$
& \textbf{28.254{\tiny$\pm$2.27}}
& 0.8361{\tiny$\pm$0.07}
& 0.0762{\tiny$\pm$0.02}
& 130.086 \\

1.6
& 28.233{\tiny$\pm$2.26}
& 0.8343{\tiny$\pm$0.07}
& 0.0773{\tiny$\pm$0.02}
& 133.309 \\

2.0
& 28.181{\tiny$\pm$2.24}
& 0.8309{\tiny$\pm$0.07}
& 0.0802{\tiny$\pm$0.02}
& 151.374 \\

\bottomrule
\end{tabular}
\end{minipage}

\vspace{1.5pt}

\textbf{(d) Boundary handling}
\vspace{0pt}

\begin{minipage}[t]{0.485\textwidth}
\centering
\textit{Noiseless: $k=5$, with central LR, $\zeta'=5.2$}

\vspace{0pt}
\begin{tabular}{@{}lcccc@{}}
\toprule
Padding &
PSNR$\uparrow$ &
SSIM$\uparrow$ &
LPIPS$\downarrow$ &
FVD$\downarrow$ \\
\midrule

Zero
& 29.685{\tiny$\pm$2.82}
& 0.8752{\tiny$\pm$0.08}
& 0.0586{\tiny$\pm$0.03}
& 68.766 \\

Mirroring $\star$
& \textbf{29.694{\tiny$\pm$2.82}}
& 0.8753{\tiny$\pm$0.08}
& 0.0584{\tiny$\pm$0.03}
& 68.720 \\

\bottomrule
\end{tabular}
\end{minipage}
\hfill
\begin{minipage}[t]{0.485\textwidth}
\centering
\textit{Noisy: $k=5$, without central LR, $\zeta'=1.2$}

\vspace{0pt}
\begin{tabular}{@{}lcccc@{}}
\toprule
Padding &
PSNR$\uparrow$ &
SSIM$\uparrow$ &
LPIPS$\downarrow$ &
FVD$\downarrow$ \\
\midrule

Zero
& 28.242{\tiny$\pm$2.27}
& 0.8358{\tiny$\pm$0.07}
& 0.0766{\tiny$\pm$0.03}
& 130.520 \\

Mirroring $\star$
& \textbf{28.254{\tiny$\pm$2.27}}
& 0.8361{\tiny$\pm$0.07}
& 0.0762{\tiny$\pm$0.02}
& 130.086 \\

\bottomrule
\end{tabular}
\end{minipage}

\vspace{1.5pt}

\textbf{(e) Shared noise trajectory}
\vspace{0pt}

\begin{minipage}[t]{0.485\textwidth}
\centering
\textit{Noiseless: $k=5$, with central LR, $\zeta'=5.2$}

\vspace{0pt}
\begin{tabular}{@{}lcccc@{}}
\toprule
Shared noise &
PSNR$\uparrow$ &
SSIM$\uparrow$ &
LPIPS$\downarrow$ &
FVD$\downarrow$ \\
\midrule

Excluded
& 29.702{\tiny$\pm$2.73}
& 0.8748{\tiny$\pm$0.07}
& 0.0579{\tiny$\pm$0.03}
& 82.795 \\

Included $\star$
& 29.694{\tiny$\pm$2.82}
& 0.8753{\tiny$\pm$0.08}
& 0.0584{\tiny$\pm$0.03}
& \textbf{68.720} \\

\bottomrule
\end{tabular}
\end{minipage}
\hfill
\begin{minipage}[t]{0.485\textwidth}
\centering
\textit{Noisy: $k=5$, without central LR, $\zeta'=1.2$}

\vspace{0pt}
\begin{tabular}{@{}lcccc@{}}
\toprule
Shared noise &
PSNR$\uparrow$ &
SSIM$\uparrow$ &
LPIPS$\downarrow$ &
FVD$\downarrow$ \\
\midrule

Excluded
& 28.240{\tiny$\pm$2.16}
& 0.8351{\tiny$\pm$0.07}
& 0.0771{\tiny$\pm$0.03}
& 164.753 \\

Included $\star$
& 28.254{\tiny$\pm$2.27}
& 0.8361{\tiny$\pm$0.07}
& 0.0762{\tiny$\pm$0.02}
& \textbf{130.086} \\

\bottomrule
\end{tabular}
\end{minipage}

\end{supptable}

\begin{multicols}{2}
\raggedcolumns

\subsection{Visual analysis of design choices}
\label{app:visual_ablations}

The quantitative ablations are also reflected in the reconstructed content.
We therefore provide representative examples illustrating the two design
choices with the clearest visual effect: the treatment of the central LR
observation and the use of a shared noise trajectory.

\paragraph{Central LR observation.}
Fig.~\ref{fig:visual_central_lr} illustrates the measurement-dependent role of
the central LR observation. Under noiseless degradation, conditioning the
denoiser on the central frame helps recover fine structures; in the shown example, the thin window stripes are more faithfully preserved. Under measurement noise, the opposite behavior is observed: omitting the noisy central frame from the denoiser input preserves these structures more clearly. This agrees with the configuration selected by validation PSNR and motivates our asymmetric conditioning policy. Importantly, the central
measurement is still used by the DPS data consistency term in both settings.

\noindent\textbf{Shared noise trajectory.}\enspace
Independent diffusion stochasticity can synthesize ambiguous fine details
differently across adjacent frames, producing visible flicker even when each
frame is individually plausible. Fig.~\ref{fig:visual_shared_noise} compares the
same five consecutive frames reconstructed with independent and shared noise
trajectories. With independent noise, fine structures fluctuate across the
sequence. Sharing the initial noise and the stepwise stochastic updates
substantially stabilizes these details while preserving the content induced by the measurements. This visual behavior is consistent with the pronounced FVD improvement observed in the quantitative ablation.
\end{multicols}

\begin{suppfigure}
\centering
\setlength{\tabcolsep}{0pt}
\renewcommand{\arraystretch}{1.0}

\begin{tabular}{@{}c c c c c@{}}
&
\small\textbf{LR input}
&
\small\textbf{Without central LR}
&
\small\textbf{With central LR}
&
\small\textbf{Ground truth}
\\

\raisebox{-.5\totalheight}{%
    \makebox[0.077\textwidth][c]{%
        \shortstack{%
            \small\bfseries Noiseless\\
            \small $\sigma_y=0$
        }%
    }%
}
&
\raisebox{-.5\totalheight}{%
    \includegraphics[interpolate=false,width=0.190\textwidth]
    {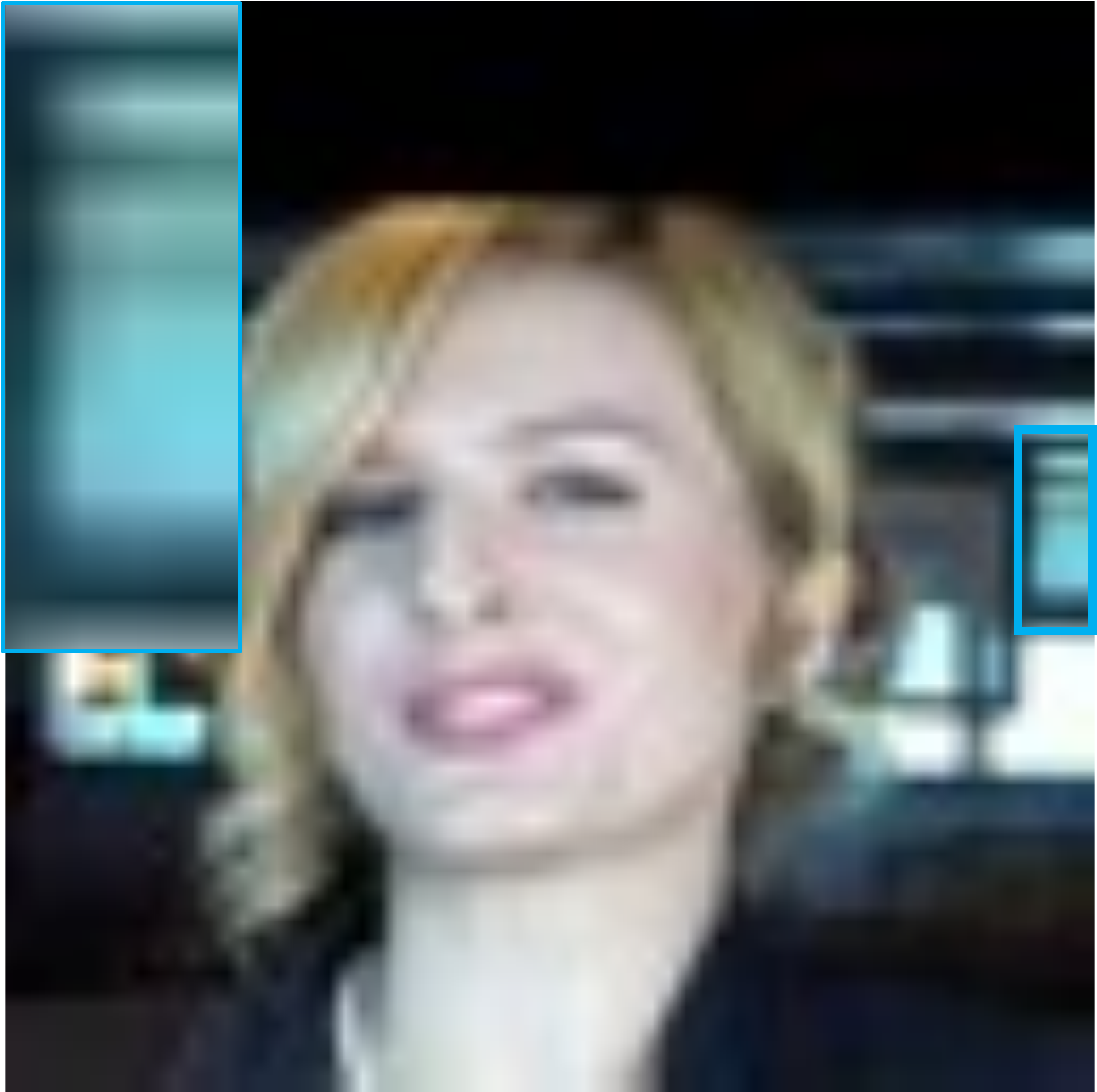}%
}
&
\raisebox{-.5\totalheight}{%
    \includegraphics[interpolate=false,width=0.190\textwidth]
    {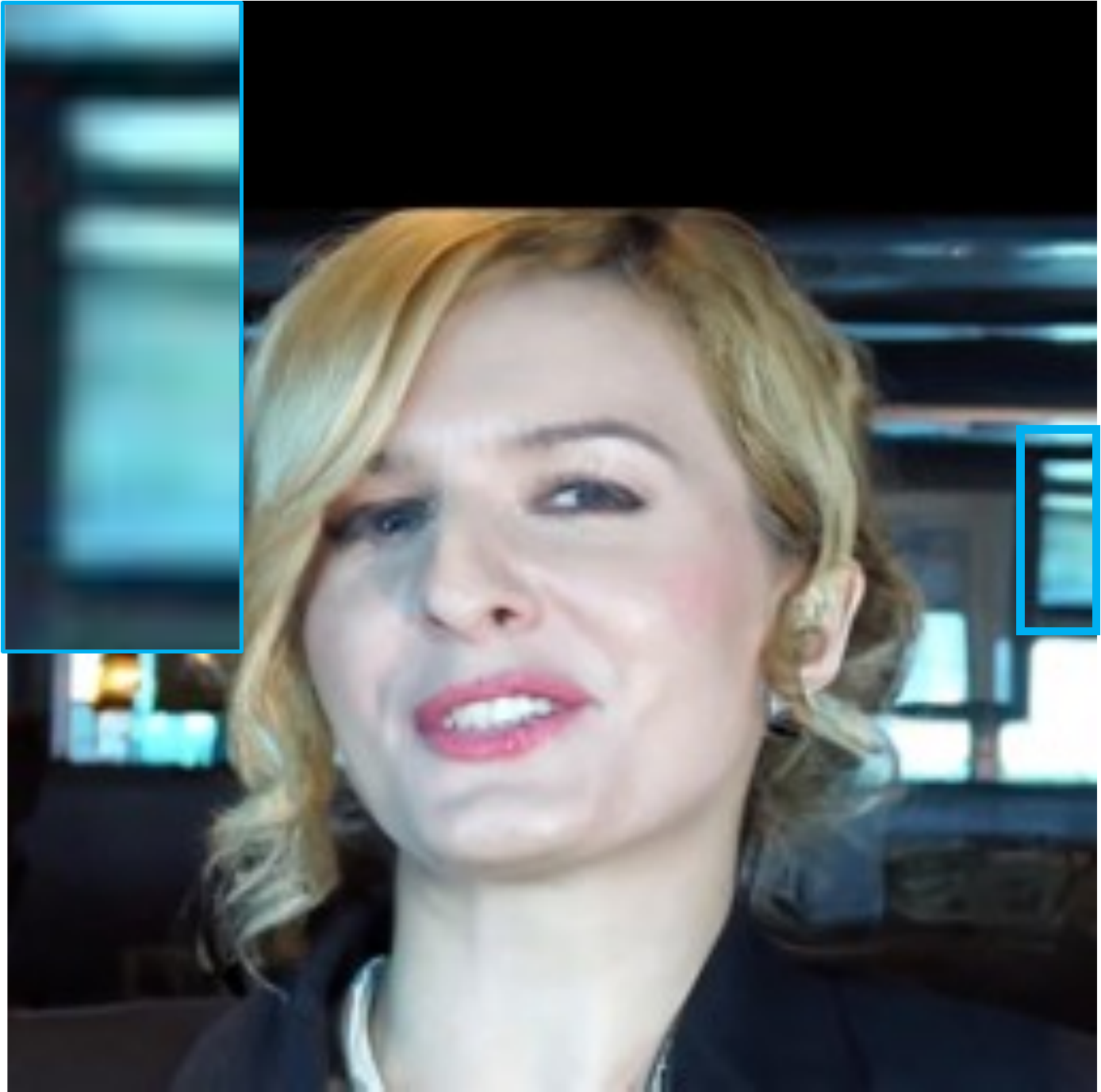}%
}
&
\raisebox{-.5\totalheight}{%
    \includegraphics[interpolate=false,width=0.190\textwidth]
    {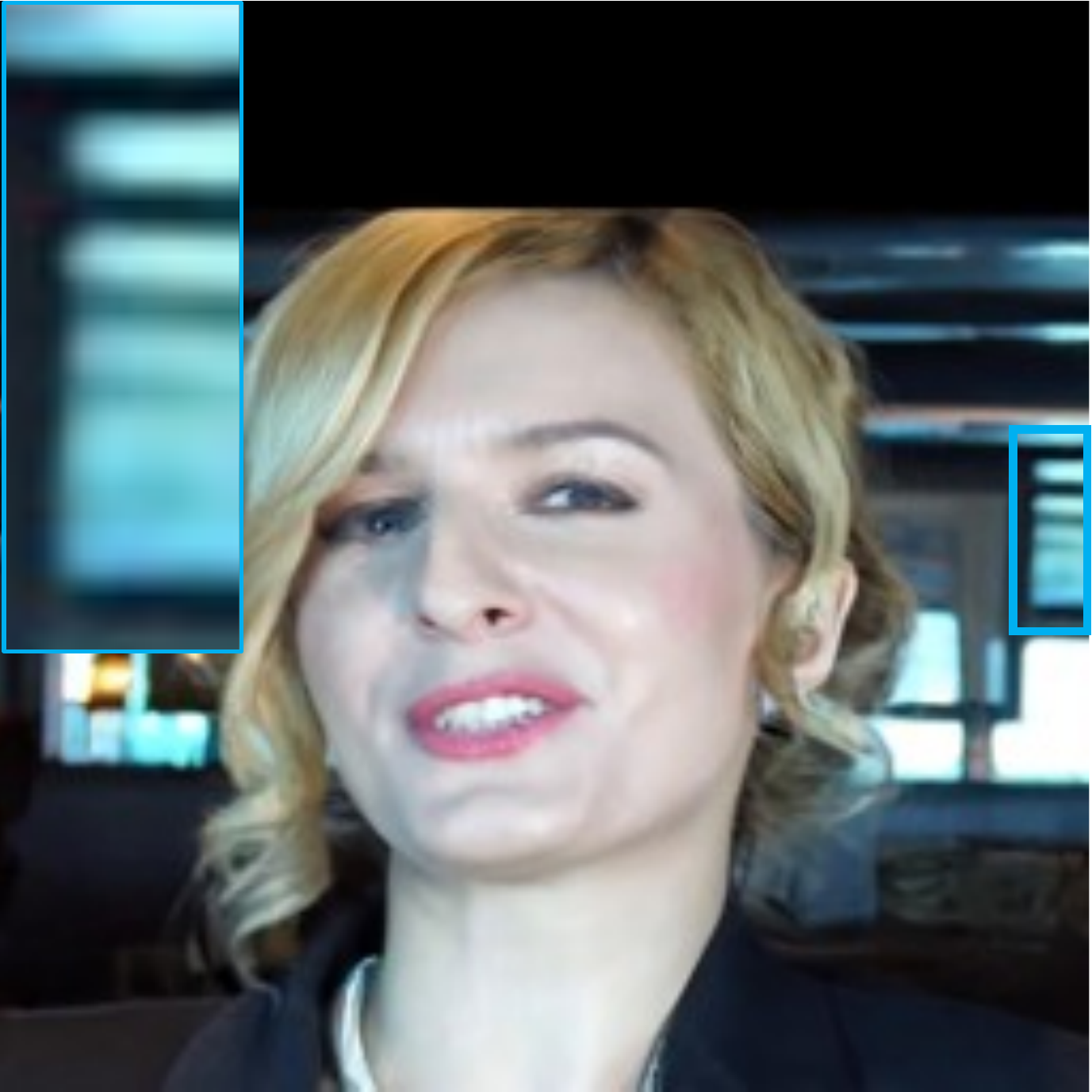}%
}
&
\raisebox{-.5\totalheight}{%
    \includegraphics[interpolate=false,width=0.190\textwidth]
    {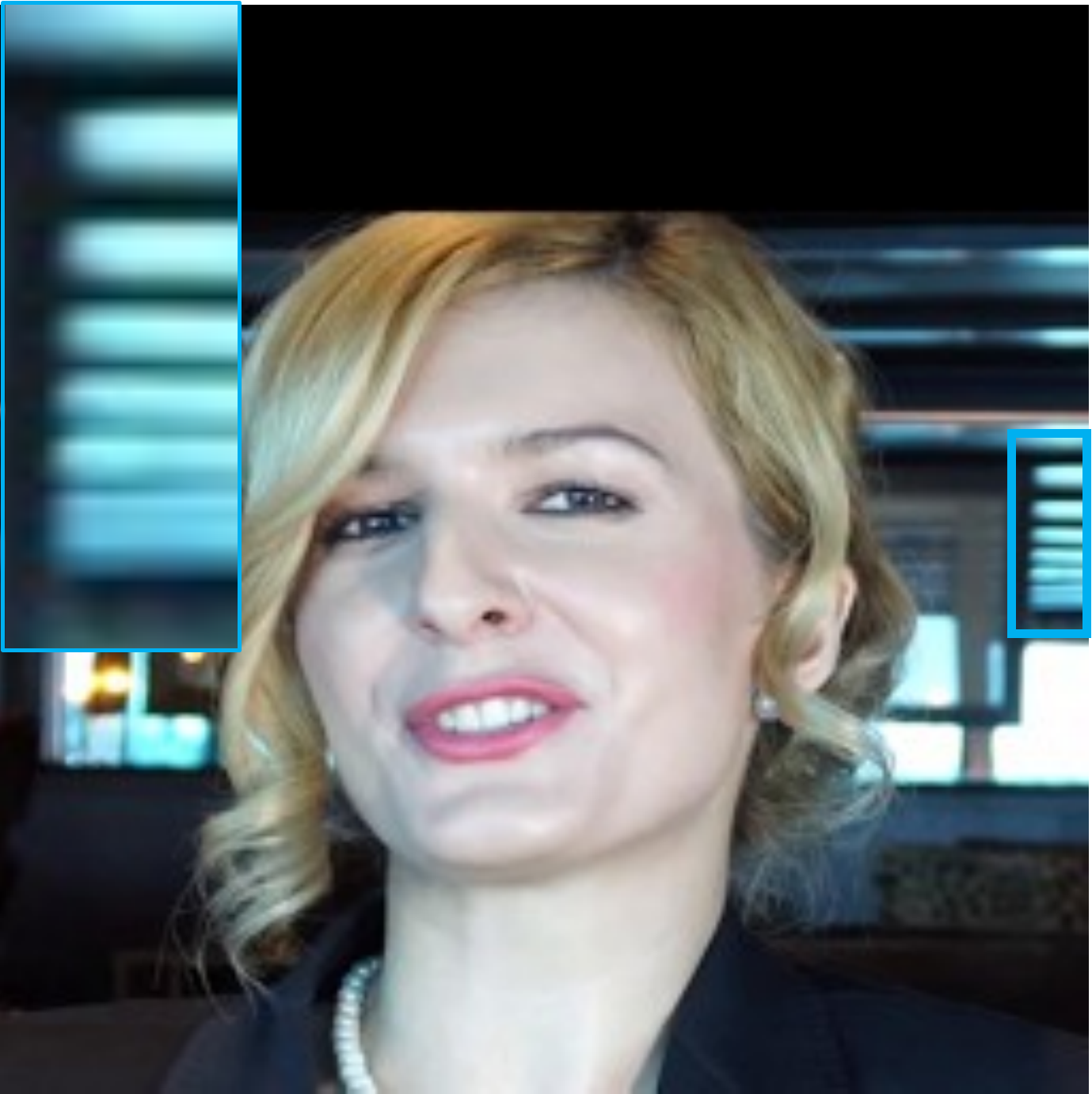}%
}
\\

\raisebox{-.5\totalheight}{%
    \makebox[0.077\textwidth][c]{%
        \shortstack{%
            \small\bfseries Noisy\\
            \small $\sigma_y=0.05$
        }%
    }%
}
&
\raisebox{-.5\totalheight}{%
    \includegraphics[interpolate=false,width=0.190\textwidth]
    {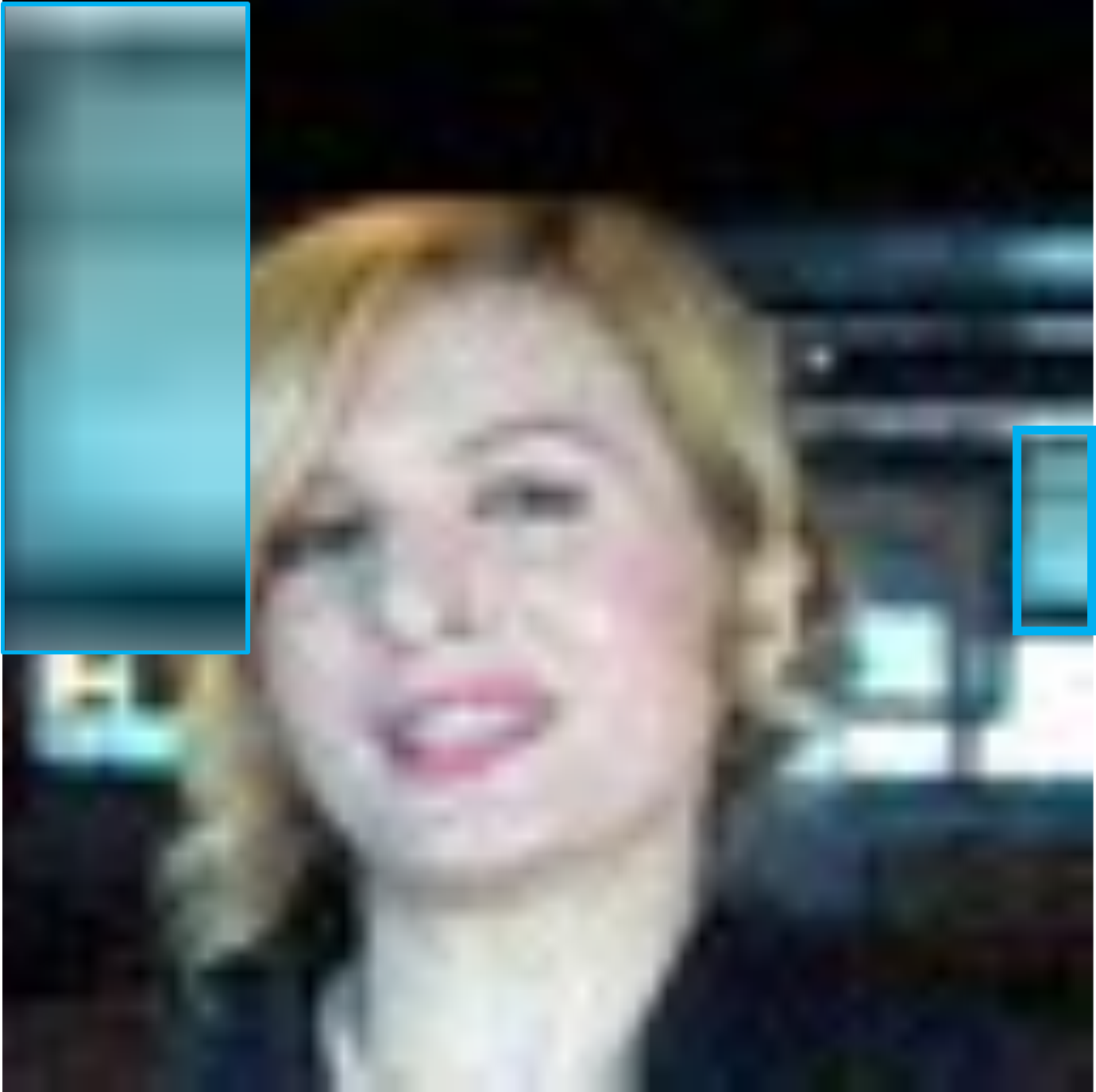}%
}
&
\raisebox{-.5\totalheight}{%
    \includegraphics[interpolate=false,width=0.190\textwidth]
    {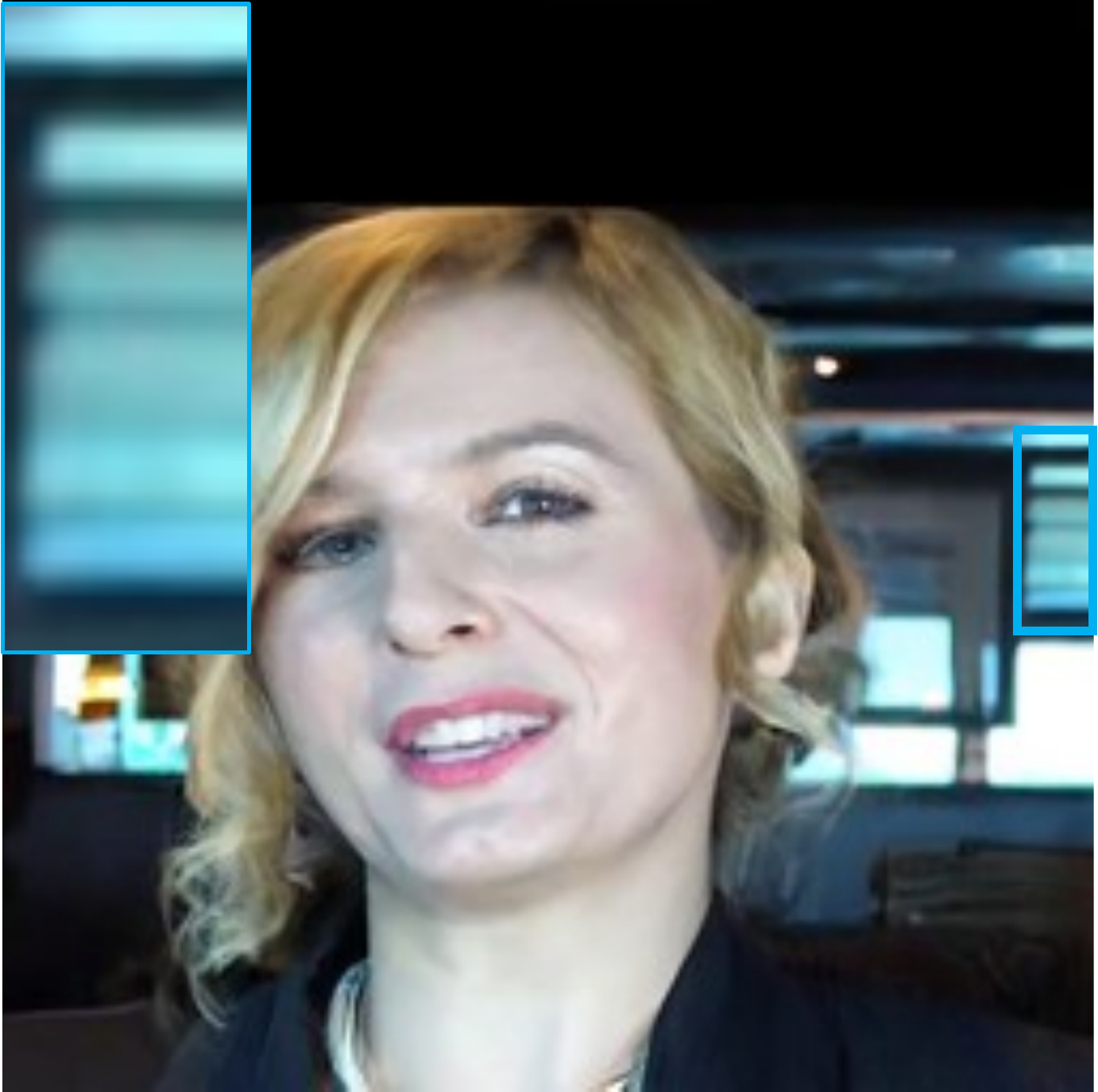}%
}
&
\raisebox{-.5\totalheight}{%
    \includegraphics[interpolate=false,width=0.190\textwidth]
    {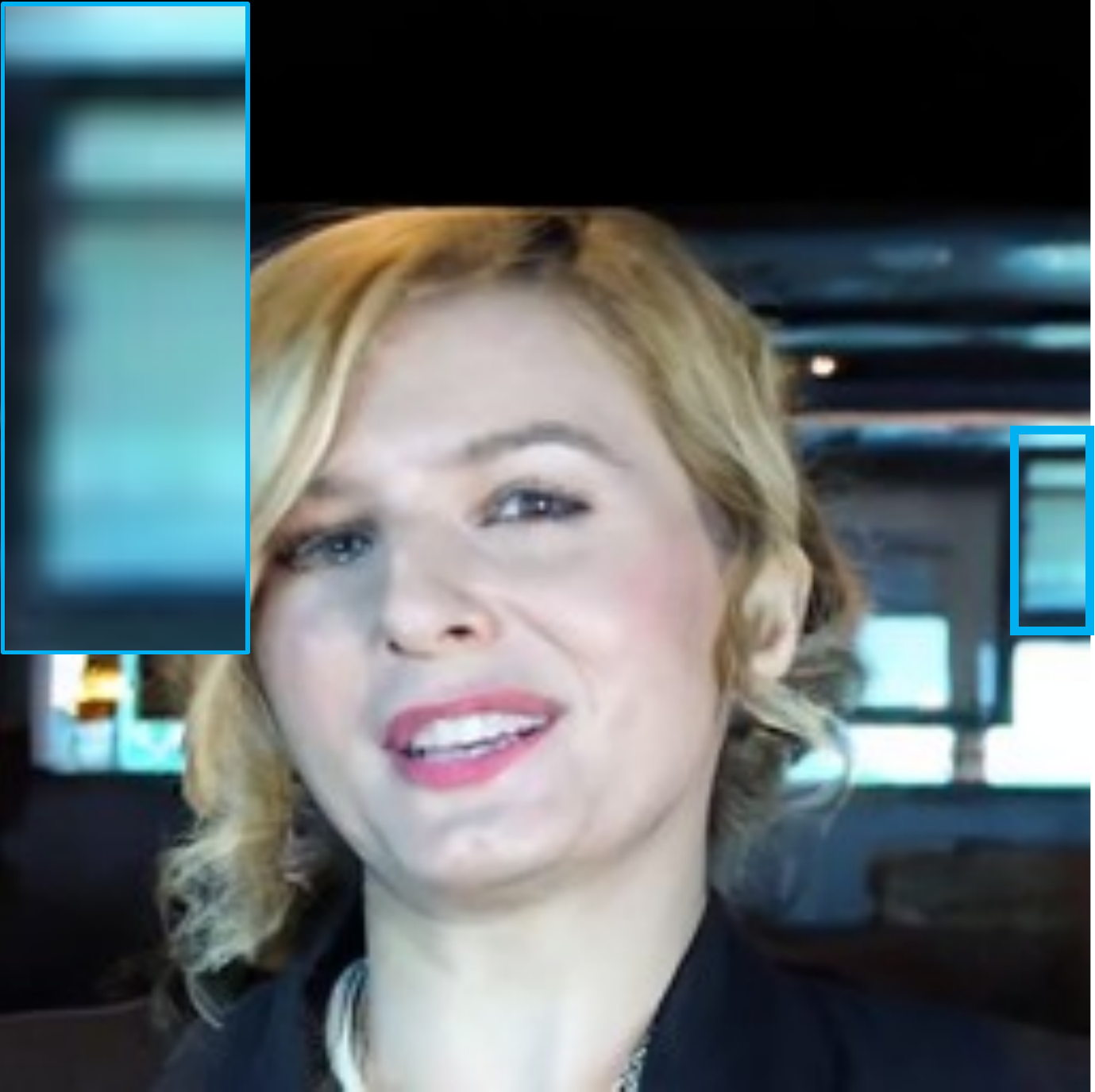}%
}
&
\raisebox{-.5\totalheight}{%
    \includegraphics[interpolate=false,width=0.190\textwidth]
    {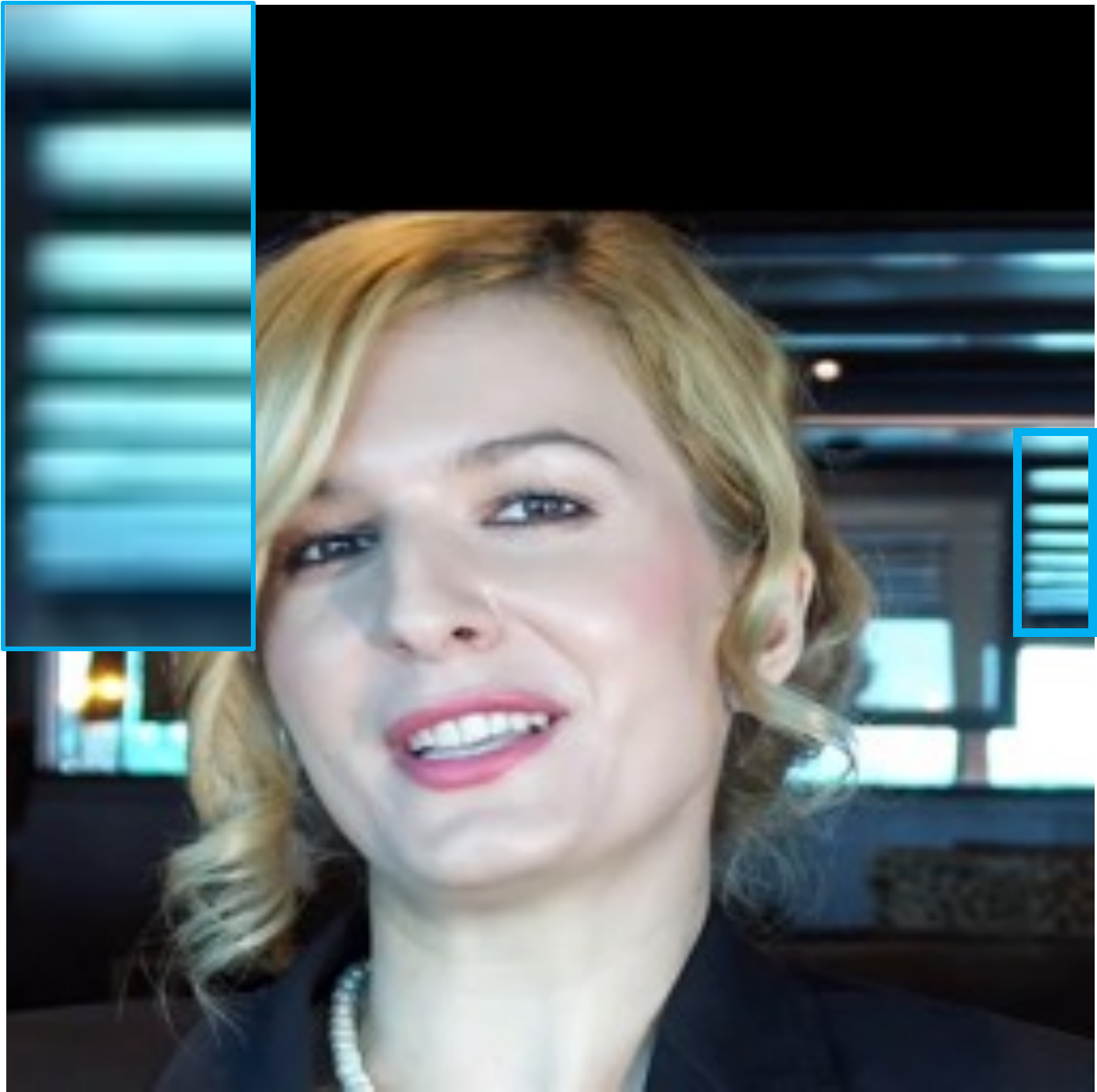}%
}
\end{tabular}

\caption{
\textbf{Visual effect of the central LR observation.}
Representative crops highlighting thin window structures.
Under noiseless degradation, conditioning on the central LR observation
preserves the window stripes more faithfully, whereas under noisy degradation
the same structures are better preserved when the noisy central observation
is omitted from the denoiser input. The enlarged LR input and ground truth are
shown for reference. The selected configurations are therefore
\emph{with} central LR conditioning for $\sigma_y=0$ and \emph{without} it for
$\sigma_y=0.05$; in both cases, the central measurement remains part of the DPS
data consistency term.
}
\label{fig:visual_central_lr}
\end{suppfigure}

\begin{suppfigure}
\centering
\setlength{\tabcolsep}{0pt}
\renewcommand{\arraystretch}{1.0}

\begin{tabular}{@{}c c c c c c@{}}
&
\small\textbf{$\mathbf{x}_{i,0}$}
&
\small\textbf{$\mathbf{x}_{i+1,0}$}
&
\small\textbf{$\mathbf{x}_{i+2,0}$}
&
\small\textbf{$\mathbf{x}_{i+3,0}$}
&
\small\textbf{$\mathbf{x}_{i+4,0}$}
\\

\raisebox{-.5\totalheight}{%
    \makebox[0.100\textwidth][c]{%
        \shortstack{%
            \small\bfseries Independent\\
            \small noise
        }%
    }%
}
&
\raisebox{-.5\totalheight}{%
    \includegraphics[interpolate=false,width=0.176\textwidth]
    {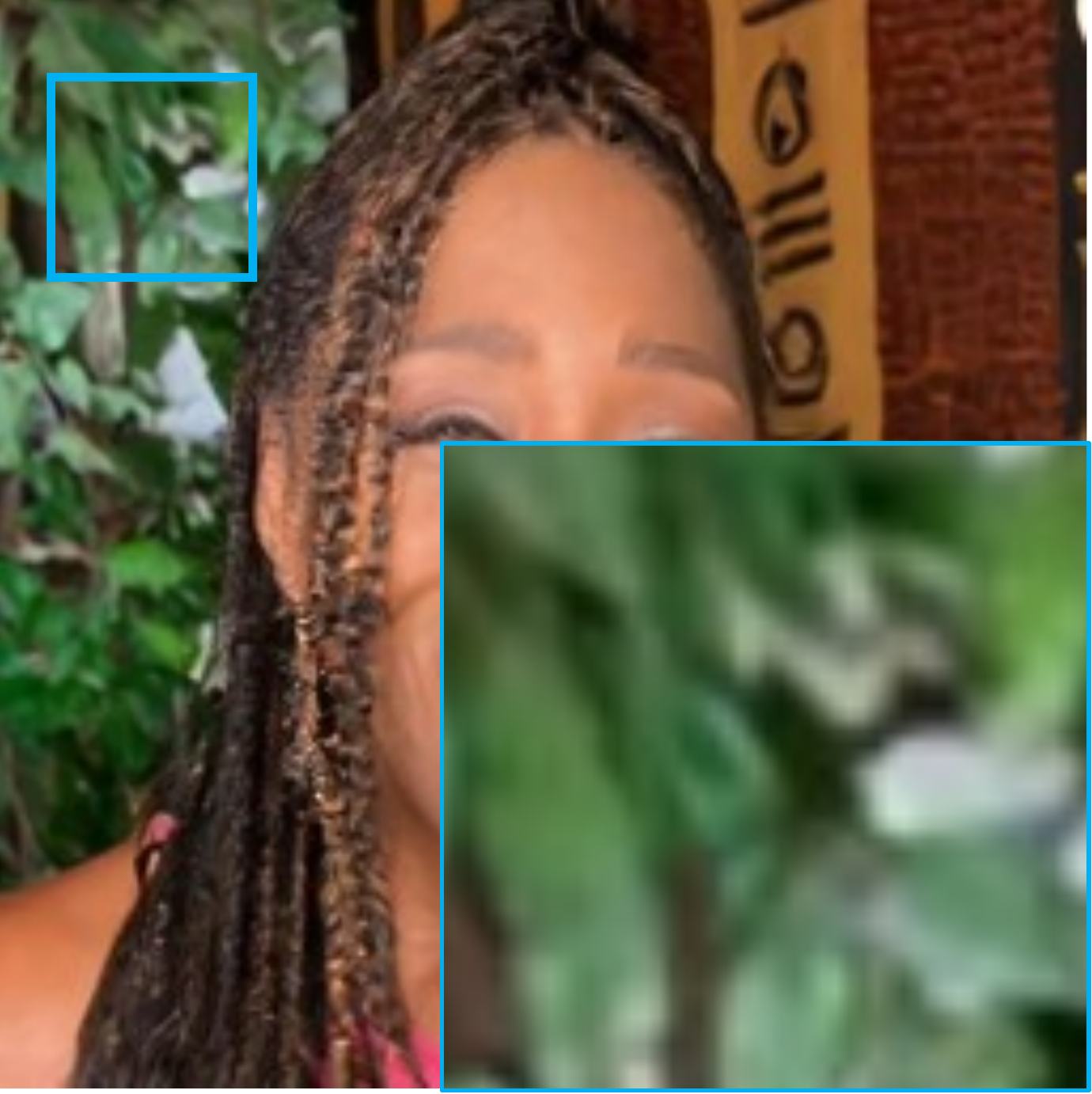}%
}
&
\raisebox{-.5\totalheight}{%
    \includegraphics[interpolate=false,width=0.176\textwidth]
    {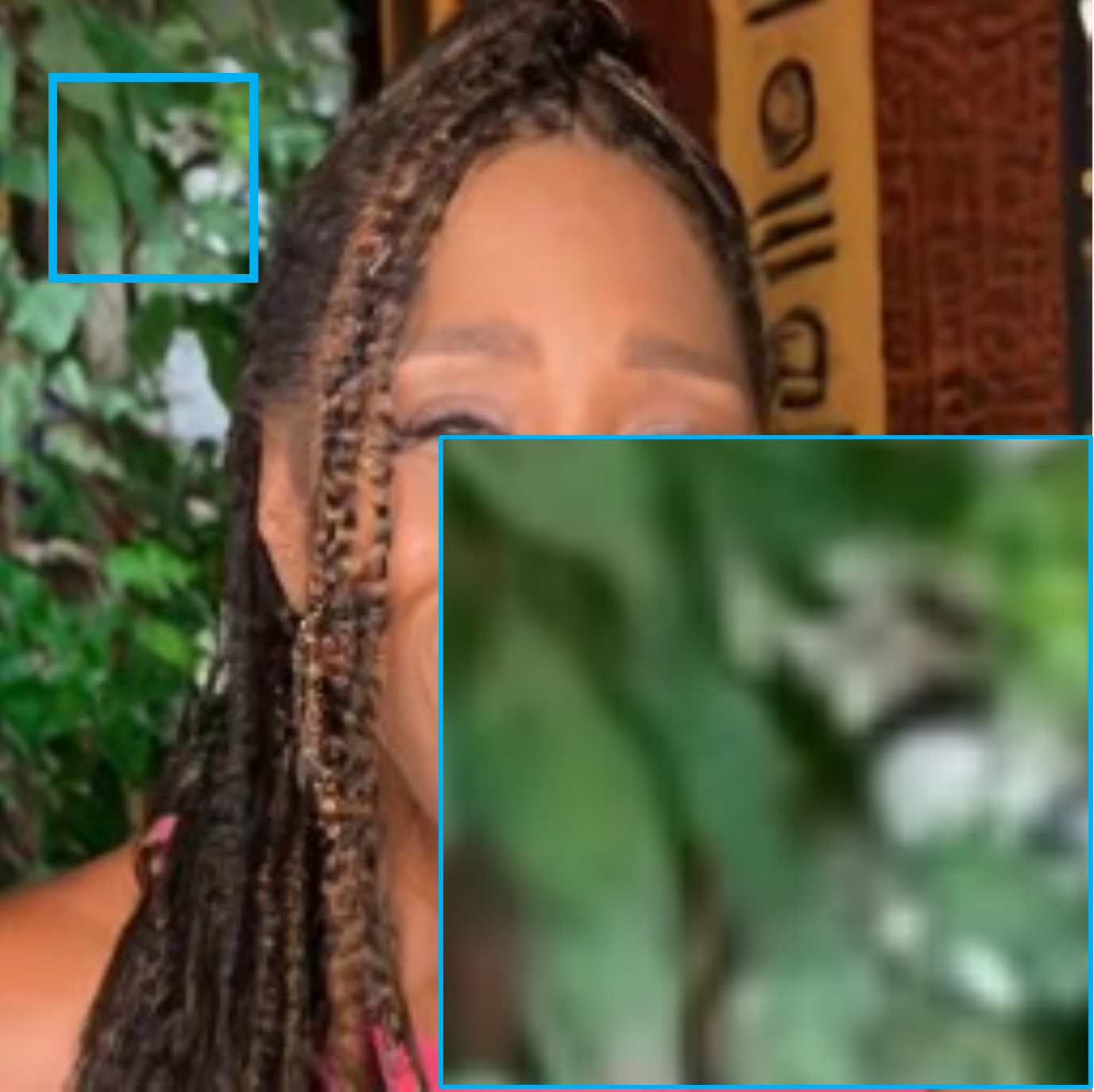}%
}
&
\raisebox{-.5\totalheight}{%
    \includegraphics[interpolate=false,width=0.176\textwidth]
    {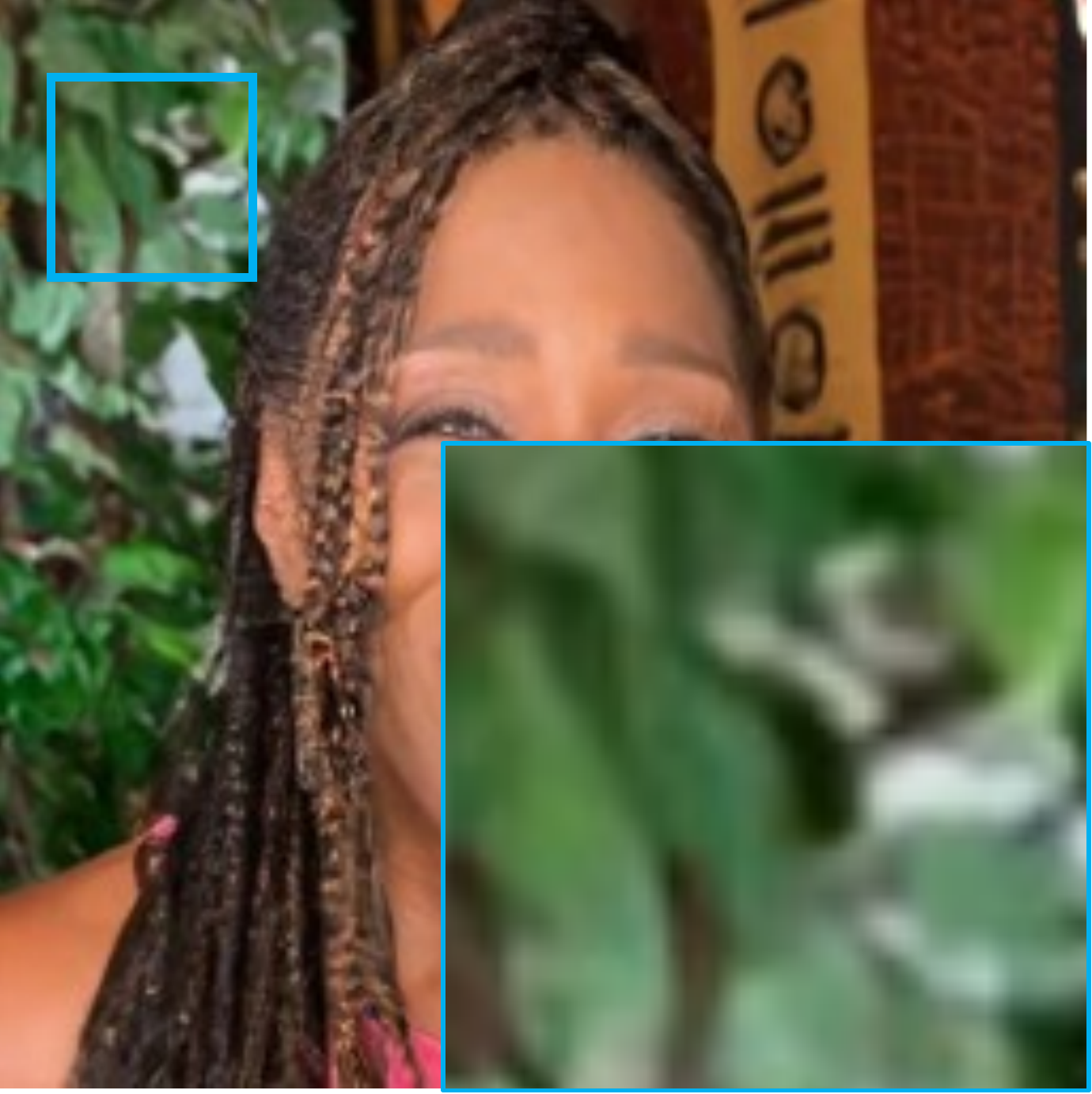}%
}
&
\raisebox{-.5\totalheight}{%
    \includegraphics[interpolate=false,width=0.176\textwidth]
    {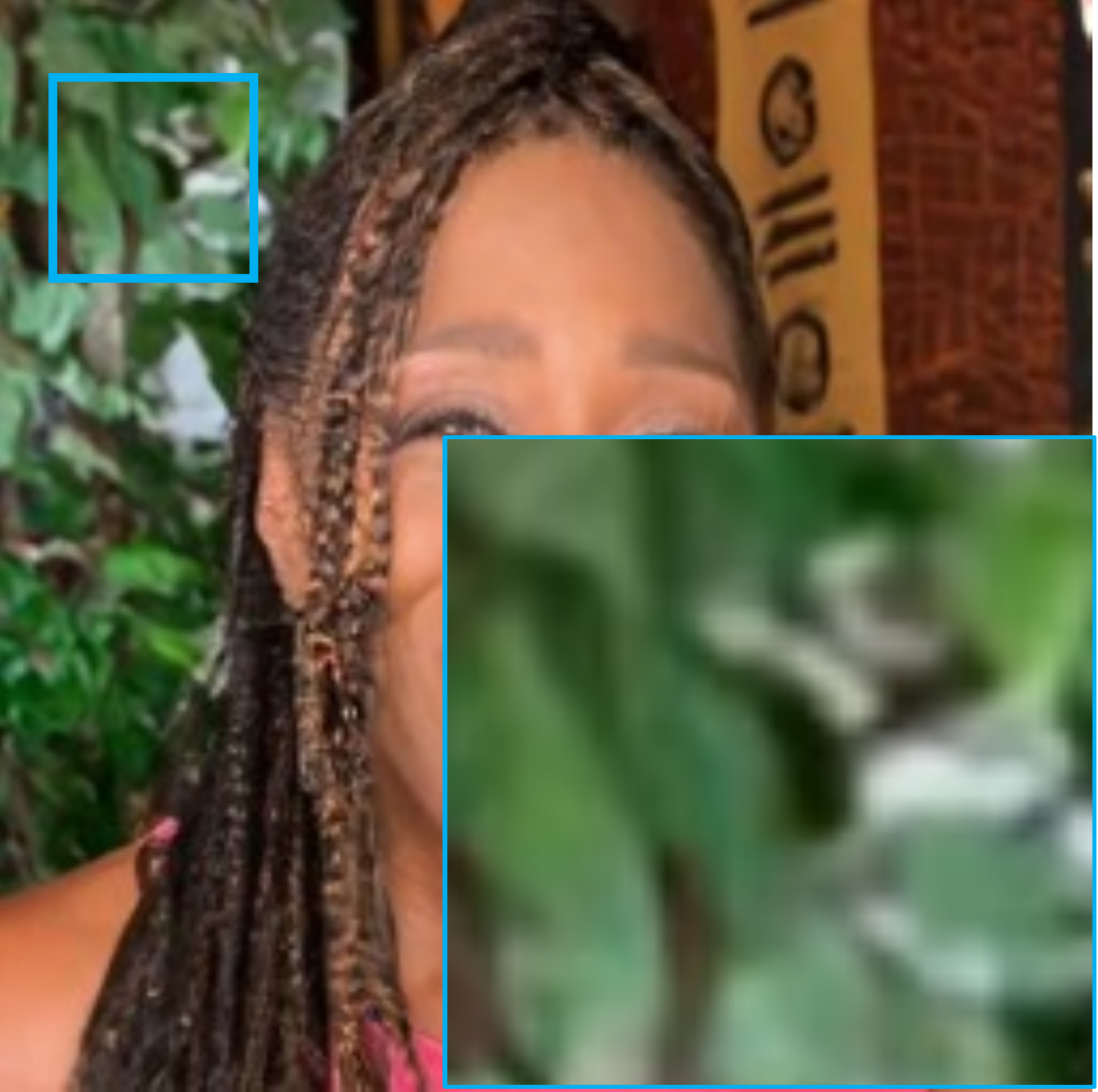}%
}
&
\raisebox{-.5\totalheight}{%
    \includegraphics[interpolate=false,width=0.176\textwidth]
    {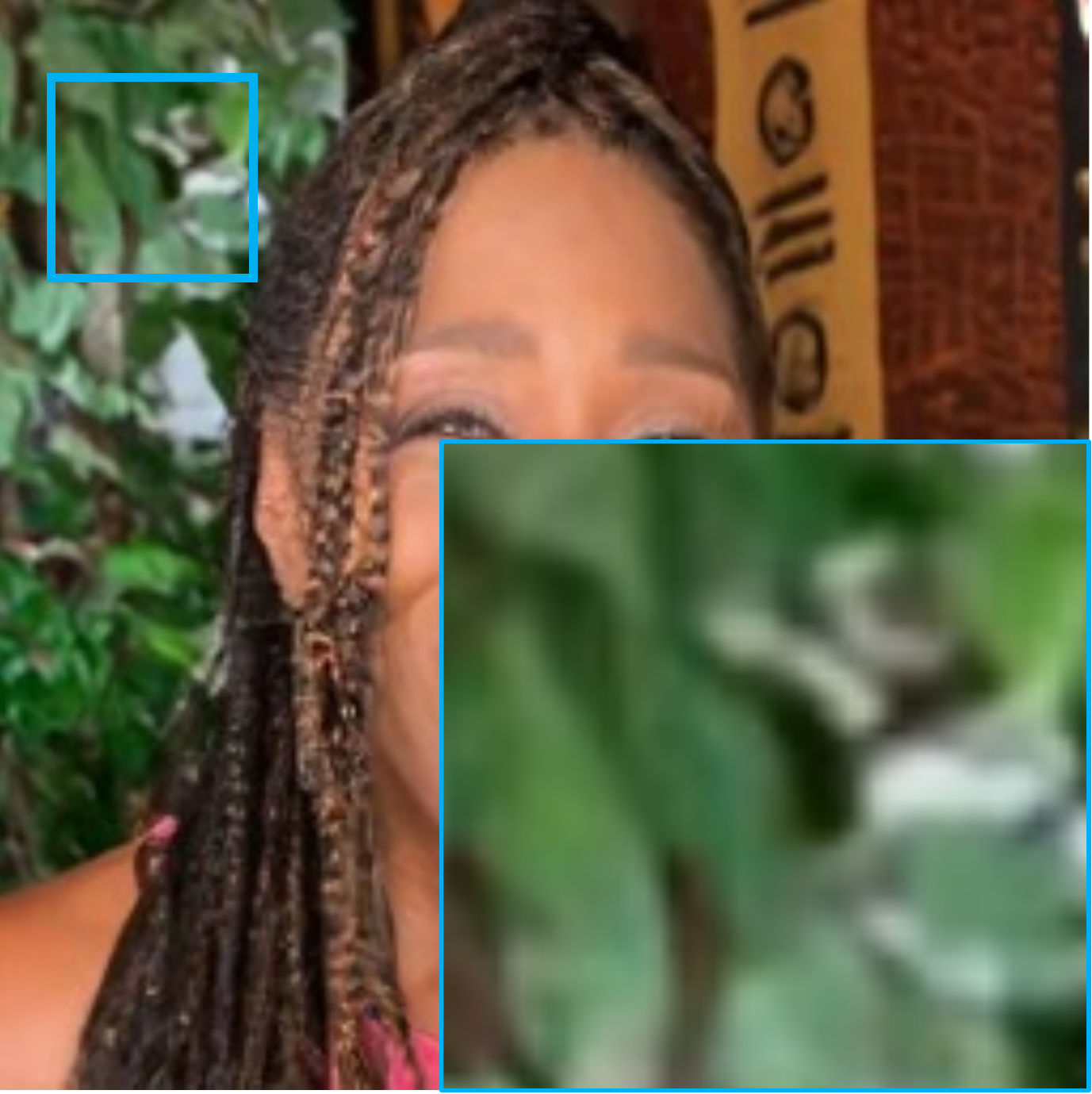}%
}
\\

\raisebox{-.5\totalheight}{%
    \makebox[0.100\textwidth][c]{%
        \shortstack{%
            \small\bfseries Shared\\
            \small noise
        }%
    }%
}
&
\raisebox{-.5\totalheight}{%
    \includegraphics[interpolate=false,width=0.176\textwidth]
    {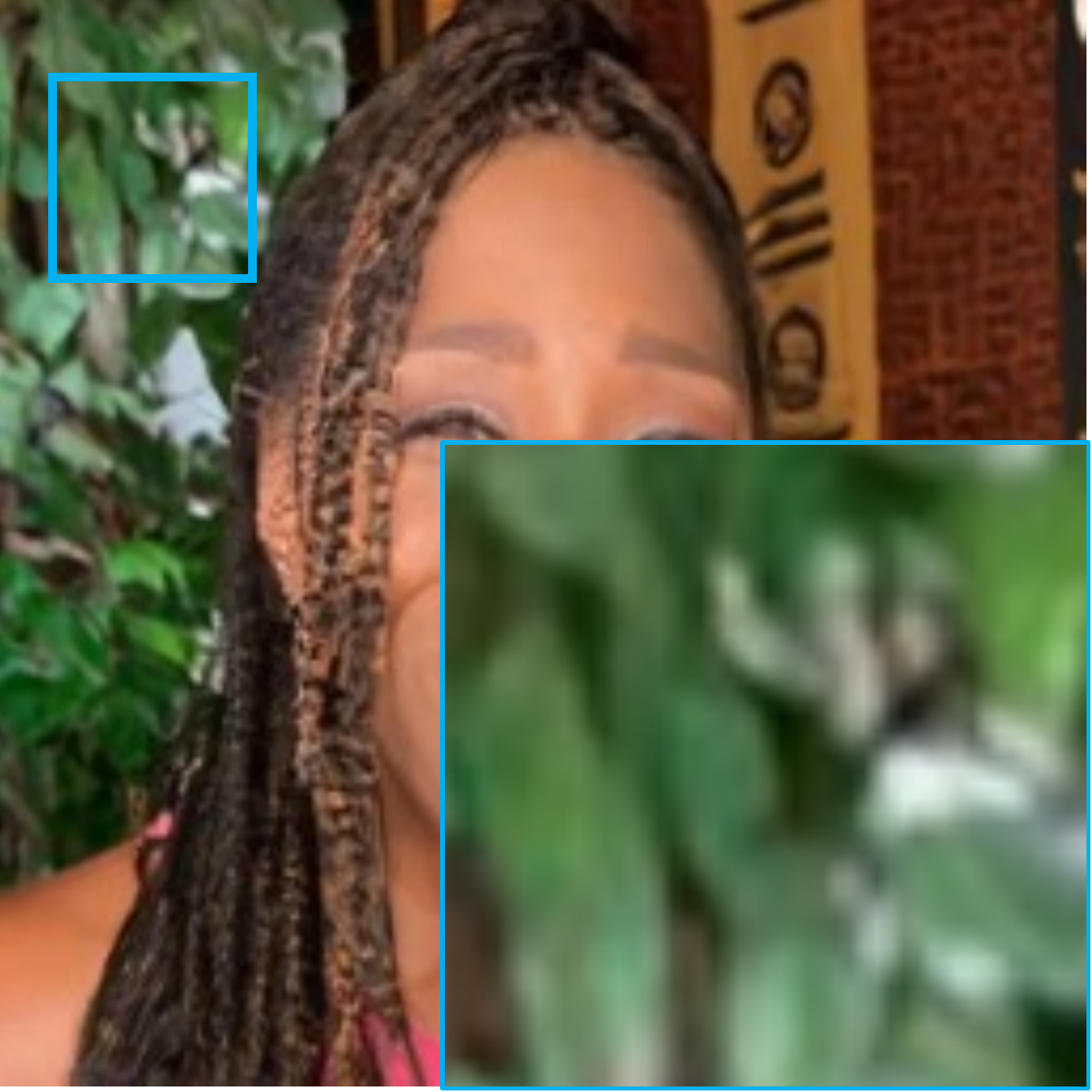}%
}
&
\raisebox{-.5\totalheight}{%
    \includegraphics[interpolate=false,width=0.176\textwidth]
    {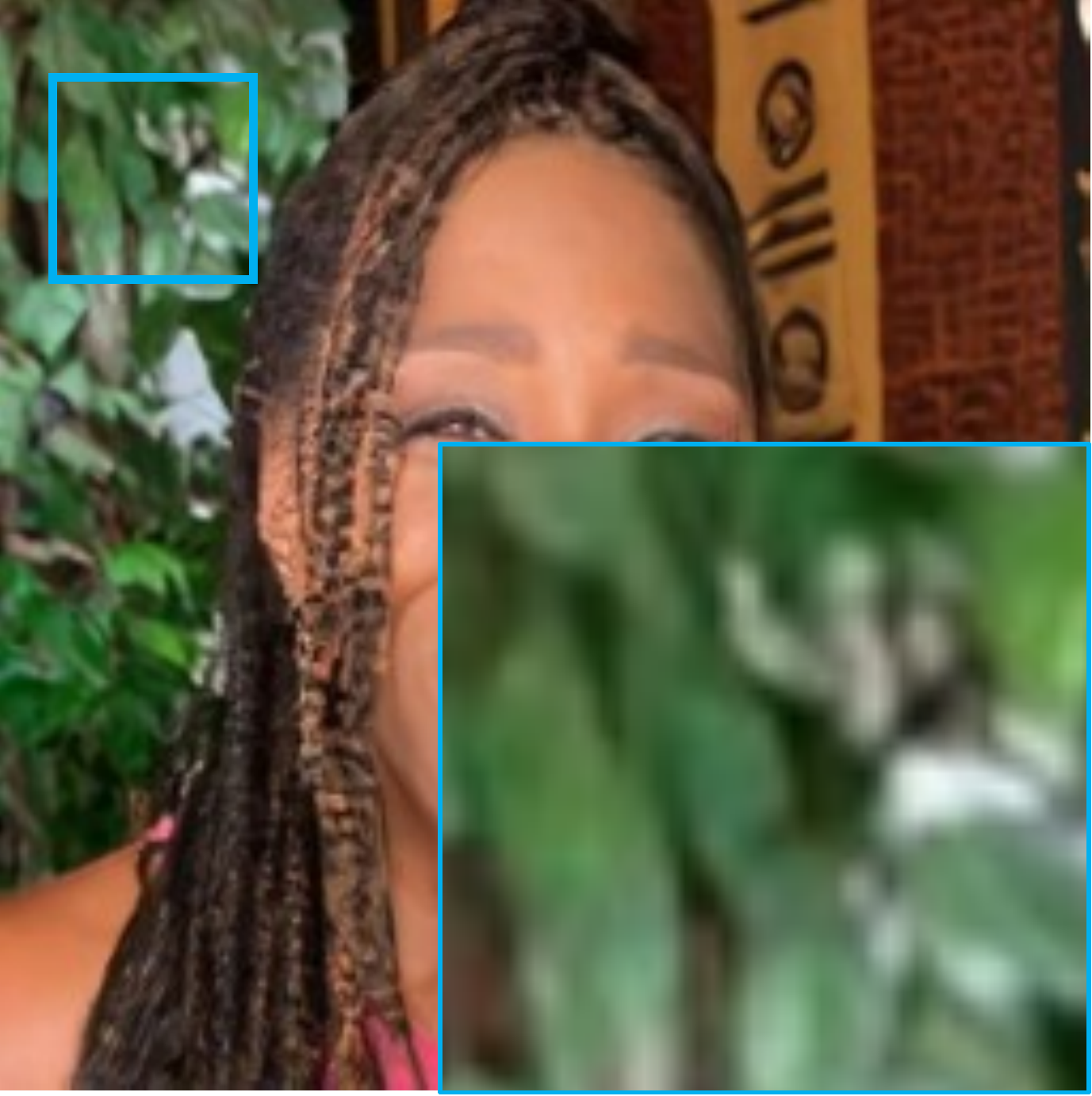}%
}
&
\raisebox{-.5\totalheight}{%
    \includegraphics[interpolate=false,width=0.176\textwidth]
    {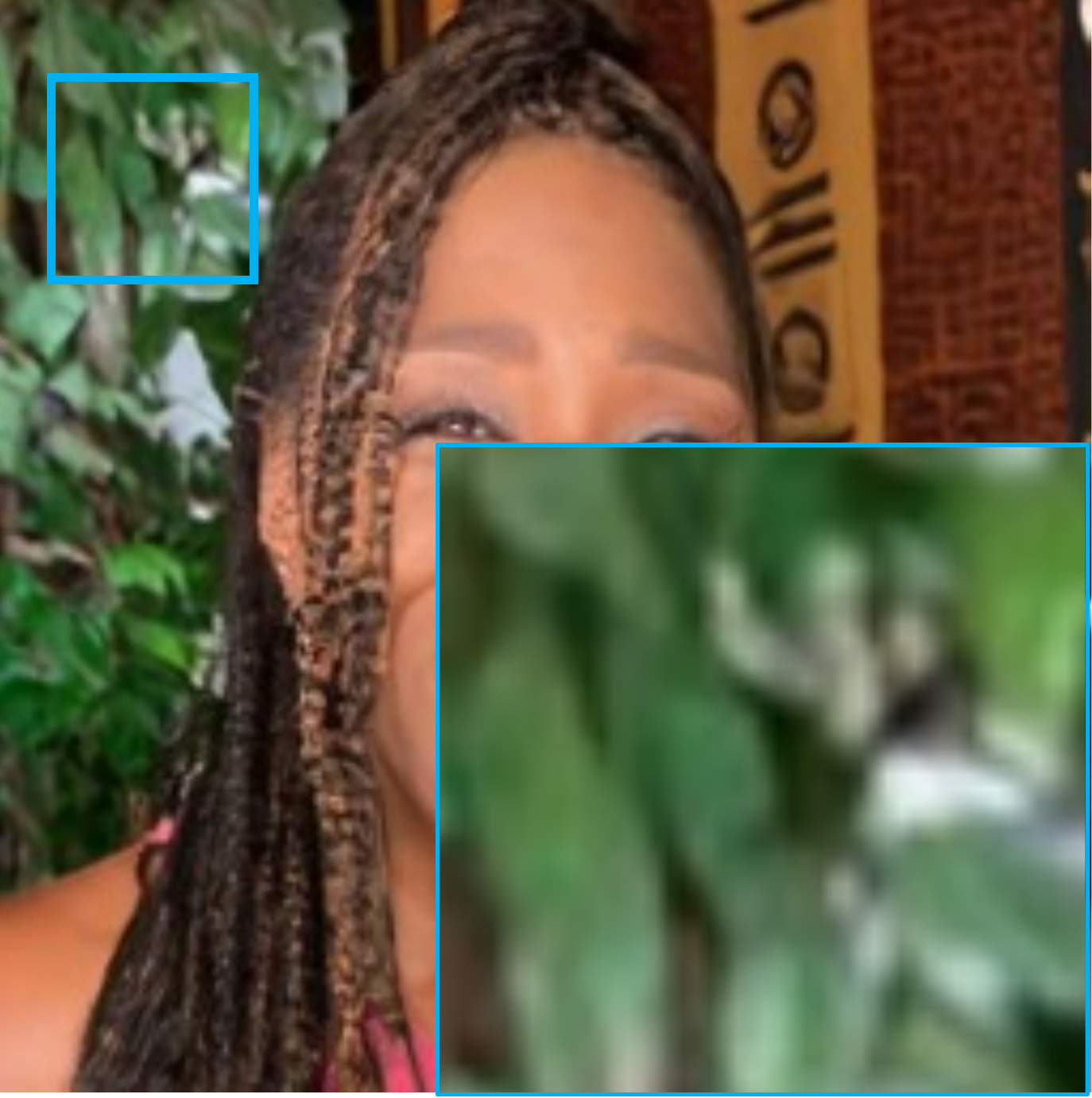}%
}
&
\raisebox{-.5\totalheight}{%
    \includegraphics[interpolate=false,width=0.176\textwidth]
    {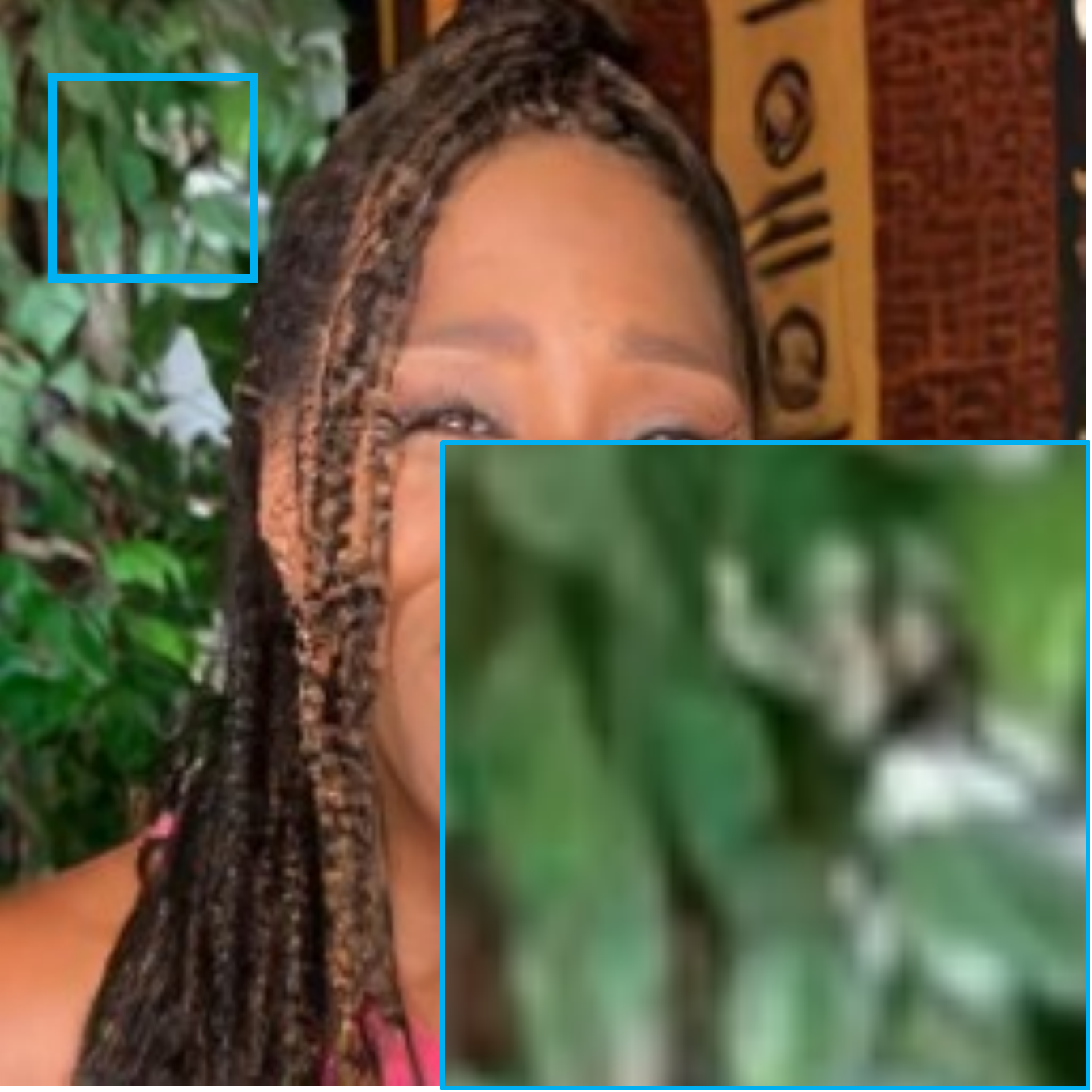}%
}
&
\raisebox{-.5\totalheight}{%
    \includegraphics[interpolate=false,width=0.176\textwidth]
    {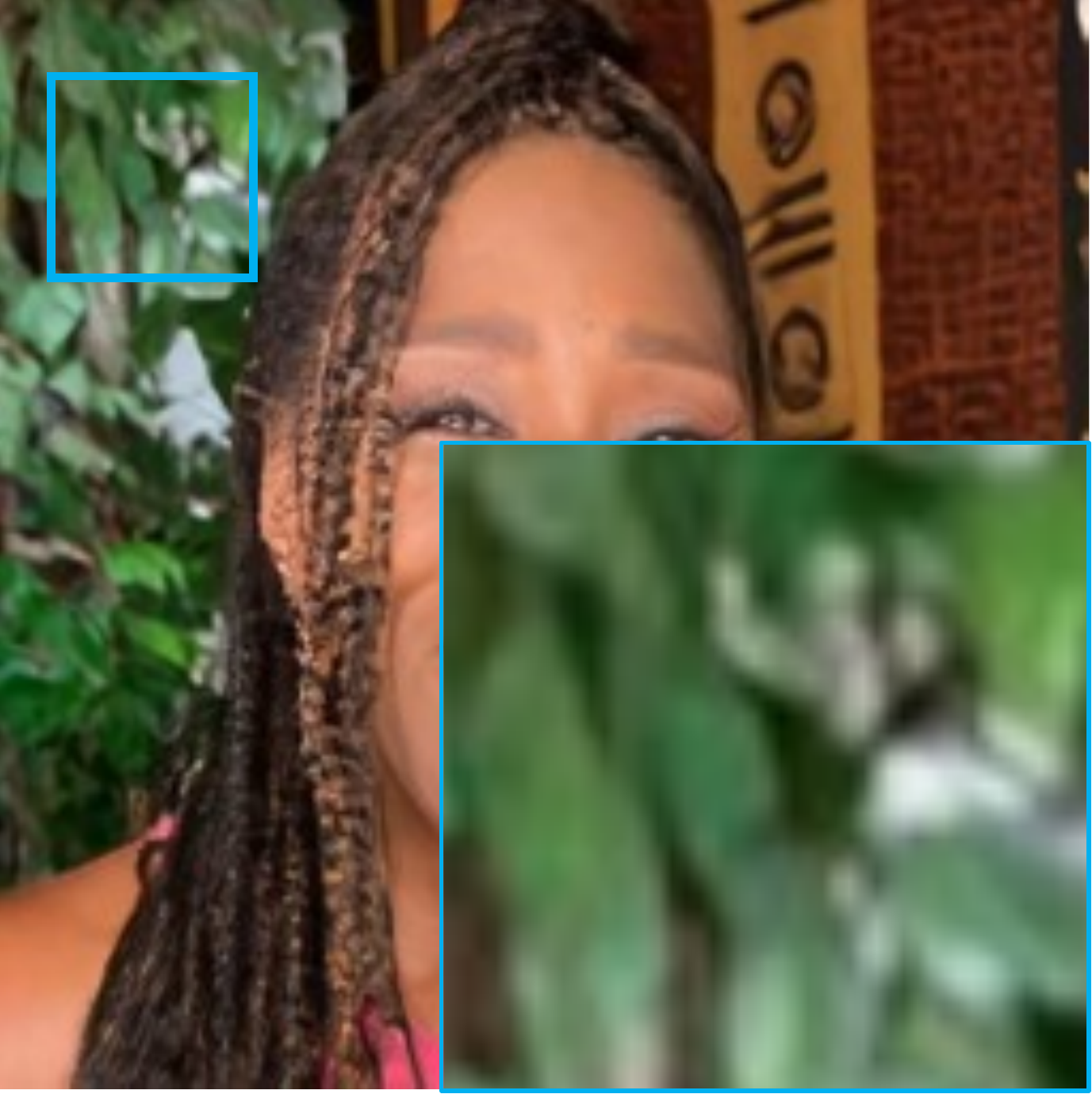}%
}
\end{tabular}

\caption{
\textbf{Visual effect of the shared noise trajectory.}
The same five consecutive frames from the noisy setting ($\sigma_y=0.05$), reconstructed using the selected architecture ($k=5$, without central LR conditioning), are shown with independent (top) and shared (bottom) diffusion noise trajectories. Independent sampling causes ambiguous fine details to vary across adjacent frames, resulting in visible flicker. Sharing the stochastic trajectory suppresses this sampling-induced variation and produces a more temporally stable sequence.
}
\label{fig:visual_shared_noise}
\end{suppfigure}

\clearpage

\begin{multicols}{2}
\raggedcolumns
\section{Additional qualitative analysis}
\label{app:additional_visuals}

Figures~\ref{fig:additional_qualitative_noiseless_1}-\ref{fig:additional_qualitative_noisy_2} extend the qualitative comparison
presented in Figure~\ref{fig:qualitative} with additional
VFHQ test examples and a broader set of competing methods. The figures report the noiseless and noisy degradation settings,
respectively, using the same degradation and evaluation protocol as in the main paper.
\end{multicols}

\begin{suppfigure}
\centering
\setlength{\tabcolsep}{0.7pt}
\renewcommand{\arraystretch}{0.92}

\begin{tabular}{@{}c@{\hspace{1pt}}c@{\hspace{1pt}}c@{\hspace{1pt}}c@{\hspace{1pt}}c@{}}
&
\small\textbf{LR input}
&
\small\textbf{Baseline}
&
\small\textbf{\method{}}
&
\small\textbf{Ground truth}
\\

\rotatebox{90}{\small\textbf{SVI-Diffusion}}
&
\includegraphics[interpolate=false,width=0.243\textwidth]{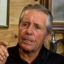}
&
\includegraphics[interpolate=false,width=0.243\textwidth]{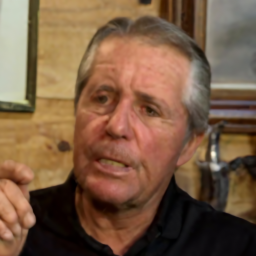}
&
\includegraphics[interpolate=false,width=0.243\textwidth]{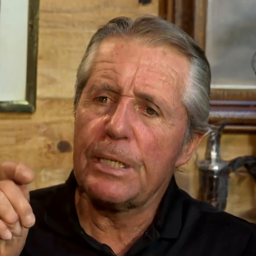}
&
\includegraphics[interpolate=false,width=0.243\textwidth]{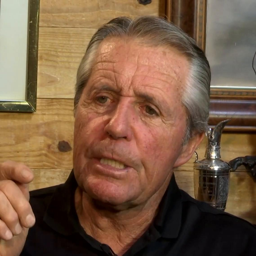}
\\

\rotatebox{90}{\small\textbf{Upscale-A-Video}}
&
\includegraphics[interpolate=false,width=0.243\textwidth]{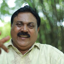}
&
\includegraphics[interpolate=false,width=0.243\textwidth]{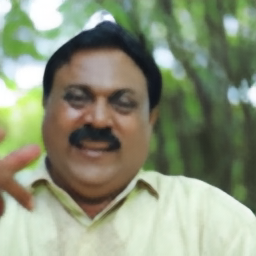}
&
\includegraphics[interpolate=false,width=0.243\textwidth]{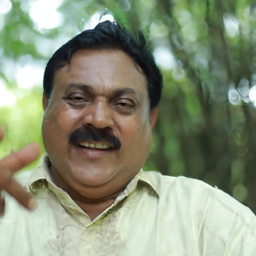}
&
\includegraphics[interpolate=false,width=0.243\textwidth]{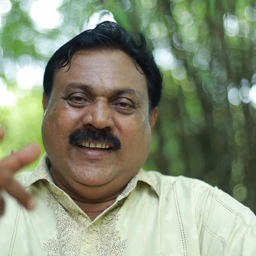}
\\

\rotatebox{90}{\small\textbf{VISION-XL}}
&
\includegraphics[interpolate=false,width=0.243\textwidth]{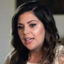}
&
\includegraphics[interpolate=false,width=0.243\textwidth]{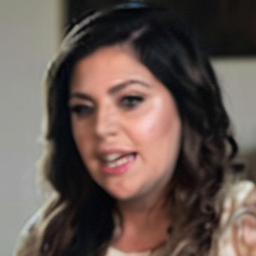}
&
\includegraphics[interpolate=false,width=0.243\textwidth]{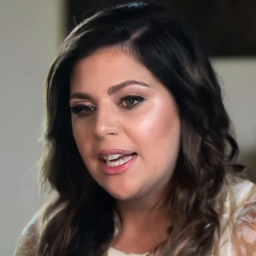}
&
\includegraphics[interpolate=false,width=0.243\textwidth]{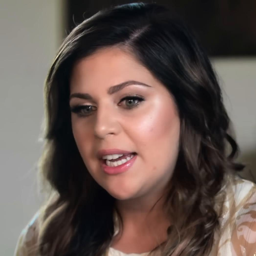}
\\

\rotatebox{90}{\small\textbf{UltraVSR}}
&
\includegraphics[interpolate=false,width=0.243\textwidth]{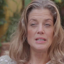}
&
\includegraphics[interpolate=false,width=0.243\textwidth]{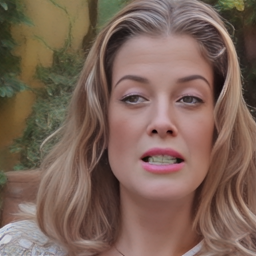}
&
\includegraphics[interpolate=false,width=0.243\textwidth]{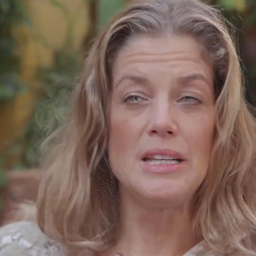}
&
\includegraphics[interpolate=false,width=0.243\textwidth]{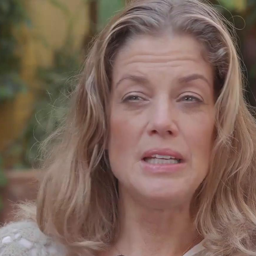}
\\

\end{tabular}

\caption{
\textbf{Extended qualitative comparison for Figure~\ref{fig:qualitative}
under noiseless degradation (Part I).}
Additional VFHQ test examples for $\times4$ VSR with $\sigma_y=0$.
Each row compares the enlarged LR input, the corresponding
baseline reconstruction, LoCoVSR, and the ground truth for the same target
frame. The examples extend the qualitative results in
Figure~\ref{fig:qualitative} to additional competing methods
and illustrate differences in the recovery of fine facial structures and
high-frequency details.
}
\label{fig:additional_qualitative_noiseless_1}
\end{suppfigure}

\clearpage

\begin{suppfigure}
\centering
\setlength{\tabcolsep}{0.7pt}
\renewcommand{\arraystretch}{0.92}

\begin{tabular}{@{}c@{\hspace{1pt}}c@{\hspace{1pt}}c@{\hspace{1pt}}c@{\hspace{1pt}}c@{}}
&
\small\textbf{LR input}
&
\small\textbf{Baseline}
&
\small\textbf{\method{}}
&
\small\textbf{Ground truth}
\\

\rotatebox{90}{\small\textbf{PS-SR}}
&
\includegraphics[interpolate=false,width=0.243\textwidth]{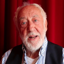}
&
\includegraphics[interpolate=false,width=0.243\textwidth]{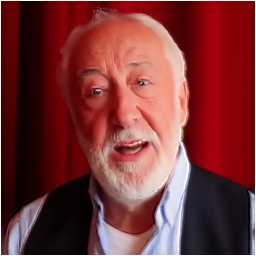}
&
\includegraphics[interpolate=false,width=0.243\textwidth]{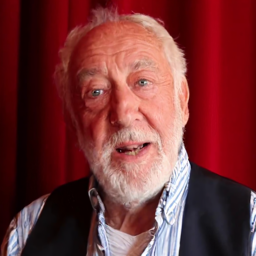}
&
\includegraphics[interpolate=false,width=0.243\textwidth]{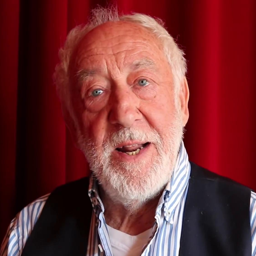}
\\

\rotatebox{90}{\small\textbf{BasicVSR++}}
&
\includegraphics[interpolate=false,width=0.243\textwidth]{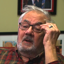}
&
\includegraphics[interpolate=false,width=0.243\textwidth]{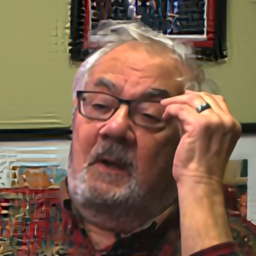}
&
\includegraphics[interpolate=false,width=0.243\textwidth]{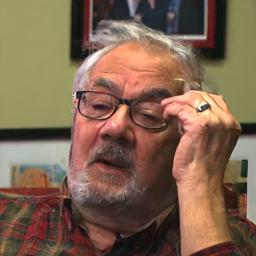}
&
\includegraphics[interpolate=false,width=0.243\textwidth]{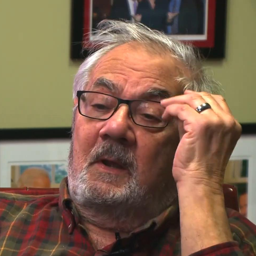}
\\

\rotatebox{90}{\small\textbf{StableVSR}}
&
\includegraphics[interpolate=false,width=0.243\textwidth]{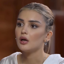}
&
\includegraphics[interpolate=false,width=0.243\textwidth]{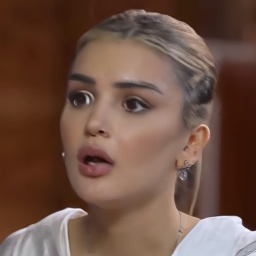}
&
\includegraphics[interpolate=false,width=0.243\textwidth]{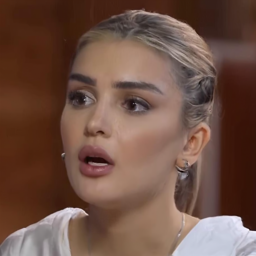}
&
\includegraphics[interpolate=false,width=0.243\textwidth]{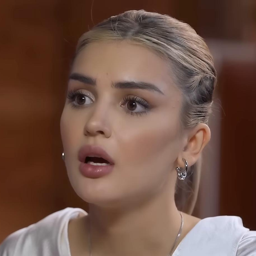}
\\

\end{tabular}

\caption{
\textbf{Extended qualitative comparison for Figure~\ref{fig:qualitative}
under noiseless degradation (Part II).}
Continuation of Figure~\ref{fig:additional_qualitative_noiseless_1} with additional competing methods. 
}
\label{fig:additional_qualitative_noiseless_2}
\end{suppfigure}

\begin{multicols}{2}
\raggedcolumns
\paragraph{Noiseless degradation.}
Figures~\ref{fig:additional_qualitative_noiseless_1} and~\ref{fig:additional_qualitative_noiseless_2} show that the main
differences are concentrated in fine facial structures.
Several competing methods recover a visually plausible face but either
smooth weak high-frequency details or introduce local structures that
deviate from the ground truth. In contrast, \method{} more consistently
preserves facial contours, eye and mouth structure, wrinkles, hair, and
other subtle textures while remaining close to the reference. The
examples also illustrate that increased apparent sharpness alone does
not necessarily correspond to faithful reconstruction: some baseline
outputs contain pronounced or synthetic local detail, whereas \method{}
better preserves the structure supported by the underlying observation.

\paragraph{Noisy degradation.}
The differences become more pronounced in
Figures~\ref{fig:additional_qualitative_noisy_1} and~\ref{fig:additional_qualitative_noisy_2}. Under additive noise,
several baselines either retain residual artifacts or suppress weak
high-frequency structure, leading to noisy, oversmoothed, or locally
distorted reconstructions. \method{}, in contrast,
operates directly on the noisy measurements: the central observation is
enforced through per-frame posterior guidance, while neighboring LR
frames provide complementary temporal evidence. This combination allows
fine facial structures to remain visible while avoiding the need for a
separate denoising stage.

Taken together, these additional examples are consistent with the
quantitative results in Tab.~\ref{tab:main-test}: the advantage of
\method{} is not merely increased sharpness, but the recovery of fine
structure that remains faithful to the observed content across both
degradation regimes.
\end{multicols}

\clearpage

\begin{suppfigure}
\centering
\setlength{\tabcolsep}{0.7pt}
\renewcommand{\arraystretch}{0.92}

\begin{tabular}{@{}c@{\hspace{1pt}}c@{\hspace{1pt}}c@{\hspace{1pt}}c@{\hspace{1pt}}c@{}}
&
\small\textbf{LR input}
&
\small\textbf{Baseline}
&
\small\textbf{\method{}}
&
\small\textbf{Ground truth}
\\

\rotatebox{90}{\small\textbf{SVI-Diffusion}}
&
\includegraphics[interpolate=false,width=0.243\textwidth]{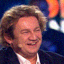}
&
\includegraphics[interpolate=false,width=0.243\textwidth]{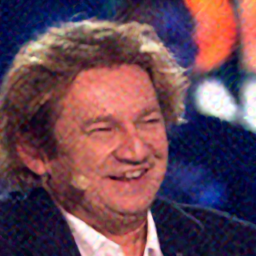}
&
\includegraphics[interpolate=false,width=0.243\textwidth]{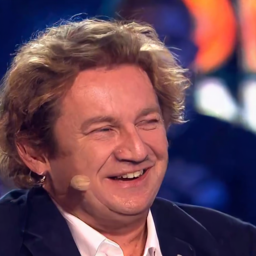}
&
\includegraphics[interpolate=false,width=0.243\textwidth]{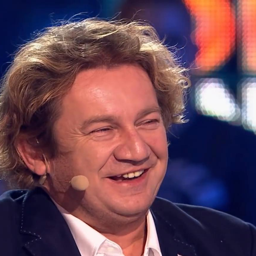}
\\

\rotatebox{90}{\small\textbf{Upscale-A-Video}}
&
\includegraphics[interpolate=false,width=0.243\textwidth]{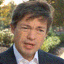}
&
\includegraphics[interpolate=false,width=0.243\textwidth]{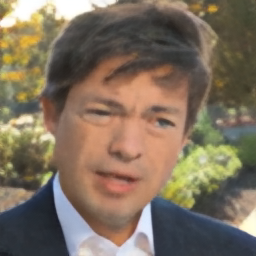}
&
\includegraphics[interpolate=false,width=0.243\textwidth]{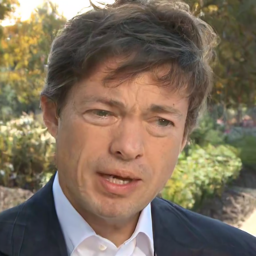}
&
\includegraphics[interpolate=false,width=0.243\textwidth]{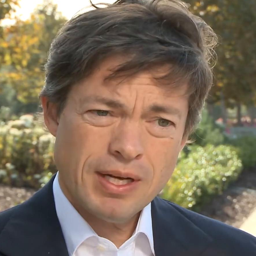}
\\

\rotatebox{90}{\small\textbf{VISION-XL}}
&
\includegraphics[interpolate=false,width=0.243\textwidth]{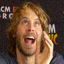}
&
\includegraphics[interpolate=false,width=0.243\textwidth]{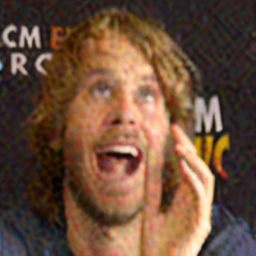}
&
\includegraphics[interpolate=false,width=0.243\textwidth]{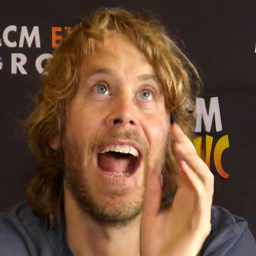}
&
\includegraphics[interpolate=false,width=0.243\textwidth]{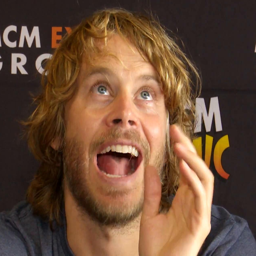}
\\

\rotatebox{90}{\small\textbf{UltraVSR}}
&
\includegraphics[interpolate=false,width=0.243\textwidth]{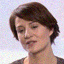}
&
\includegraphics[interpolate=false,width=0.243\textwidth]{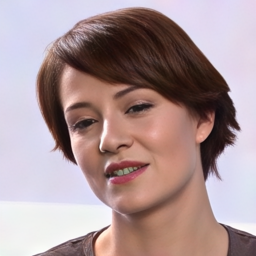}
&
\includegraphics[interpolate=false,width=0.243\textwidth]{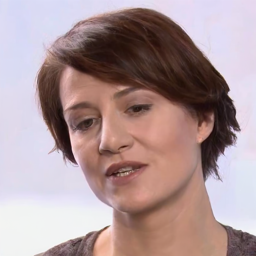}
&
\includegraphics[interpolate=false,width=0.243\textwidth]{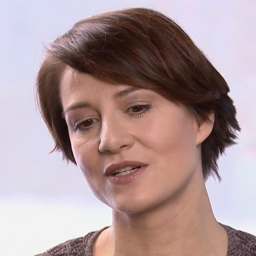}

\end{tabular}

\caption{
\textbf{Extended qualitative comparison for Figure~\ref{fig:qualitative}
under noisy degradation (Part I).}
Additional VFHQ test examples for $\times4$ VSR with additive Gaussian
noise ($\sigma_y=0.05$). Each row compares the enlarged noisy
LR input, the corresponding baseline reconstruction, LoCoVSR, and the
ground truth for the same target frame. The examples extend the qualitative
results in Figure~\ref{fig:qualitative} to additional competing
methods, and highlight the recovery of fine
structures under measurement noise. 
}
\label{fig:additional_qualitative_noisy_1}
\end{suppfigure}

\clearpage

\begin{suppfigure}
\centering
\setlength{\tabcolsep}{0.7pt}
\renewcommand{\arraystretch}{0.7}

\begin{tabular}{@{}c@{\hspace{1pt}}c@{\hspace{1pt}}c@{\hspace{1pt}}c@{\hspace{1pt}}c@{}}
&
\small\textbf{LR input}
&
\small\textbf{Baseline}
&
\small\textbf{\method{}}
&
\small\textbf{Ground truth}
\\

\rotatebox{90}{\small\textbf{PS-SR}}
&
\includegraphics[interpolate=false,width=0.239\textwidth]{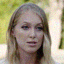}
&
\includegraphics[interpolate=false,width=0.239\textwidth]{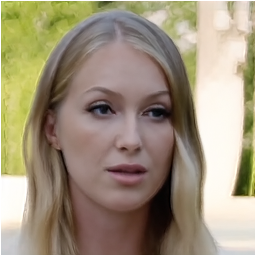}
&
\includegraphics[interpolate=false,width=0.239\textwidth]{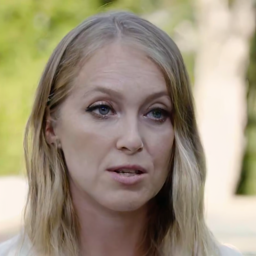}
&
\includegraphics[interpolate=false,width=0.239\textwidth]{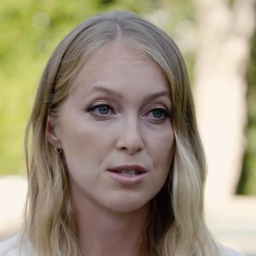}
\\

\rotatebox{90}{\small\textbf{BasicVSR++}}
&
\includegraphics[interpolate=false,width=0.239\textwidth]{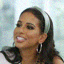}
&
\includegraphics[interpolate=false,width=0.239\textwidth]{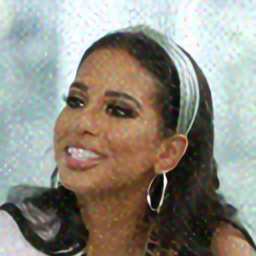}
&
\includegraphics[interpolate=false,width=0.239\textwidth]{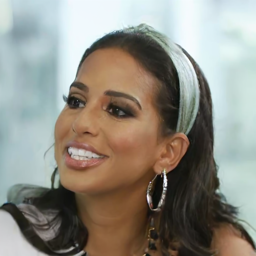}
&
\includegraphics[interpolate=false,width=0.239\textwidth]{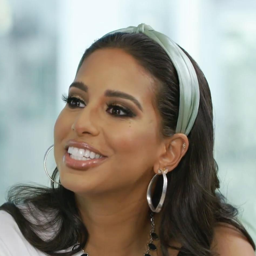}
\\

\rotatebox{90}{\small\textbf{BasicVSR++ +BM3D}}
&
\includegraphics[interpolate=false,width=0.239\textwidth]{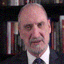}
&
\includegraphics[interpolate=false,width=0.239\textwidth]{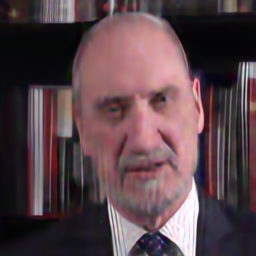}
&
\includegraphics[interpolate=false,width=0.239\textwidth]{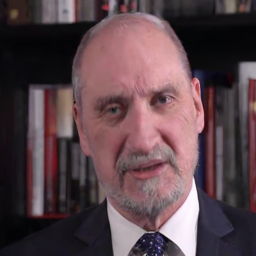}
&
\includegraphics[interpolate=false,width=0.239\textwidth]{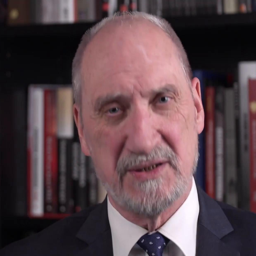}
\\

\rotatebox{90}{\small\textbf{StableVSR}}
&
\includegraphics[interpolate=false,width=0.239\textwidth]{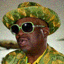}
&
\includegraphics[interpolate=false,width=0.239\textwidth]{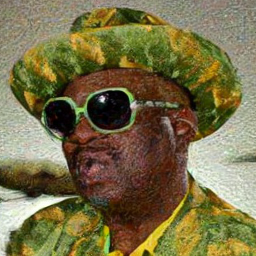}
&
\includegraphics[interpolate=false,width=0.239\textwidth]{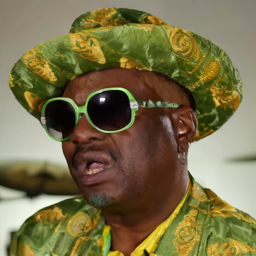}
&
\includegraphics[interpolate=false,width=0.239\textwidth]{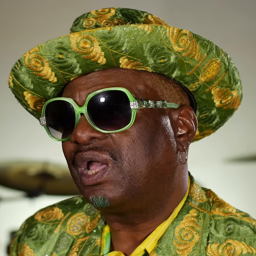}
\\

\rotatebox{90}{\small\textbf{StableVSR+BM3D}}
&
\includegraphics[interpolate=false,width=0.239\textwidth]{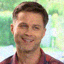}
&
\includegraphics[interpolate=false,width=0.239\textwidth]{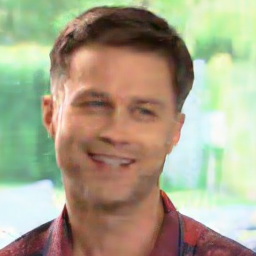}
&
\includegraphics[interpolate=false,width=0.239\textwidth]{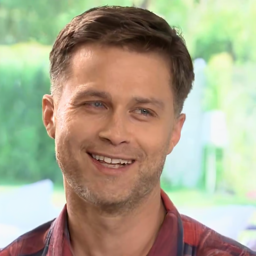}
&
\includegraphics[interpolate=false,width=0.239\textwidth]{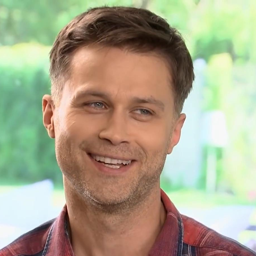}

\end{tabular}

\caption{
\textbf{Extended qualitative comparison for Figure~\ref{fig:qualitative}
under noisy degradation (Part II).}
Continuation of Figure~\ref{fig:additional_qualitative_noisy_1} with additional competing methods. 
}
\label{fig:additional_qualitative_noisy_2}
\end{suppfigure}

\end{document}